\documentclass{article}

\usepackage[final,nonatbib]{neurips_2020}

\usepackage{sidecap}
\usepackage[utf8]{inputenc} %
\usepackage[T1]{fontenc}    %
\usepackage{hyperref}       %
\usepackage{url}            %
\usepackage{booktabs}       %
\usepackage{amsfonts}       %
\usepackage{nicefrac}       %
\usepackage{microtype}      %

\usepackage{grffile} %
\usepackage[percent]{overpic}
\usepackage{floatrow}

\usepackage{sidecap}

\usepackage{enumitem}
\usepackage{amssymb}
\usepackage{amsmath}
\usepackage{mathtools}
\usepackage{graphicx}
\usepackage{dsfont}  %
\usepackage{booktabs} %
\usepackage{multirow}
\usepackage{algorithm}
\usepackage{algorithmic}
\usepackage{booktabs} %
\usepackage{color}
\usepackage{mwe,tikz}
\usepackage[percent]{overpic}
\usepackage{rotating} %
\usepackage{upgreek}

\newcommand{\mywidth}{0.21}

\newcommand{\tildeapprox}{{\raise.17ex\hbox{$\scriptstyle\sim$}}}
\newcommand{\secref}[1]{Section \ref{#1}}
\newcommand{\figref}[1]{Figure \ref{#1}}

\renewcommand{\eqref}[1]{Eq.~(\ref{#1})}

\DeclareMathOperator*{\argmin}{\arg\!\min}
\DeclareMathOperator*{\argmax}{\arg\!\max}

\newcommand\reals{\mathbb{R}}
\newcommand{\phio}{\Phi_O}
\newcommand{\phia}{\Phi_A}
\newcommand{\lowphio}{\phi_o}
\newcommand{\lowphia}{\phi_a}
\newcommand{\hatphio}{\hat{\phi}_o}
\newcommand{\hatphiotag}{\hat{\phi}_{o'}}
\newcommand{\hatphia}{\hat{\phi}_a}
\newcommand{\hatphiatag}{\hat{\phi}_{a'}}
\newcommand{\phiao}{\phi_{ao}}
\newcommand{\hao}{h_{ao}}
\newcommand{\nao}{n_{ao}}

\newcommand{\ho}{h_o}
\newcommand{\ha}{h_a}

\renewcommand\vec[1]{\mathbf{#1}}
\newcommand{\x}{\vec{x}}
\renewcommand{\S}{\mathcal{S}}
\newcommand{\U}{\mathcal{U}}
\newcommand{\A}{\mathcal{A}}
\newcommand{\X}{\mathcal{X}}

\renewcommand{\L}{\mathcal{L}}
\newcommand{\I}{\mathcal{I}}
\renewcommand{\O}{\mathcal{O}}

\makeatletter
\newcommand*{\indep}{%
  \mathbin{%
    \mathpalette{\@indep}{}%
  }%
}
\newcommand*{\nindep}{%
  \mathbin{%
    \mathpalette{\@indep}{\not}%
  }%
}
\newcommand*{\@indep}[2]{%
  \sbox0{$#1\perp\m@th$}%
  \sbox2{$#1=$}%
  \sbox4{$#1\vcenter{}$}%
  \rlap{\copy0}%
  \dimen@=\dimexpr\ht2-\ht4-.2pt\relax
  \kern\dimen@
  {#2}%
  \kern\dimen@
  \copy0 %
} 

\definecolor{atomictangerine}{rgb}{0.8, 0.2, 0.1}
\definecolor{turq}{rgb}{0.0, 0.5, 0.5}
\definecolor{darkturq}{rgb}{0.0, 0.4, 0.4}
\definecolor{bright}{rgb}{0.8, 0.1, 0}
\definecolor{darkgray}{gray}{0.3}
\definecolor{gray}{gray}{0.5}
\definecolor{mahogany}{rgb}{0.6, 0.05, 0.05}
\definecolor{myblue}{rgb}{0.3,0.05,0.9}
\definecolor{darkgreen}{rgb}{0.1,0.5,0.0}

\definecolor{olive}{rgb}{0.537, 0.627, 0.318}
\definecolor{green}{rgb}{0.22, 0.463, 0.114}
\definecolor{grey}{rgb}{0.4, 0.4, 0.4}
\definecolor{blue}{rgb}{0.435, 0.659, 0.863}
\definecolor{pink}{rgb}{0.761, 0.482, 0.627}
\definecolor{darkpink}{rgb}{0.561, 0.282, 0.427}

\newcommand\edit[1]{#1}

\title{\edit{A causal view of compositional zero-shot recognition}}

\author{
Yuval Atzmon\textsuperscript{1}
Felix Kreuk \textsuperscript{1,2}
Uri Shalit \textsuperscript{3}
Gal Chechik\textsuperscript{1,2} \\
\textsuperscript{1}NVIDIA Research, Tel Aviv, Israel \\
\textsuperscript{2}Bar-Ilan University, Ramat Gan, Israel \\
\textsuperscript{3}Technion - Israel Institute of Technology \\
yatzmon@nvidia.com, gchechik@nvidia.com,
\\}

\begin{document}

\maketitle
\begin{abstract}
People easily recognize new visual categories that are new combinations of known components. This \textit{compositional generalization} capacity is critical for learning in real-world domains like vision and language because the long tail of new combinations dominates the distribution. Unfortunately, learning systems struggle with compositional generalization because they often build on features that are \textit{correlated} with class labels even if they are not ``essential'' for the class. This leads to consistent misclassification of samples from a new distribution, like new combinations of known components.

Here we describe an approach for compositional generalization that builds on causal ideas. First, we describe compositional zero-shot learning from a causal perspective, and propose to view zero-shot inference as finding ``\textit{which intervention caused the image?}''. Second, we present a causal-inspired embedding model that learns disentangled representations of elementary components of visual objects from correlated (confounded) training data.
We evaluate this approach on two datasets for predicting new combinations of attribute-object pairs: A  well-controlled synthesized images dataset and a real-world dataset which consists of fine-grained types of shoes. We show improvements compared to strong baselines. \edit{Code and data are provided in \url{https://github.com/nv-research-israel/causal_comp}}
\end{abstract}

\section{Introduction}
\begin{figure}[t]
    \centering
    \begin{overpic}[width=0.28\linewidth, trim={8.2cm 1cm 7.2cm 0cm},clip]{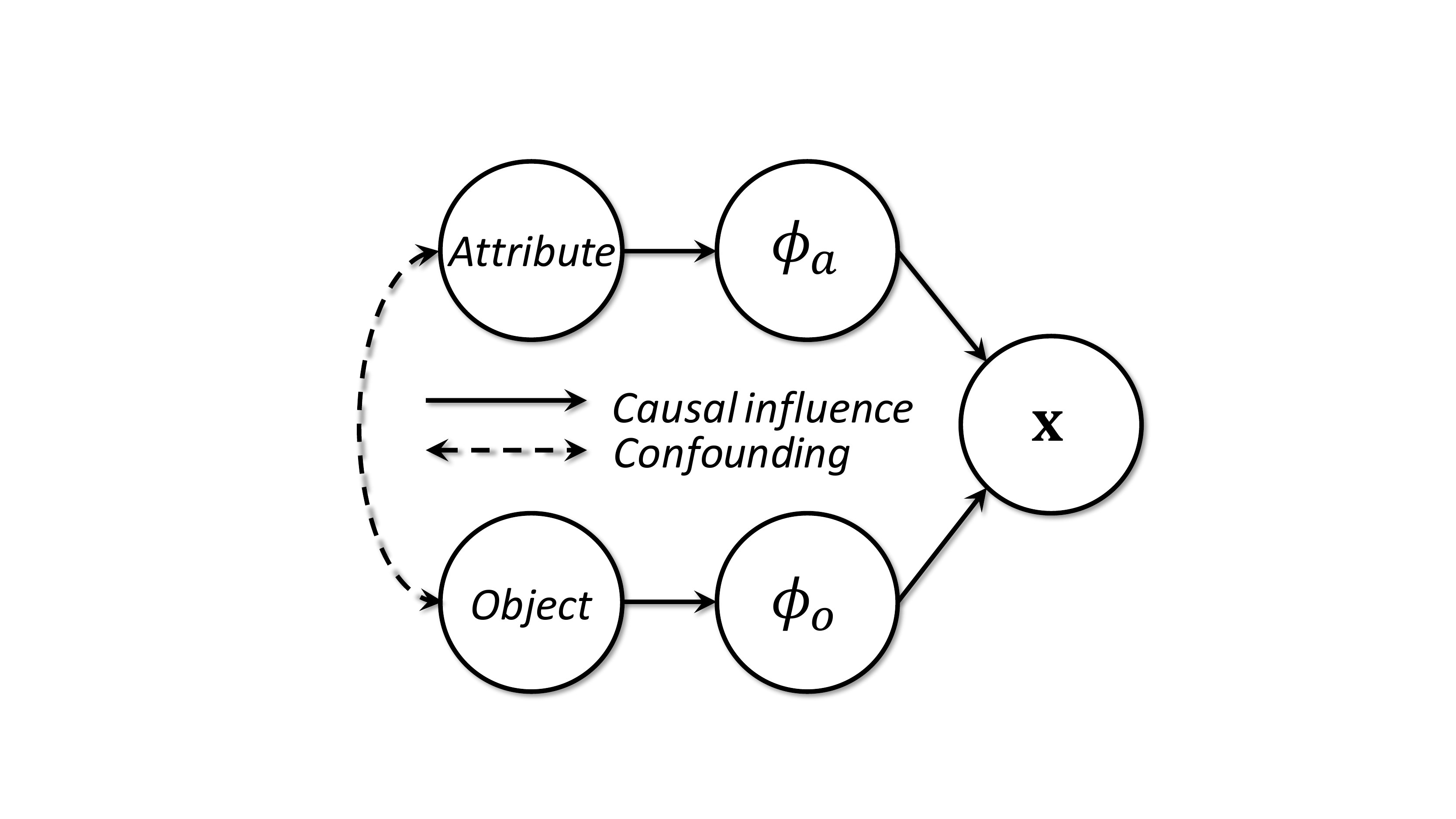}
    \put (90,10) {(a)}
    \end{overpic}
    ~\hspace{5pt}~
    \begin{overpic}[width=0.29\linewidth, trim={6.7cm 1cm 7.5cm 0cm},clip]{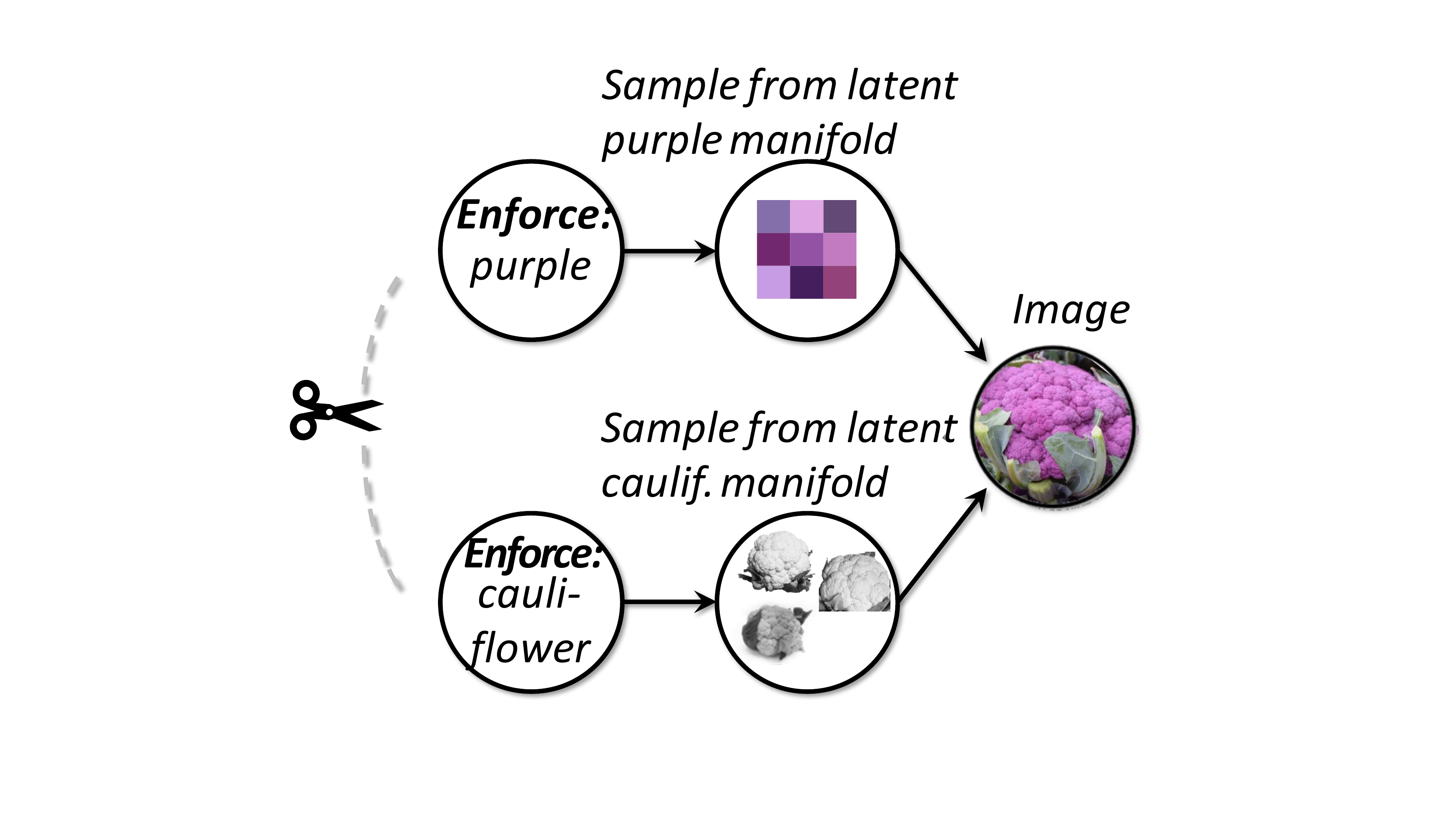}
    \put (90,10) {(b)}
    \end{overpic}
    ~\hspace{5pt}~
    \begin{overpic}[width=0.35\linewidth, trim={7.5cm 1cm 4.4cm 0cm},clip]{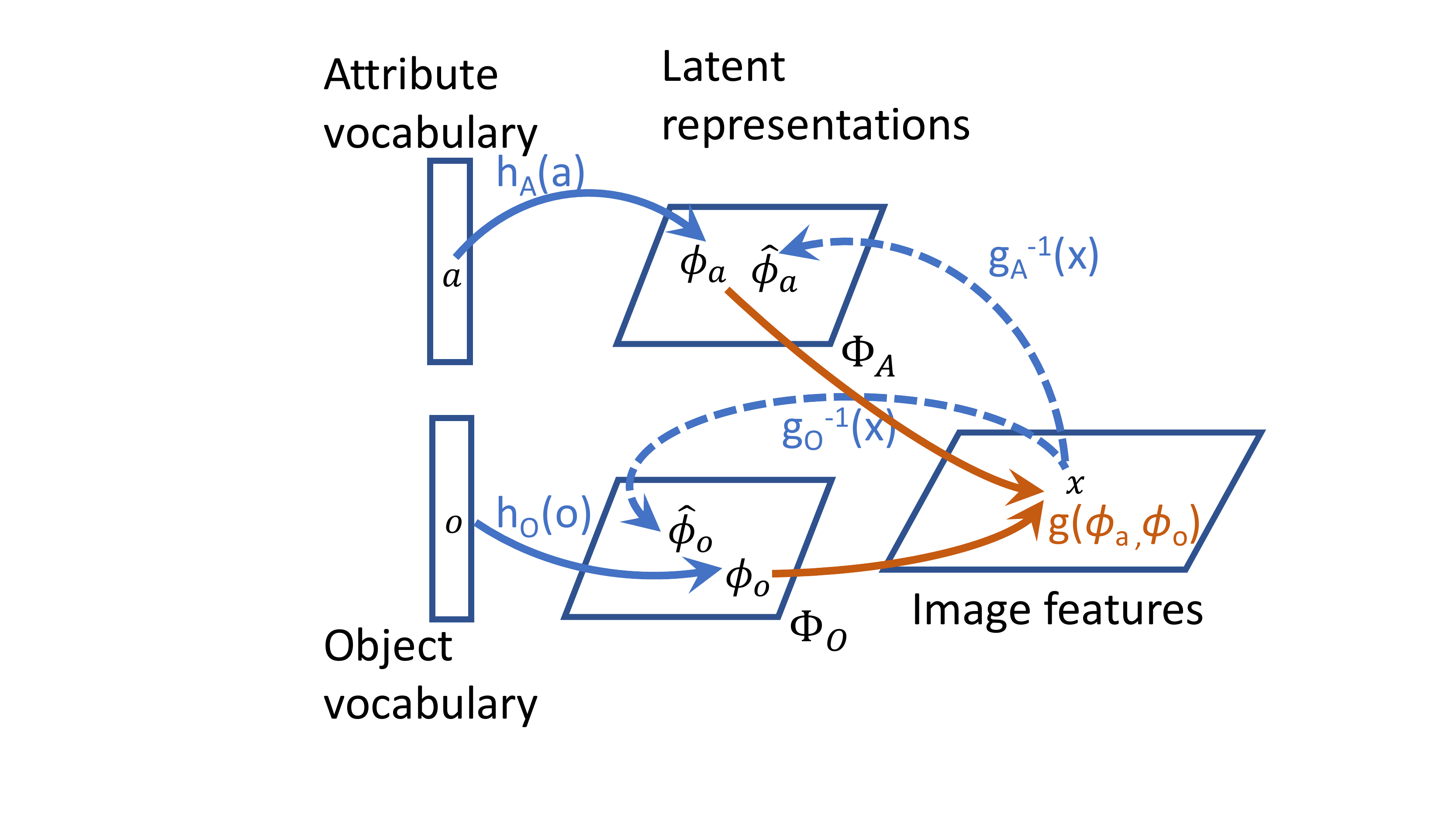}
    \put (90,8) {(c)}
    \end{overpic}
    \caption{\textbf{(a)} The causal graph that generates an image. The solid arrows represent the real-world processes by which the two categorical variables ``Object'' and ``Attribute'' each generate ``core features'' \cite{heinze2017conditional, gong2016domain} $\lowphio$ and $\lowphia$. The core features then jointly generate an image feature vector $\x$. The core features are assumed to be stable for unseen combinations of objects and attributes.
    The dotted double-edged arrows between the Object and Attribute nodes indicates that there is a process ``confounding'' the two: they are not independent of each other. \textbf{(b)} An intervention that generates a test image of a purple-cauliflower, by enforcing $a=purple$ and $o=cauliflower$. It cuts the confounding link between the two nodes \cite{pearl2000causality} and changes the joint distribution of the nodes to the ``\textit{interventional distribution}''. \textbf{(c)} Illustration of the learned mappings, detailed in \secref{sec:approach}}.
    \label{fig:causal_graph}
    \vspace{-15pt}
\end{figure}

Compositional zero-shot recognition is the problem of learning to recognize new combinations of known components. 
People seamlessly recognize and generate new compositions from known elements and 
\textit{Compositional Reasoning} is considered a hallmark of human intelligence  \cite{Lake2014, Lake2017, atzmon2016learning, andreasNAACL}. As a simple example, people can recognize a purple cauliflower even if they have never seen one, based on their familiarity with cauliflowers and with other purple objects (Figure \ref{fig:causal_graph}b). Unfortunately, although \textit{feature compositionality} is a key design consideration of deep networks, 
current deep models struggle when required to generalize to new \textit{label compositions}.
This limitation has grave implications for machine learning because the heavy tail of unfamiliar compositions dominates the distribution of labels in perception, language, and decision-making problems. 

Models trained from data tend to fail with compositional generalization for two fundamental reasons: \textbf{distribution-shift} and \textbf{entanglement}. First, recognizing new combinations is an extreme case of distribution-shift inference, where label combinations at test time were never observed during training (zero-shot learning). As a result, models learn correlations during training that hurt inference at test time. For instance, if all cauliflowers in the training set are white, the correlation between the color and the class label is predictive and useful. A correlation-based model like (most) deep networks will learn to associate cauliflowers with the color white during training, and may fail when presented with a purple cauliflower at test time. \edit{For the current scope, we put aside the fundamental semantic question about what defines the class of an object (cauliflower), and assume that it is given or determined by human observers}. 

The second challenge is that the training samples themselves are often labeled in a compositional way,  and disentangling their ``elementary'' components from examples is often an ill-defined problem \cite{locatello2018challenging}. For example, for an image labeled as white cauliflower, it is hard to tell which visual features capture being a cauliflower, and which, being white. In models that learn from data the representation of these terms may be inherently entangled, and it would be hard to separate which visual features represent white and which represent a cauliflower.

These two challenges are encountered when learning deep \textit{discriminative} models from data. For example, consider a simple model that learns the concept ``cauliflower'', by training a deep model over all cauliflower images (VisProd \cite{ATTOP}), and the same for the concept "white". At  inference time, simply select the most likely attribute $\hat{a}=\arg\max_a p(a|\x)$ and, independently, the most likely object $\hat{o}=\arg\max_o p(o|\x)$.  Unfortunately, this model, while quite powerful, tends to be sensitive to training-specific correlations in its input.

Here we propose to address compositional recognition by modelling images as being generated, or caused, by real-world entities (labels) (\figref{fig:causal_graph}). This model recognizes that the distribution $p(Image\!\!=\!\!\x|Attr\!\!=\!\!a,Obj\!\!=\!\!o)$ is more likely to be stable across the train and test environments  \big($p_{test}(\x|a,o) = p_{trn}(\x|a,o)$\big) \cite{scholkopf2019causality, pearl2000causality,suter2018robustly}: it means that unlike objects or attributes by themselves, \emph{combinations} of objects and attributes generate the same distribution over images in train and test.
We propose to consider images of unseen combinations as generated by interventions on the attribute and object labels. In causal inference, intervention means that the value of a random variable is forced to some value, without affecting its causes (but affecting other variables that  depend on it, \figref{fig:causal_graph}b). We cast zero-shot inference as the problem of finding which intervention caused a given image.

In the general case, the conditional distribution $p(\x|a,o)$ can have arbitrary an complex structure and may be hard to learn. We explain how treating labels as causes, rather than as effects of the image, 
reveals an independence structure that makes it easier to learn $p(\x|a,o)$. We propose conditional independence constraints applied to the structure of this distribution and show how the model can be learned effectively from data.

The paper makes the following novel contributions: First, we provide a new formulation of compositional zero-shot recognition using a causal perspective. Specifically, we formalize inference as a problem of finding the most likely intervention. 
Second, we describe a new embedding-based architecture that infers causally stable representations for compositional recognition. %
Finally, we demonstrate empirically that in two challenging datasets, our architecture better recognizes new unseen attribute-object compositions  compared to previous methods.

\vspace{-5pt}
\section{Related work}
\vspace{-5pt}

\textbf{Attribute - object compositionality:}
\cite{misra2017red} studied decomposing attribute-object combinations. They embedded attributes and object classes using deep networks pre-trained on other large datasets.
\cite{nagarajan2018attributes} proposed to view attributes as linear operators in the embedding space of object  word-embeddings. Operators are trained to keep  transformed objects similar to the corresponding image representation. %
\cite{nan2019recognizing} proposed a method similar to \cite{misra2017red} with an additional decoding loss.  \cite{wei2019adversarial} used GANs to generate a feature vector from the label embedding.
\cite{TMN2019} trained a set of network modules, jointly with a gating network that rewires the modules according to embeddings of attribute and object labels.
\cite{SymNet2020} is a very recent framework inspired by group theory that incorporates symmetries in label-space. %

\textbf{Compositional generalizations:}
Several papers devised datasets to directly evaluate compositional generalization for vision problems by creating a test set with new combinations of train-set components. \cite{johnson2017clevr} introduced a synthetic dataset inherently built with compositionality splits. \cite{agrawal2017c, bahdanau2019closure} introduced new compositional splits of VQA datasets \cite{agrawal2017vqa} and show that the performance of existing models degrades under their new setting. \cite{kato2018compositional} used a knowledge graph for composing classifiers for verb-noun pairs.  \edit{\cite{atzmon2016learning} proposed an experimental framework for measuring compositional generalization and showed that structured prediction models outperform image-captioning models.}

\edit{\textbf{Zero-shot Learning (ZSL):} Compositional generalization can be viewed as a special case of zero-shot learning \cite{xianCVPR, DAP, COSMO}, where a classifier is trained to recognize (new) unseen classes based on their semantic description, which can include a natural-language textual description \cite{ZEST} or predefined attributes \cite{DAP, LAGO}. To discourage attributes that belong to different groups from sharing low-level features, \cite{Jayaraman} proposed a group-sparse regularization term.}

\textbf{Causal inference for domain adaptation:}
Several recent papers take a causal approach to describe and address domain adaptation problems. This includes early work by
\cite{scholkopf2012causal} and \cite{zhang2013domain}. Adapting these ideas to computer vision, \cite{gong2016domain} were one of the first papers to propose a causal DAG describing the generative process of an image as being generated by a ``domain'', which generates a label and an image. They use this graph for learning invariant components that transfer across domains. \cite{CIDDG, AFLAC} extended \cite{gong2016domain} with adversarial training \cite{ganin2016domain}. It learned a single discriminative classifier, $p(o|\x)$, that is robust against  domain shifts and accounts for the dependency of classes on domains.
When viewing attributes as ``domains'', one of the independence terms in their model corresponds to one term (c) in \eqref{eq:indepCondAO}. %
 \cite{kilbertus2018generalization, scholkopf2012causal} discusses image recognition as an ``anti-causal'' problem, inferring causes from effects. 
\cite{bengio2019meta, ke2019learning} studied learning causal structures under sparse distributional shifts. \edit{\cite{lopez2017discovering} proposed to learn causal relationships in images by detecting the causal direction of two random variables.
\cite{CausalImitationLearn2019} used targeted interventions to address distribution shifts in imitation learning. \cite{kocaoglu2018causalgan} learned a conditional-GAN model jointly with a causal-model of label distribution.} \cite{arjovsky2019invariant} proposed a regularization term to improve robustness against distributional changes. Their view is complementary to this paper in that they model the labeling processes, where images cause labels, while our work focuses on the data generating process (labels cause images). 
\cite{heinze2017conditional} proposed a similar causal DAG for images, while adding \textit{auxiliary} information such as that some images are of the \textit{same instance} with a different ``style''. This allowed the model to identify core features. The approach described in this paper does not use such auxiliary information. It also views the ``object'' and ``attribute'' as playing mutual roles, making their inferred representations invariant to each-other.

\textbf{Unsupervised disentanglement of representations:}
Several works use a VAE \cite{kingma2013auto} approach for unsupervised disentanglement of representations \cite{locatello2019disentangling, higgins2017beta, burgess2018understanding, chen2018isolating, rubenstein2018learning, mathieu2018disentangling}.
This paper focuses on a different problem setup: (1) Our goal is to infer a joint attribute-object pair, disentangling the representation is a useful byproduct. (2) In our setup, attribute-object combinations are dependent in the training data, and new combinations may be observed at test time. (3) We do not use unsupervised learning. (4) We take a simpler embedding based approach. %

\vspace{-7pt}
\section{Method overview} 
\vspace{-7pt}
\label{sec:method_overview}
We start with a descriptive overview of our approach. For simplicity, we skip here the causal motivation and describe the model in an informal way from an embedding viewpoint.

Our model is designed to estimate $p(\x|a,o)$, the likelihood of an image feature vector $\x$, conditioned on a tuple $(a,o)$ of attribute-object labels. For inference, we iterate over all combinations of labels and select $(\hat{a}, \hat{o}) = \argmax_{a,o} p(\x|a,o)$.

\textbf{To estimate} the distribution $p(\x|a,o)$, our model learns two embedding spaces: $\Phi_A$ for attributes, and $\Phi_O$ for objects (see  \figref{fig:causal_graph}a). These spaces can be thought of as semantic embedding spaces, where an attribute $a$ (say, ``white") has some dense prototypical representation $\phi_a\in\Phi_A$, and an object $o$ (say, a cauliflower) has a dense representation $\phi_o\in\Phi_o$.  
Given a new image $\x$, we learn a mapping to three spaces.
First, an inferred attribute embedding $\hatphia\in\Phi_A$ represents the attribute seen in the image (say, how white is the object). Second, an inferred object embedding $\hatphio\in\Phi_O$ represents the object seen in the image (say, how ``cauliflowered'' it is). Finally, we also represent the image in a general space of image features. %

\textbf{Learning} the parameters of the model involves learning the three mappings above.  
In addition, we learn the representation of the attribute prototype (``White'') $\lowphia\in\phia$ and the representation of the object prototype (``Cauliflower") $\lowphio\in\phio$.  Very naturally, we want that a perceived attribute $\hatphia$ would be embedded close to its attribute prototype $\lowphia$. Our loss captures this intuition. 
Finally, we also aim to have the representation spaces of attributes and objects statistically independent. The intuition is that we want to keep the  representation of an object (cauliflower) independent of the attribute (white), so we can recognize that object when seen with new attributes (purple cauliflower). 

At this point, %
the challenge remains to build a principled model that can be learned efficiently from data. 
We now turn to the formal and detailed description of the approach.

\vspace{-5pt}
\section{A causal formulation of compositional zero-shot recognition}
\vspace{-5pt}
\label{sec:approach}

We put forward a causal perspective that treats labels as causes of an image, rather than its effects. This direction of dependencies is consistent with the mechanism underlying natural image generation and, as we show below, allows us to recognize unseen label combinations. 

Figure \ref{fig:causal_graph} presents our causal generative model. %
We consider two ``elementary factors'' which are categorical variables called ``Attribute'' $a\in\A$ and ``Object" $o\in\O$, and are dependent (confounded) in the training data. {\edit{As one example, European swans (Cygnus) are white but Australian ones are black. The data collection process may make `white' and `swan' confounded if collected in Europe, even-though this dependency does not apply in Australia.}}. %
The model also has two semantic representation spaces: one for attributes $\phia=\reals^{d_A}$ and another for objects $\phio=\reals^{d_O}$. 
An attribute $a$ induces a distribution $p(\lowphia|a)$ over the representation space, which we model as a Gaussian distribution. We denote by $\ha$ a function that maps a categorical attributes to the center of this distribution in the semantic space $\ha:\A\rightarrow\phia$ (\figref{fig:causal_graph}c). The conditional distribution is therefore  $\lowphia\sim\mathcal{N}(\ha, \sigma^2_a I)$.  
We have a similar setup for $p(\lowphio|o)\sim \mathcal{N}(\ho, \sigma^2_o I)$.

Given the semantic embedding of the attribute and object, the probability of an image feature vector $\x\in\X$ is determined by the representations $p(\x|\lowphia, \lowphio)$, which we model as Gaussian, w.r.t a mapping $g$,  $\x\sim\mathcal{N}(g(\lowphia, \lowphio), \sigma^2_x I)$. 
$\lowphia$ and $\lowphio$ can be viewed as an encoding of ``core features'', namely encoding a representation of attribute and object that is ``stable" in the training set and test set, as proposed by \cite{heinze2017conditional, gong2016domain}. Namely, the conditional distributions $p(\lowphia|a)$ and $p(\lowphio|o)$ do not substantially change for unseen combinations of attributes and objects.

We emphasize that our causal graph is premised on the belief that what we use as objects and attributes are truly distinct aspects of the world, giving rise to different core features. For attributes that have no physical meaning in the world, it may not be possible to postulate two distinct processes giving rise to separate core features.

\vspace{-5pt}
\subsection{Interventions on elementary factors}
\vspace{-5pt}
Causal inference provides a formal mechanism to address the confounding effect through a ``do-intervention''\footnote{Our formalism and model can be extended to include other types of intervention on the joint distribution of attributes and objects. For simplicity, we focus here on the most-common ``do-intervention''.}. A ''Do-intervention`` \textit{overrides} the joint distribution $p_{trn}(a,o)$, enforcing $a,o$ to specific values and propagates them through the causal graph. %
With this propagation, an intervention \textit{changes the joint distribution} of nodes in the graph. 
Therefore, a test image is generated according to a new joint-distribution, denoted in causal language as the \textit{interventional distribution} $p^{do(A=a,O=o)}(\x)$. 
Thus, for zero-shot learning, we postulate that inference about a test image is equivalent to asking: \textbf{Which intervention on attributes and objects caused the image?}

\vspace{-5pt}
\section{Inference}
\vspace{-5pt}
\label{sec:the_model}

We propose to infer the attribute and object by choosing the most likely interventional distribution:
\begin{eqnarray}
    (\hat{a},\hat{o}) &=& \argmax_{a,o \in \A\times \O} \,\, p^{do(A=a,O=o)}(\x).
    \label{eq_px_do}
\end{eqnarray}
This inference procedure is more stable than the discriminative zero-shot approach, since the generative conditional distribution is equivalent to the interventional distribution \cite{scholkopf2012causal}.
\begin{eqnarray}
    p^{do(A=a,O=o)}(\x) = p(\x|a,o).
    \label{eq_p_inference}
\end{eqnarray}
This holds both for training and test, so we simply write $p(\x|a,o)$.
This likelihood depends on the core features $\phi_A, \phi_O$ which are latent variables; Computing the likelihood exactly requires to marginalize (integrate) over the latent variables. Since this integral is very hard to compute, we take a ``hard" approach instead, evaluating the integrand at its most likely value. Since  $\phi_A, \phi_O$ 
are not known, we estimate them from the image $x$, by learning a mapping function $\hatphia = g_A^{-1}(x)$ (see \figref{fig:causal_graph}c). 
The supplemental describes  these approximation steps in details. It shows that the negative log-likelihood  $-\log p(\x|a,o)$ can be approximated \edit{by}
\begin{equation}
    \hat{L}(a,o) = 
    \frac{1}{\sigma^2_a}||\hatphia - \ha||^2 +
    \frac{1}{\sigma^2_o}||\hatphio - \ho||^2 + 
    \frac{1}{\sigma^2_x}||\x - g(\ha,\ho)||^2 \quad.
    \label{eq_inference_NLL}
\end{equation}
Here, $\ha$, $\ho$ and $g(\ha,\ho)$ are the \edit{parameters of the Gaussian distributions} of $\lowphia$, $\lowphio$ and $\x$. %
The factors $a$ and $o$ are inferred by taking the $\argmin_{a,o} \hat{L}(a,o)$ of \eqref{eq_inference_NLL}.
\edit{Note that in the causal graph, $\lowphia, \lowphio$ are parent nodes of the image $\x$, but $\hatphia, \hatphio$ are estimated from $x$ and are therefore child nodes of $\x$, and therefore do not immediately follow the conditional independence relations that  $\lowphia, \lowphio$ obey. We elaborate on this point in section \ref{sec:learning}.  }

\vspace{-8pt}
\section{Learning}
\label{sec:learning}
\vspace{-8pt}

Our model consists of five learned mappings:  $h_A$, $h_O$, $g$, $g^{-1}_A$ and $g^{-1}_O$, illustrated in  \figref{fig:causal_graph}c. All mappings are modelled using MLPs.%
We aim to learn the parameters of these mappings such that the (approximated) negative log-likelihood of \eqref{eq_inference_NLL} is minimized. In addition, we also include in the objective several regularization terms designed to encourage properties that we want to induce on these mappings. Specifically, the model is trained with a linear combination of three losses.
\begin{eqnarray}
    \L = \L_{data} + \lambda_{indep} \L_{indep} + \lambda_{invert}\L_{invert},
\end{eqnarray}
where $\lambda_{indep}\ge0$ and $\lambda_{invert}\ge0$ are hyperparameters. We now discuss these losses in detail.

\noindent\textbf{(1) Data Likelihood loss.} The first component of the loss, $\L_{data}$, corresponds to the (approximate) negative log likelihood of the model, as described by \eqref{eq_inference_NLL} 
\begin{eqnarray}
    \L_{data} = ||\ha - g^{-1}_A(\x)||^2  + ||\ho - g^{-1}_O(\x)||^2  + \lambda_{ao}\L_{triplet}\Big(\x, (a, o), (a,o)_{neg} \Big). %
\end{eqnarray}
For easier comparisons with \cite{ATTOP}, we replaced the rightmost term in \eqref{eq_inference_NLL} with the standard triplet loss $\L_{triplet}$ with Euclidean distance $||\x - g(\ha,\ho)||^2$. $\lambda_{ao}\ge0$ is a hyperparameter.

\noindent\textbf{(2) Independence loss.} The second component of the loss $\L_{indep}$ is designed to capture conditional-independence relations \edit{and apply them to the reconstructed core factors $\hatphia, \hatphio$. By that, the following property is encouraged: $p^{do(O=o)}(\hatphio)\!\!\approx\!\! p^{do(A=a,O=o)}(\hatphio)$ and  $p^{do(A=a)}(\hatphia)\!\!\approx\!\! p^{do(A=a,O=o)}(\hatphia)$. Namely,  learning a representation of objects that is robust to attribute interventions, and vice versa.%
}

\edit{In more detail, }the causal graph (\figref{fig:causal_graph}a) dictates  conditional-independence relations for the latent core factors $\lowphia, \lowphio$: 
\begin{align}
    \label{eq:indepCondAO}
    (a) \quad \lowphia &\indep O |A=a \quad\quad (b) \quad \lowphia \indep \lowphio |A=a  , \\ \nonumber
    (c) \quad \lowphio &\indep A |O=o \quad\quad (d) \quad \lowphia \indep \lowphio |O=o.
\end{align} 
\edit{These relations reflect the independent mechanisms that generate the training data.}

Since the core factors $\lowphia$, $\lowphio$, are latent and not observed, we wish that their reconstructions $\hatphia$ and $\hatphio$ maintain approximately the same independence relations. 
\edit{For example, we encourage $(\hatphio \indep A |O=o)$ to capture the independence in Eq. (\ref{eq:indepCondAO})c. %
}

To learn mappings that adhere to these statistical independences over $\hatphia$ and $\hatphio$, we regularize the learned mappings using a differentiable measure of statistical dependence. Specifically, we use the Hilbert-Schmidt Information Criterion (HSIC)  \cite{gretton2005measuring,gretton2008kernel}. HSIC is a non-parametric method for estimating the statistical dependence between samples of two random variables, based on an implicit embedding into a universal reproducing kernel Hilbert space. In the infinite-sample limit, the HSIC between two random variables is $0$ if and only if they are independent \cite{gretton2008kernel}. HSIC also has a simple finite-sample estimator which is easily calculated and is differentiable w.r.t. the input variables. In supplemental \secref{sec:indep_loss}, we describe the details of $\L_{indep}$ and how it is optimized with HSIC, \edit{and why minimizing $L_{indep}$ indeed encourages the property $p^{do(O=o)}(\hatphio) \!\!\approx\!\! p^{do(A=a,O=o)}(\hatphio)$}. This minimization can be viewed as minimizing the ``Post Interventional Disagreement'' (PIDA) metric of \cite{PIDA}, a recently proposed measure of disentanglement of representations. We explain this perspective in more detail in the supplemental    \edit{(\ref{sec:PIDA})}.

There exist alternative measures for encouraging statistical independence, such as adversarial training \cite{ganin2016domain, CIDDG, AFLAC} or techniques based on mutual information \cite{CMI, sanchez2019learning,kim2019learning}. \edit{HSIC has the benefit that it is non-parametric and therefore does not require training an additional network. It was} easy to optimize, and was provide useful in previous literature \cite{ tsai2017improving, song2012feature, mukherjee2017deep, greenfeld2019robust}.

\noindent\textbf{(3) Invertible embedding loss.} 
The third component of the loss, $\L_{invert}$,  encourages the label-embedding mappings $\ha$, $\ho$, and $g(\ha,\ho)$ to preserve information about their source labels when minimizing $\L_{data}$. \edit{Without this term, minimizing $||\hatphia-h_a||^2$ may reach trivial solutions because the model has no access to ground-truth values for $\lowphia$ or $\ha$ (same for $\lowphio, \ho$).} Similar to \cite{ATTOP}, we use a cross-entropy (CE) loss with a linear layer that classifies the labels that generate each embedding, and a hyperparameter  $\lambda_{g}$:
\begin{equation*}
    \L_{invert} = CE(a, f_a(\ha)) + CE(o, f_o(\ho)) + \lambda_{g}\big[CE(a,f_{ga}(g(\ha,\ho))) + CE(a,f_{go}(g(\ha,\ho)))\big].
\end{equation*}

\vspace{-5pt}
\section{Experiments}
\vspace{-5pt}

\subsection{Data}

Despite several studies of compositionality, current datasets used for evaluations are quite limited. 
Two main benchmarks were used in evaluations of previous literature: \textit{MIT states} \cite{isola2015discovering} and \textit{UT-Zappos50K} \cite{yu2014finegrained}. 

The MIT-states dataset was labeled automatically using early technology of image search engine based on text surrounding images. As a result, labels are often incorrect. We quantified label quality using human raters \edit{, and found that they are too noisy to be useful for proper evaluations.} \edit{In more detail, we presented images to human raters, along with candidate attribute labels from the dataset. Raters were asked to select the best and second-best attributes that describe the noun (multiple-choice setup). 
Only 32\% of raters selected the correct attribute for their first choice (top-1 accuracy), and only 47\% of raters had the correct attribute in one of their choices (top-2 accuracy).
The top-2 accuracy was only slightly higher than adding a random label on top of the top-1 label (yielding 42\%).
To verify that raters were attentive, we also injected  30 ``sanity'' questions that had two ``easy"  attributes, yielding top-2 accuracy of 100\%. See supplemental (\ref{sec:AMT}) for further details.
We conclude that this level of $\sim 70\%$ label noise is too noisy for evaluating noun-attribute compositionality. }

\noindent\textbf{Zappos:}
We evaluate our approach on the Zappos dataset, which consists of fine-grained types of shoes, like ``leather sandal'' or ``rubber sneaker''. It has 33K images, 16 attribute classes, and 12 object classes. We use the split of \cite{TMN2019} and the provided ResNet18 pretrained features. The split contains both seen pairs and unseen pairs for validation and test. It uses 23K images for training of 83 seen pairs, a validation set with 3K images from 15 seen and 15 unseen pairs, and a test set with 3K images from 18 seen and 18 unseen pairs. All the metrics we report for our approach and compared baselines are averaged over 5 random initializations of the model.

\begin{SCfigure}
    \centering   
    \includegraphics[width=0.8\textwidth, trim={0.cm 0cm 0cm 1cm},clip]{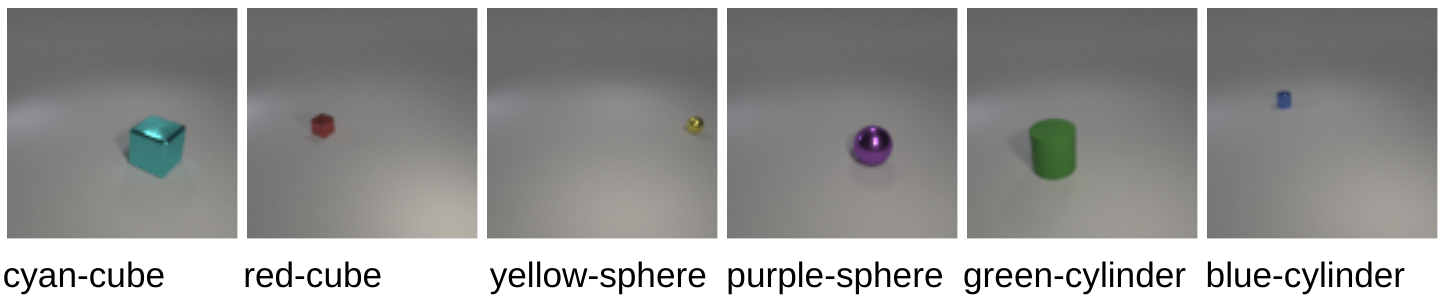}  
    \hspace{-10pt}
    \vspace{-20pt}
    \caption{Example images of AO-CLEVr dataset and their labels.}
    \label{fig_examples_clevr}
\end{SCfigure}

\noindent\textbf{AO-CLEVr:}
To evaluate compositional methods on a well-controlled clean dataset, we generated a synthetic-images dataset containing images of ``easy'' Attribute-Object categories. We used the CLEVr framework \cite{johnson2017clevr}, hence we name the dataset \textit{AO-CLEVr}.  %
AO-CLEVr has attribute-object pairs created from 8 attributes: \{ red, purple, yellow, blue, green, cyan, gray, brown \} and 3 objects \{sphere, cube, cylinder\}, yielding 24 attribute-object pairs. Each pair consists of 7500 images. Each image has a single object that consists of the attribute-object pair. The object is randomly assigned one of two sizes (small/large), one of two materials (rubber/metallic), a random position, and random lightning according to CLEVr defaults. 
See Figure \ref{fig_examples_clevr} for examples.

For cross-validation, we used two types of splits. The first uses the same unseen pairs for validation and test. This split allows us to quantify the potential generalization capability of each method. In the second split, which is harder, unseen validation pairs are not overlapping with the unseen test pairs. 
Importantly, we vary the ratio of unseen:seen pairs on a range of (2:8, 3:7, \ldots ,7:3), and for each ratio we draw 3 random seen-unseen splits. We report the average and the standard error of the mean (S.E.M.) over the three random splits and three random model initialization for each split.  We provide more details about the splits in the suppl.

\subsection{Compared methods} 

\noindent\textbf{(1) Causal}. Our approach as described in \secref{sec:the_model}. For Zappos it also \textit{learns} a single layer network to project the pretrained image features to the feature space $\X$. \edit{We also evaluate a variant named \textbf{Causal $\lambda_{indep}\!\!=\!\!0$}, which nulls the loss terms that encourage the conditional independence relations}

\noindent\textbf{(2) VisProd}: A common discriminatively-trained baseline \cite{nagarajan2018attributes, misra2017red}. It uses two  classifiers over image features to predict the attribute and object independently, and approximates $p(a, o|\x) \tildeapprox p(a|\x)p(o|\x)$.

\noindent\textbf{(3) VisProd\&CI}: A discriminatively-trained variant of our model. We use VisProd as a vanilla model, regularized by the conditional independence loss $\L_{indep}$, where we use the top network layer activations of attributes and objects as proxies for $\hatphia, \hatphio$. 

\noindent\textbf{(4) LE:} \textit{Label embedding} \cite{nagarajan2018attributes} trains a neural network to embed images and attribute-object labels to a joint feature space. LE is a vanilla baseline because it approximately models $p(\x|a,o)$, but without modelling the core-features. %

\noindent\textbf{(5) ATTOP:} \textit{Attributes-as-operators}
\cite{nagarajan2018attributes} views attributes as operators over the embedding space of object label-embeddings. We use the code of \cite{nagarajan2018attributes} to train ATTOP and LE.  %

\noindent\textbf{(6) TMN:} \textit{Task-modular-networks} \cite{TMN2019} trains a set of network modules jointly with a gating network. The gate rewires the modules according to embeddings of attributes and objects.
We used the implementation provided by the authors and followed their grid for hyperparameter search (details in suppl.). Our results differ on Zappos because we report an average over 5 random initializations rather than a single initialization as reported in \cite{TMN2019}. Some random initializations reproduce well their reported AUC metric.

We explicitly avoid using prior knowledge in the form of \textit{pretrained} label embeddings, because we are interested to quantify the effectiveness of our approach to naturally avoid unreliable correlations.
Yet, most of the methods we compare with, rely on pretrained embeddings. Thus, we provide additional results using random initialization for the compared methods, denoted by an asterisk (e.g. LE*). %

\textbf{Implementation details} of our approach and reproduced baselines are given in the supplemental.%
\label{sec:implementation_details}

\subsection{Evaluation} 

In zero-shot (ZS) attribute-object recognition, a training set $\mathcal{D}$ has $N$ labeled samples of images: $\mathcal{D} = \{ (\x_i, a_i, o_i), i=1 \dots N \}$ , where each $\x_i$ is a feature vector, $a_i \in \A$ is an attribute label, $o_i \in \O$ is an object label and each pair $(a_i, o_i)$ is from a set of (source) \textit{seen} pairs $\S \subset \A \times \O$. 
At test time, a new set of samples $\mathcal{D}'=\{\x_i, i=N+1 \dots N+M\}$ is given from a set of target pairs. The target set is a union of the set of seen pairs $\S$ with new \textit{unseen} pairs $\U \subset \A \times \O$, $\U \cap \S = \emptyset $. Our goal is to predict the correct pair of each sample.

\textbf{Evaluation metrics:} 
\label{sec:eval-metrics}
We evaluate methods by the accuracy of their top-1 prediction for recognizing seen and unseen attribute-object pairs. 
In AO-CLEVr, we compute the balanced accuracy across pairs,  namely, the average of per-class accuracy. This is the common metric to evaluate zero-shot benchmarks \cite{xian_2018,xianCVPR}. Yet, in Zappos, we used the standard (imbalanced) accuracy, to be consistent with the protocol of \cite{TMN2019},

We compute metrics in two main evaluations setups, which differ in their \textit{output} label space, namely, which classes can be predicted.
\noindent \textbf{\textit{(1)~Closed}}: Predictions can be from unseen class-pairs \textit{only}. 
\textbf{\textit{(2)~Open}}: Predictions can be from all pairs in the dataset, seen or unseen. This setup is also known as the \textit{generalized} zero-shot learning setup \cite{xianCVPR, chao}. Specifically, we compute:
\textbf{\textit{Seen}}: Accuracy is computed on \textit{samples from} seen class-pairs. 
\textbf{\textit{Unseen}}: Accuracy is computed on \textit{samples from} unseen class-pairs. 
\textbf{\textit{Harmonic mean}}:  A metric that quantifies the overall performance of both Open-Seen and Open-Unseen accuracy. 
It is defined as: $H = 2 (Acc_{seen} * Acc_{unseen}) /(Acc_{seen} + Acc_{unseen})$. 
For the harmonic metric, we follow the protocol of \cite{xian_2018, xianCVPR}, which does not take an additional post-processing step. We note that some papers \cite{TMN2019, SymNet2020, chao} used a different protocol averaging between seen and unseen. 
Finally, we report the \textbf{\textit{Area Under Seen-Unseen Curve} (AUSUC)}, which uses a post-processing step \cite{chao} to balance the seen-unseen accuracy. Inspired by the area-under-the-curve procedure, it adjusts the level of confidence of seen pairs by adding (or subtracting) a constant (see \cite{chao} for further details). To compute AUSUC, we sweep over a range of constants and compute the area under the seen-unseen curve.

For early stopping, we use (i) The Harmonic for the open setup. (ii) The closed accuracy for the closed setup. In Zappos, we followed \cite{TMN2019} and use the AUSUC for early stopping at the closed setup. 

\edit{All experiments were performed on a cluster of DGX-V100 machines. Training a single model for 1000 epochs on the $5:5$ AO-CLEVr split (with $\tildeapprox$80K samples) takes 2-3 hours.}

\vspace{-5pt}
\section{Results}
\vspace{-5pt}
We describe here the results obtained on AO-CLEVr (overlapping-split) and Zappos. Additional results are reported in the supplemental, including the full set of metrics and numeric values; using random initialization; results with the non-overlapping split  (showing a similar trend to the overlapping split); studying our approach in greater depth through ablation experiments; \edit{and an error analysis}.

\begin{figure}[t]
  \centering   
  \hspace{-35pt}
  \begin{overpic}[width=0.32\textwidth, trim={0cm 0cm 2.5cm 1.5cm},clip]{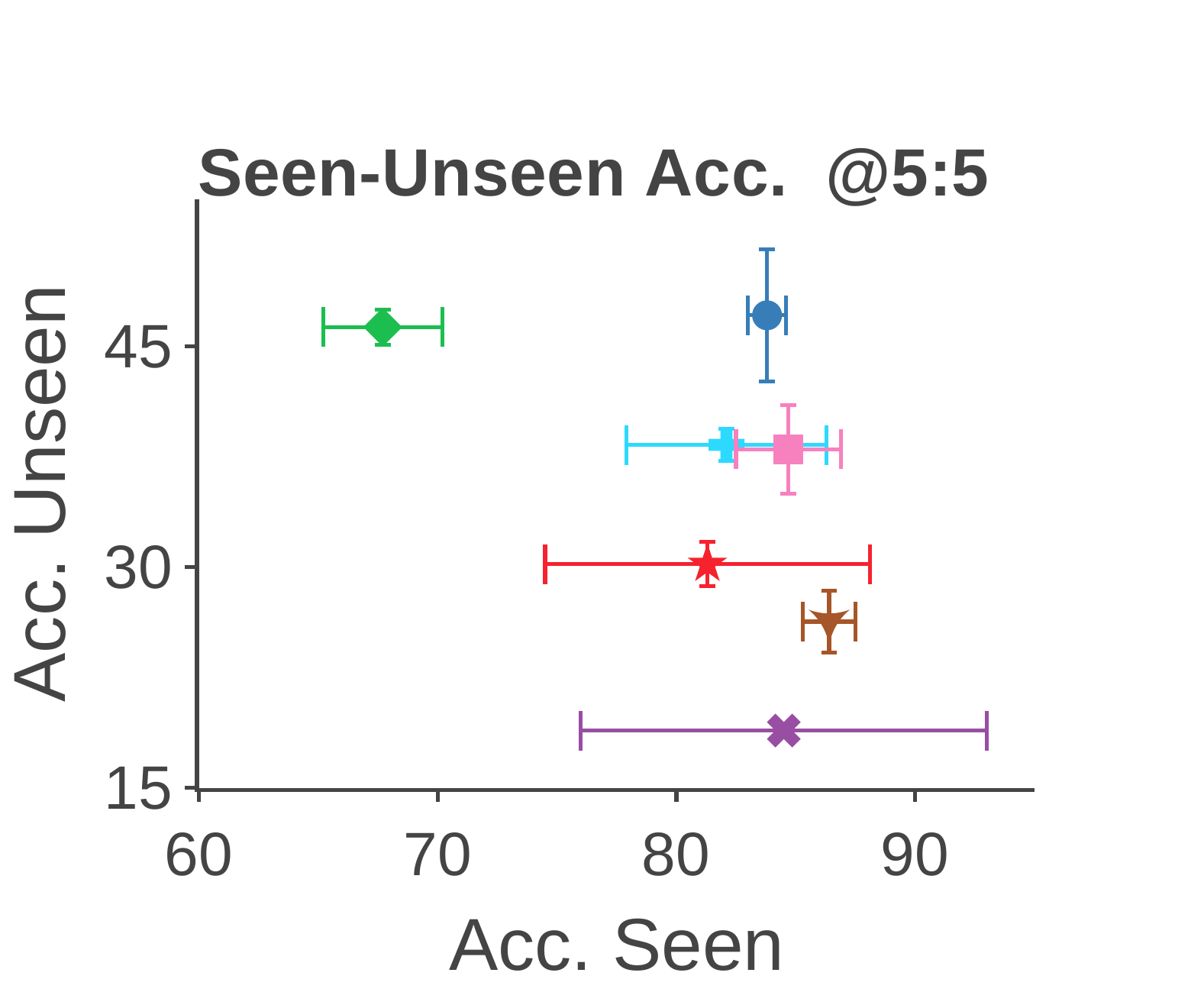} 
    \end{overpic}
  \hspace{20pt}
  \vspace{-5pt} 
\begin{overpic}[width=0.2\textwidth, trim={13.9cm 2cm 0.5cm 1.5cm},clip]{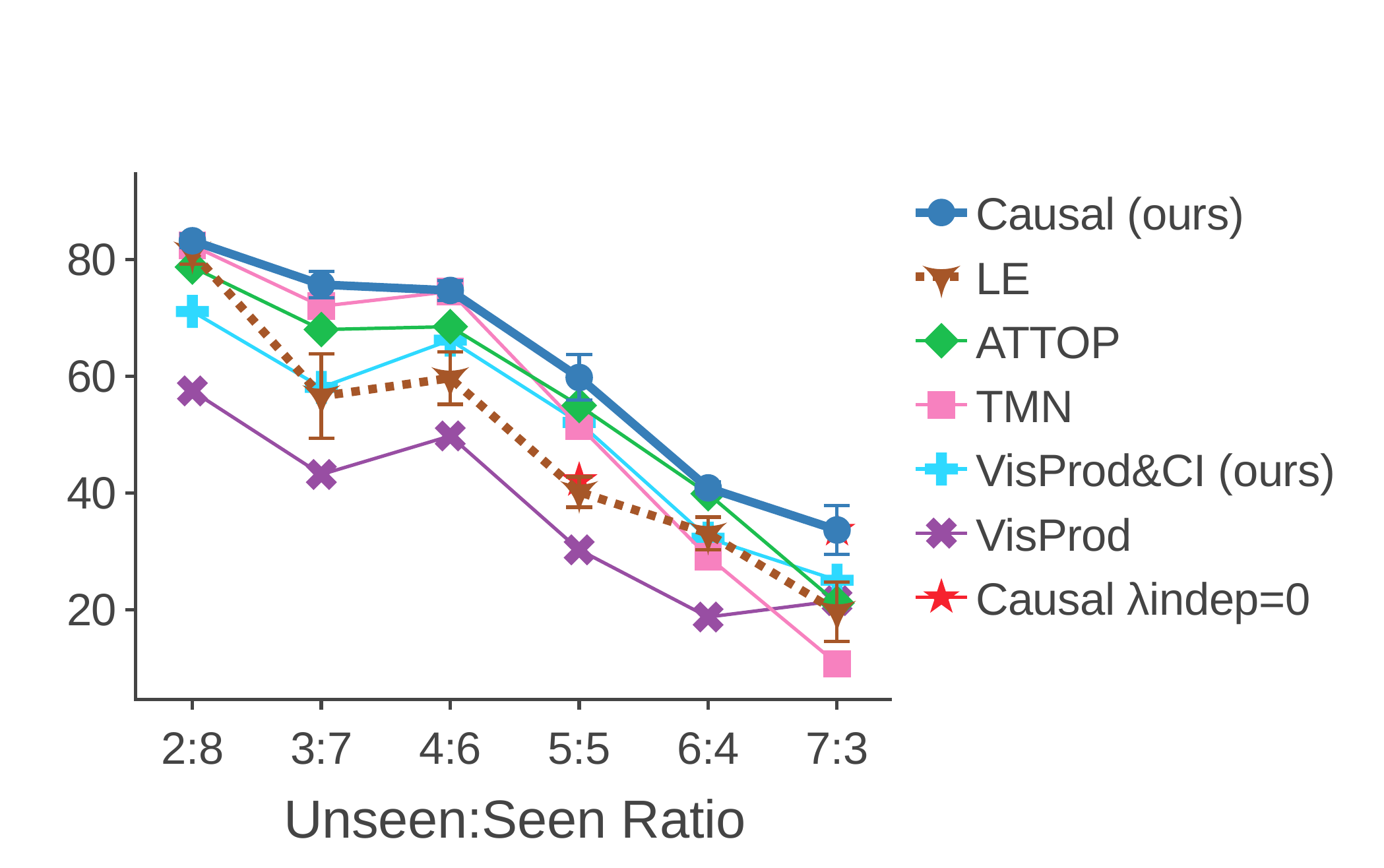}
    \end{overpic}
  \hspace{17pt}
\begin{overpic}[width=0.32\textwidth, trim={0cm 0cm 2.5cm 1.5cm},clip]{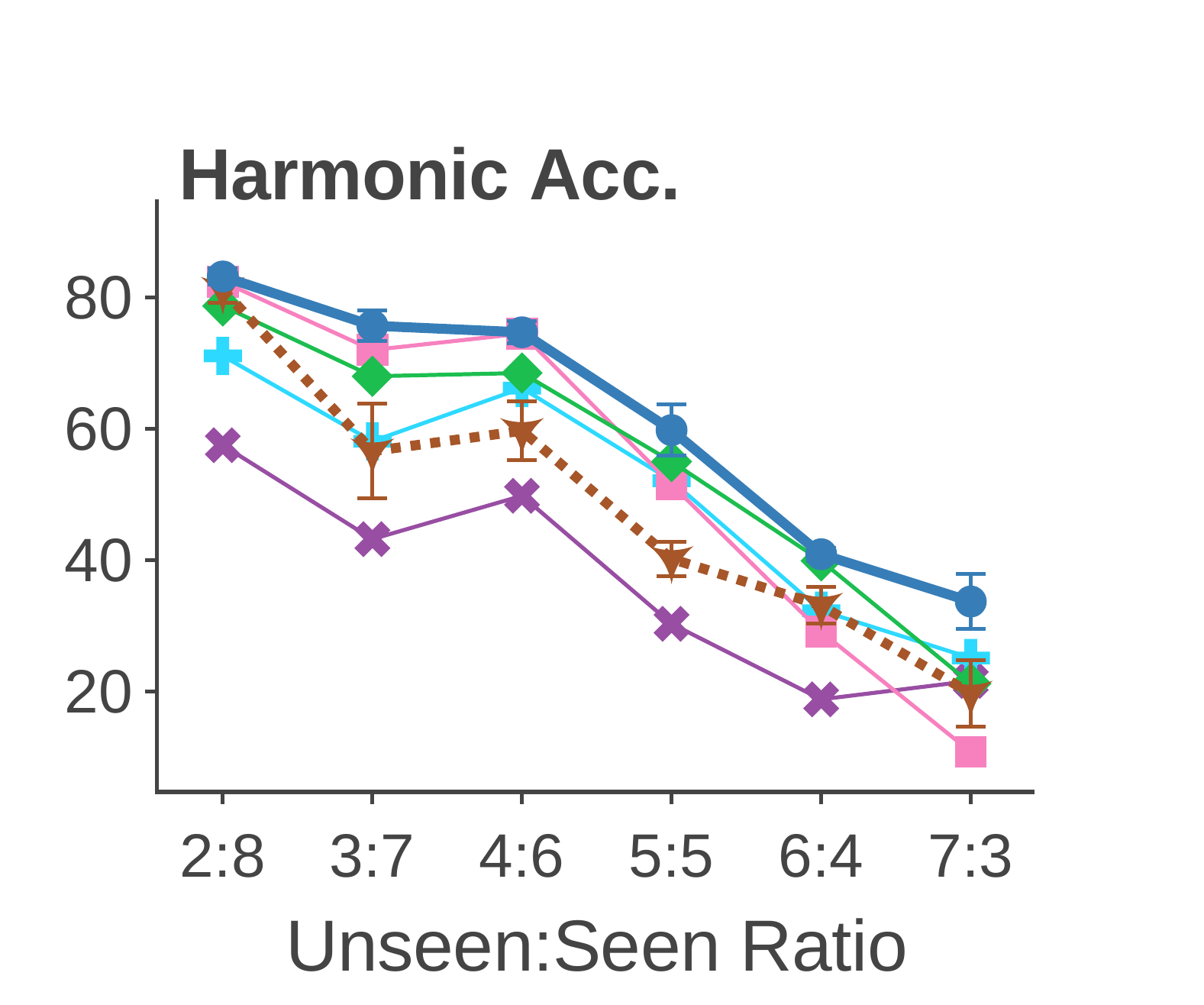}  %
 \end{overpic}

\caption{
\textbf{Left:} The seen-unseen plane for the 5:5 split. Modelling the core features largely improves the unseen accuracy: Compare \textit{Causal} to   \textit{LE} \edit{or to $\lambda_{indep}\!\!=\!\!0$ and compare} \textit{VisPros\&CI} to \textit{VisProd}. Error bars denote Standard Error of the Mean (S.E.M.) over 3 random splits \edit{and three random seeds.}
\textbf{Right:} Harmonic mean of seen-unseen accuracy for AO-CLEVR on a range of 20\% to 70\% unseen classes.  To reduce visual clutter, error bars are shown only for our \textit{Causal} method and for a vanilla baseline (\textit{LE}). }
    \label{fig_results_clevr}
\end{figure}

\textbf{AO-CLEVr:}
\figref{fig_results_clevr} (right) shows the Harmonic metric for AO-CLEVr for the whole range of unseen:seen ratios. 
Unsurprisingly, the more seen pairs are available, the better all models perform for unseen combinations of attributes and objects. Our approach ``Causal'', performs better than or equivalent to all the compared methods. 
VisProd easily confuses the Unseen classes. 
ATTOP, is better than LE on the open unseen pairs but performs substantially worse than all methods on the seen pairs.
TMN performs equally well as our approach for splits with mostly seen pairs but degrades when the fraction of seen pairs is below 4:6. 

\noindent \textbf{The seen-unseen plane:} 
By definition, our task aims to perform well in two different metrics (multi-objective): accuracy on seen pairs and unseen pairs. It is therefore natural to compare approaches by their performance on the seen-unseen plane. 
This is important, because different approaches may select different operating points to trade-off accuracy on unseen for accuracy on seen.  \figref{fig_results_clevr} (left) shows how the compared approaches trade-off the metrics for the 5:5 split.
ATTOP tends to favor unseen-pairs accuracy over the accuracy of seen pairs, vanilla models like VisProd, LE tend to favor seen classes. 
\textbf{Importantly}, it reveals that modelling the core-features largely improves the unseen accuracy, without hurting much the seen accuracy. Specifically, comparing \textit{Causal} to vanilla baseline \textit{LE}, improves the unseen acc. from 26\% to \edit{47\%} and reduces the seen acc. from 86\% to \edit{84\%}. Comparing \textit{VisPros\&CI} to \textit{VisProd}  improves the unseen acc. from 19\% to \edit{38\%} and reduces the seen Acc. from 85\% to \edit{82\%}.

\begin{SCtable}[][h]
    {
    \begin{small}\begin{sc}
    \scalebox{0.83}{
    \begin{tabular}{llllll}
    \toprule
    {} &          Unseen &          Seen &      Harmonic &          Closed &           AUSUC \\
    \midrule
    \multicolumn{5}{l}{{With prior embeddings}} \\
    LE               &  10.7 $\pm$ 0.8 &  52.9 $\pm$ 1.3 &  17.8 $\pm$ 1.1 &  55.1 $\pm$ 2.3 &  19.4 $\pm$ 0.3 \\
    ATTOP            &  22.6 $\pm$ 2.9 &  35.2 $\pm$ 2.7 &  26.5 $\pm$ 1.4 &  52.2 $\pm$ 1.8 &  20.3 $\pm$ 1.8 \\
    TMN              &   9.7 $\pm$ 0.6 &  51.9 $\pm$ 2.4 &  16.4 $\pm$ 1.0 &  \textbf{60.9 $\pm$ 1.1} &  \edit{\textbf{24.6 $\pm$ 0.8}} \\
    \midrule
    \multicolumn{5}{l}{{No prior embeddings}} \\
    LE*              &  15.6 $\pm$ 0.6 &  52.0 $\pm$ 1.0 &  24.0 $\pm$ 0.7 &  58.1 $\pm$ 1.2 &  22.0 $\pm$ 0.9 \\
    ATTOP*           &  16.5 $\pm$ 1.5 &  15.8 $\pm$ 1.9 &  15.8 $\pm$ 1.4 &  42.3 $\pm$ 1.5 &  16.7 $\pm$ 1.1 \\
    TMN*             &   6.3 $\pm$ 1.4 &  \textbf{55.3 $\pm$ 1.6} &  11.1 $\pm$ 2.3 & 58.4 $\pm$ 1.5 &  24.5 $\pm$ 0.8 \\
    \midrule
    \edit{Causal $\lambda_{indep}\!\!=\!\!0$} &  22.5 $\pm$ 2.0 &  45.5 $\pm$ 3.7 &  29.4 $\pm$ 1.5 &  55.3 $\pm$ 1.1 &  22.2 $\pm$ 0.9 \\
    \edit{Causal}         &  \textbf{26.6 $\pm$ 1.6} &  39.7 $\pm$ 2.2 &  \textbf{31.8 $\pm$ 1.7 } &  55.4 $\pm$ 0.8 &  23.3 $\pm$ 0.3 \\
    \bottomrule
    \end{tabular}
    }
    \end{sc}\end{small}
    }
    \hspace{-.4cm}
    \caption{Results for Zappos. $\pm$ denotes the Standard Error of the Mean (S.E.M.) over 5 random model initializations.}
    \label{tab:test}
\end{SCtable}

\textbf{Zappos:}
Our approach improves the Unseen and Harmonic metrics. For the ``Closed'' \edit{and ``AUSUC''} metrics it loses compared to TMN.%
We note that results on the Closed metric are less interesting from a causal point of view: A model cannot easily rely on the knowledge of which attribute-object pairs tend to appear in the test set.

\edit{For both AO-CLEVr and Zappos, the independence loss improves recognition on unseen pairs but hurts recognition of seen pairs. This is a known and important trade-off when learning models that are robust for interventions \cite{rothenhausler2018anchor}. The independence loss discourages certain types of correlations, hence models \textit{do not} benefit from them when the test and train distributions are identical (seen-pairs). However, the loss is constructed in such a way that these are exactly the correlations that fail to hold once the test distribution changes (unseen-pairs). Thus, ignoring these correlations improves performance for samples of unseen-pairs.} %

\vspace{-8pt}
\section{Discussion}
\vspace{-8pt}

\edit{We present a new causal perspective on the problem of recognizing new attribute-object combinations in images.}
We propose to view inference in this setup as answering the question ``which intervention on attribute and object caused the image''.
This perspective gives rise to a causal-inspired embedding model. The model learns \edit{disentangled} representations of attributes and objects although they are dependent in the training data. It provides better accuracy on two benchmark datasets.

\edit{The trade-off between seen accuracy and unseen accuracy reflects the fact that prior information about co-occurrence of attributes and objects in the training set is useful and predictive. A related problem has been studied in the setting of (non-compositional) generalized zero-shot learning \cite{COSMO}. We suggest that some of these techniques could be adapted to the compositional setting.}

\edit{Several aspects of the model can be further extended by relaxing its assumptions.}
First, the assumption that image features are normally distributed may be limiting, and alternative ways to model this distribution may improve the model accuracy. Second, \edit{the model is}  premised on the prior knowledge that the attributes and objects have distinct and stable generation processes. \edit{However, this prior knowledge may not always be available, or some attribute may not have an obvious physical meaning. E.g.  ``cute'', ``comfortable'' or } ``sporty'', and   in a multi-label setting \cite{patterson2016coco} it is challenging to reveal what are the independent generation mechanisms themselves from confounded training data.

This paper focused on the case where attributes and objects are fully disentangled. Clearly, in natural language, many attributes and object are used in a way are that makes them dependent. 
For example, white wine is actually yellow, and the attribute a small bear is larger than a large bird. It remains an interesting question to extend the fully disentangled case to learn specific dependencies while leveraging the power of disentangled representations.

\section*{Broader Impact}

Compositional generalization, the key challenge of this work, is critical for learning in real-world domains where the long tail of new combinations dominates the distribution, like in vision-and-language tasks or for the perception modules of autonomous driving.

A causality-based approach, like the one we propose, may allow vision systems to make more robust inference, and debias correlations that naturally exist in the training data, allowing to use vision systems in complex environments where the distribution of labels and their combinations is varying. It has been shown in the past that vision systems may emphasize biases in the data, and the ideas put forward in the paper may help make systems more robust to such biases.

Such approach may be useful for improving fairness in various applications, for example by providing a more balanced visual recognition of individuals from minority groups.

\section*{Acknowledgements}
\edit{We thank Daniel Greenfeld, Idan Schwartz, Eli Meirom, Haggai Maron, Lior Bracha and Ohad Amosy for their helpful feedback on the early version.}
\edit{Uri Shalit was partially supported by the Israel Science Foundation (grant No. 1950/19).}

\section*{Funding Disclosure}
\edit{Uri Shalit was partially supported by the Israel Science Foundation. Yuval Atzmon was supported by the Israel Science Foundation and Bar-Ilan University during his Ph.D. studies.}

{\small
\bibliographystyle{ieee_fullname}
\bibliography{compo}
}

{\huge{Supplementary Information:\\}}
{\huge{\edit{A causal view of compositional zero-shot recognition}}}

\appendix

\newcommand{\E}{\mathbb{E}}
\renewcommand\thefigure{S.\arabic{figure}}    
\renewcommand\thetable{S.\arabic{table}}   
\renewcommand\thesection{\Alph{section}}   
\renewcommand{\theequation}{S.\arabic{equation}}
\setcounter{figure}{0}  
\setcounter{table}{0}  
\setcounter{section}{0}

\section{Approximating \texorpdfstring{$\argmax_{a,o} p(\x|a,o)$}{}}
\label{sec:approximate_px_ao}
The conditional likelihood $p(\x|a,o)$ can be written by marginalizing over the latent factors $\lowphia$ and $\lowphio$
\begin{eqnarray}
    p(\x|a,o) \!\!\! &=& \!\!\! \int\!\!\!\int_{\lowphia, \lowphio} p(\x, \lowphia,\lowphio|a,o) = \int\!\!\!\int_{\lowphia, \lowphio} p(\x|\lowphia,\lowphio)p(\lowphia|a)p(\lowphio|o) d\lowphio d\lowphia \,.
    \label{eq:integral}
\end{eqnarray}
Computing this integral is hard in the general case. Variational autoencoders (VAEs) \cite{kingma2013auto} approximate a similar integral by learning a smaller support using an auxiliary probability function $Q$ over the latent space. 
Here we make another approximation, taking a "hard" approach and find the single most likely integrand. %
\begin{equation}
    \argmax_{(a,o) \in \A\times \O} \,\, p(\x|\lowphia,\lowphio)p(\lowphia|a)p(\lowphio|o)
    \label{eq:pointestimate}
\end{equation}

Based on \secref{sec:approach}, the three factors of the distributions are Gaussians. Therefore, maximizing \eqref{eq:pointestimate} is equivalent to minimizing the negative log likelihood \begin{equation}
    \argmin_{(a,o) \in \A\times \O} \,\, \frac{1}{\sigma^2_x}||\x - g(\lowphia, \lowphio)||^2 +
    \frac{1}{\sigma^2_a}||\lowphia - \ha||^2 +
    \frac{1}{\sigma^2_o}||\lowphio - \ho||^2 \,.
    \label{eq:nll_point_exact}
\end{equation}
This expression is composed of three components. The components allow to infer $(a,o)$ by evaluating distances in three representation spaces $\X$, $\phia$ and $\phio$ (\secref{sec:approach}).

However, we cannot apply \eqref{eq:nll_point_exact} to infer $(a,o)$ because the core features $\lowphia,\lowphio$ are latent.
Next, we introduce two additional approximations we use to apply \eqref{eq:nll_point_exact}. The first, approximates $||\x - g(\lowphia,\lowphio)||^2$ by $||\x - g(\ha,\ho)||^2$, using a Taylor expansion at the means $(\lowphia,\lowphio)=(\ha,\ho)$. The second, recovers (infers) the core features $\lowphia,\lowphio$ from the image, and substitute the recovered features in $||\lowphia - \ha||^2$, $||\lowphio - \ho||^2$.%

\subsection{Approximating \texorpdfstring{$||\x - g(\lowphia,\lowphio)||^2$}{}}
\label{sec:approx_g_phi}

A causal model can be equivalently represented using a``Structural Causal Model'' (SCM) \cite{pearl2000causality}.  An SCM matches a set of assignments to a causal graph. Each node in the graph is assigned a deterministic function of its parent nodes and an independent noise term. Specifically, based on the Gaussian assumptions in \secref{sec:approach}, the SCM of our causal graph (\figref{fig:causal_graph}a) is
\begin{eqnarray}
     \lowphia &=& n_a + \ha \label{SCM_phia} \\ 
     \lowphio &=& n_o + \ho \label{SCM_phio} \\ 
     \x &=& n_x + g(\lowphia, \lowphio), \label{SCM_x} \,,
\end{eqnarray} 
where $n_a, n_o, n_x$ are jointly independent Gaussian random variables 
$n_a\!\sim\! \mathcal{N}(0, \sigma^2_a I)$, $n_o\!\sim\! \mathcal{N}(0, \sigma^2_o I)$, $n_x\!\sim\! \mathcal{N}(0, \sigma^2_x I)$. %
$n_a, n_o, n_x$ represent sampling from the manifold of attributes, objects and images near their prototypes $\ha$ and $\ho$.

We use a zeroth-order Taylor expansion of $g(\lowphia, \lowphio)$ at $(\lowphia, \lowphio) = (\ha,\ho)$ and make the following approximation
\begin{equation}
    \big(\x - g(\lowphia, \lowphio)\big) \approx \big(\x - g(\ha,\ho)\big) \quad.
    \label{eq:taylor_distance}
\end{equation}

Next, we discuss the first-order \textit{approximation error}. For brevity,  we denote with $\phiao$ the concatenation of the elements in $(\lowphia, \lowphio)$ into a single vector. Similarly $(\ha,\ho)$ into $\hao$ and $(n_a, n_o)$ into $\nao$.

We approximate $g(\phiao)$ by a first-order Taylor expansion  at $\phiao=\hao$ to
\begin{equation}
    g(\phiao)  \approx g(\hao) + [Jg](\hao)\cdot(\phiao-\hao) = g(\hao) + [Jg](\hao)\cdot\nao,
    \label{eq:taylor}
\end{equation}
where $[Jg]$ is the Jacobian of $g$ and the last equality stems from the SCM (Eqs. \ref{SCM_phia}, \ref{SCM_phio}).

Using \eqref{eq:taylor} and the Cauchy-Schwarz inequality, the first-order squared error approximation of \eqref{eq:taylor_distance} is:
\begin{eqnarray}
    \E||\big(\x - g(\phiao)\big) - \big(\x - g(\hao)\big)||^2 = \E||g(\phiao) - g(\hao)||^2  \nonumber\\ 
     \approx \E||[Jg](\hao)\nao||^2 \leq  ||[Jg](\hao)||_F^2 \E ||\nao||^2     
\end{eqnarray}

This implies that the error of the approximation \eqref{eq:taylor_distance} is mainly dominated by the gradients of $g$ at $\hao$, and the variance of $\nao$. If the gradients and the variance of $\nao$ are too large, then this approximation may be too coarse, and one may resort to more complex models like variational methods \cite{kingma2013auto}. Empirically, we observe that this approximation is useful.

\subsection{Recovering the core features}

To apply \eqref{eq:nll_point_exact}, we approximate $\lowphia,\lowphio$, by reconstructing (inferring) them from the image.

The main assumption we make is that images preserve the information about the core features they were generated from, at least to a level that allows us to differentiate what were the semantic prototypes ($\ha$ and $\ho$) of the core feature. Otherwise, inference on test data and labeling the training data by human raters will render to random guessing.

Therefore, we assume that there exists an inverse mappings that can infer $\lowphia$ and $\lowphio$ from the image up to a reasonably small error:
\begin{eqnarray}
\hatphia \equiv g^{-1}_A(\x) = \lowphia + \epsilon_A(x) \\
\hatphio \equiv g^{-1}_O(\x) = \lowphio + \epsilon_O(x)\,,
\end{eqnarray}
where $\epsilon_A(x), \epsilon_O(x)$ denote the image-based error of inferring the attribute and object core features.

Specifically, we assume that the error in substituting $\lowphia$ by $\hatphia$ and $\lowphio$ by $\hatphio$ in \eqref{eq:nll_point_exact} is small enough to keep them close to their prototypes ($\ha$ and $\ho$).
Namely, we make the following approximation
\begin{eqnarray}
    ||\lowphia - \ha||^2 &=& ||\hatphia - \epsilon_A(x) - \ha||^2 \leq ||\hatphia  - \ha||^2 + ||\epsilon_A(x)||^2 \approx ||\hatphia - \ha||^2 \nonumber \\
    ||\lowphio - \ho||^2 &=& ||\hatphio - \epsilon_O(x) - \ho||^2 \leq  ||\hatphio - \ho||^2 + ||\epsilon_O(x)||^2 \approx ||\hatphio - \ho||^2 \,.
    \label{eq:approx_core_distance}
\end{eqnarray}

To conclude, we use \eqref{eq:taylor_distance} and  \eqref{eq:approx_core_distance} to approximate  \eqref{eq:nll_point_exact} by :
\begin{equation*}
    \argmin_{(a,o) \in \A\times \O} \,\, \frac{1}{\sigma^2_x}||\x - g(\ha,\ho)||^2 +
    \frac{1}{\sigma^2_a}||\hatphia - \ha||^2 +
    \frac{1}{\sigma^2_o}||\hatphio - \ho||^2,
\end{equation*}
which is the expression for \eqref{eq_inference_NLL} in the main paper.

\section{Independence Loss}
\label{sec:indep_loss}

Our loss includes a component $\L_{indep}$, which is designed to capture the conditional-independence relations that the causal graph dictates (\eqref{eq:indepCondAO}). We now describe $\L_{indep}$ in detail.

Since we do not have the actual values of the latent core features $\lowphia$, $\lowphio$, we wish that their reconstructions $\hatphia$ and $\hatphio$ maintain approximately the same independence relations as \eqref{eq:indepCondAO}. 
To learn mappings that adhere to these statistical independences over $\hatphia$ and $\hatphio$, we regularize the learned mappings using a differentiable measure of statistical dependence.

Specifically, we use a positive differentiable measure of the statistical dependence, denoted by $\I$. For two variables $(u, v)$, conditioned on a categorical variable $Y$, we denote by $\I(u, v|Y)$ the positive differentiable measure of the statistical conditional dependence of $(u, v|Y)$.
For example we encourage approaching the equality in Eq. \ref{eq:indepCondAO}b by minimizing $\I(\hatphia, \hatphio|A)$.
$\I$ is based on measuring the Hilbert-Schmidt Information Criterion (HSIC)  \cite{gretton2005measuring,gretton2008kernel}, which is a non-parametric method for estimating the statistical dependence between samples of two variables.
We adapt the HSIC criterion to measure \textit{conditional} dependencies. $\I$   penalizes conditional dependencies in a batch of samples $B=\big\{(u_i, v_i, y_i)\big\}_{i=1}^{|B|}$, by summing over groups of samples that have the same label:

\begin{eqnarray}
    \I(u, v|Y) = \frac{1}{|Y|} \sum_{y \in Y} { \text{HSIC}(U|Y=y, V|Y=y) }  \\
    \text{where } \,\,\, (U|Y=y, V|Y=y) = \big\{ (u_i, v_i) \in B \mid y_i = y \big\}. \nonumber
\end{eqnarray}

Finally, we have $\L_{indep}$, a loss term that encourages the four conditional independence relations of \ref{eq:indepCondAO}:
\begin{eqnarray}
    \L_{indep} &=& \L_{oh} + \lambda_{rep} \L_{rep}
    \label{eq:Lindep}\\
    \L_{oh} &=& \I\Big(\hatphia, O |A\Big) + \I\Big( \hatphio, A |O\Big)  \label{eq:Loh} \\ 
    \L_{rep} &=& \I\Big(\hatphia, \hatphio |A\Big) + \I\Big(\hatphia, \hatphio |O\Big) \label{eq:Lrep},
\end{eqnarray}
where $\lambda_{rep}$ is a hyper parameter. %

Minimizing \eqref{eq:Loh} encourages the representation of the inferred attribute $\hatphia$, to be invariant to the categorical (``one-hot'') representation of an object $O$. Minimizing \eqref{eq:Lrep} encourages $\hatphia$ to be invariant to the ``appearance'' of an object ($\hatphio$).

\subsection{An expression for Hilbert-Schmidt Information Criterion (HSIC) with linear kernel}
\edit{The linear-kernel HSIC between two batches of vectors $\vec{U}, \vec{V}$ is calculated in the following manner \cite{gretton2008kernel}: It uses two linear kernel matrices 
$K_{i,j} = \vec{u}_{[i, :]}\vec{u}_{[j, :]}^{T}$, $L_{i,j} = \vec{v}_{[i, :]}\vec{v}^{T}_{[j, :]}$. Then the HSIC is calculated as the scaled Hilbert-Schmidt norm of their cross-covariance matrix:}
\begin{equation*}
    HSIC(U, V) = \frac{1}{(n-1)^2}  \cdot \textbf{tr}(KHLH),
\end{equation*}
where $H_{ij}=\delta_{i,j}-\frac{1}{n}$ is a centering matrix \edit{, and $n$ is the batch size} 

\subsection{\edit{Minimizing \texorpdfstring{$\L_{indep}$}{} encourages robustness of the reconstructed core features  $\hatphia, \hatphio$.}}
\label{sec:PIDA}

The conditional-independence term of $\L_{oh}$ within $\L_{indep}$ is related to a metric named ``Post Interventional Disagreement'' (PIDA), recently introduced by \cite{PIDA}. PIDA measures disentanglement of representations for models that are trained from unsupervised data. This  section explains  their relation in more detail, \edit{by showing that minimizing $\L_{indep}$ encourages the following properties: %
$p^{do(O=o)}(\hatphio)\!\!\approx\!\! p^{do(A=a,O=o)}(\hatphio)$ and  $p^{do(A=a)}(\hatphia))\!\!\approx\!\! p^{do(A=a,O=o)}(\hatphia)$.
}

The PIDA metric for attributes is measured by  
\begin{equation}
    PIDA(a'|a, o) := d\big( \mathbb{E}^{do(a)}[\hatphiatag], \mathbb{E}^{do(a, o)}[\hatphiatag] \big),
    \label{eq:PIDA_A}
\end{equation}
where $d$ is loosely described as ``a suitable'' positive distance function (like the $L_2$ distance). $PIDA(a'|a, o)$ quantifies the shifts in the inferred features $\hatphiatag$ when the object is enforced to $o$. Similarly, the PIDA term for objects is $PIDA(o'| o, a)$.

Below we show that encouraging the conditional independence $(\hatphia \indep O | A)$, is equivalent to minimizing $d\big( p^{do(a)}(\hatphiatag), p^{do(a, o)}(\hatphiatag) \big)$ and therefore minimizes $PIDA(a'|a, o)$.

First, in our causal graph (\figref{fig:causal_graph}a) a do-intervention on both $a$ and $o$ is equivalent to conditioning on $(a,o)$ 
\begin{equation}
    p^{do(a, o)}(\hatphiatag) = p(\hatphiatag|a,o).
\end{equation}
Second, minimizing \eqref{eq:Loh} encourages the conditional independence $(\hatphiatag \indep O|A)$, which is equivalent to minimization of $d\big(p(\hatphiatag|a), p(\hatphiatag|a, o)\big)$ \footnote{When $(\hatphiatag \indep O|A)$, then by definition  $p(\hatphiatag|a, o) = p(\hatphiatag|a)$. Therefore encouraging $(\hatphiatag \indep O|A)$, makes $p(\hatphiatag|a, o)$ approach $p(\hatphiatag|a)$}.
Third, when $d\big(p(\hatphia|a), p(\hatphia|a, o)\big)$ approaches zero,  then $p^{do(a)}(\hatphia)$ approaches $p(\hatphia|a)$. 
This stems from the adjustment formula, $p^{do(a)}(\hatphiatag) = \sum_{o} p(\hatphiatag|a,o)p(o)$:
When $d\big(p(\hatphiatag|a), p(\hatphiatag|a, o)\big)$ approaches zero, then $\sum_{o} p(\hatphiatag|a,o)p(o)$ approaches $\sum_{o} p(\hatphiatag|a)p(o) = p(\hatphiatag|a)$. Therefore $p^{do(a)}(\hatphiatag)$ approaches $p(\hatphiatag|a)$.

As a result, we have the following: Minimizing \eqref{eq:Loh} leads to $p^{do(a, o)}(\hatphiatag)$ approaching $p(\hatphiatag|a)$, which as we have just shown, leads to $p(\hatphiatag|a)$ approaching $p^{do(a)}(\hatphiatag)$. Therefore, $d\big( p^{do(a)}(\hatphiatag), p^{do(a, o)}(\hatphiatag) \big)$ is minimized and accordingly, $PIDA(a'|a, o)$ (Eq. \ref{eq:PIDA_A}) is minimized.

Similarly, for objects, encouraging the conditional independence $(\hatphiotag \indep A | O)$ minimizes $PIDA(o'|a, o)$. Therefore, minimizing $\L_{oh}$ (Eq. \ref{eq:Loh}), optimizes both $PIDA(a'|a, o)$ and $PIDA(o'|a, o)$.

\section{Implementation details}

\subsection{Architecture}
Similar to LE \cite{ATTOP}, we implemented $g$, $\ha$ and $\ho$ by MLPs with ReLU activation. For $g^{-1}_A$ and $g^{-1}_O$, every layer used a batch-norm and leaky-relu activation.  All the MLPs share the same size of hidden layer, denoted by $d_h$. 

For experiments on Zappos, we also learned a single layer network $f$ to project pretrained image features to the feature space $\X$. This strategy was inspired by the baseline models LE and ATTOP. Learning a projection $f$ finds better solutions on the validation set than using the pretrained features as $\X$.

For HSIC we used the implementation of \cite{greenfeld2019robust} and applied a linear kernel as it does not require tuning additional hyper-parameters.
 
\subsection{Optimization}
\textbf{AO-CLEVr}:
We optimized AO-CLEVr in an alternating fashion: First we trained $\ha$, $g^{-1}_A$ and $g$, keeping $\ho$, $g^{-1}_O$ frozen. Then we froze $\ha$, $g^{-1}_A$ and trained $\ho$, $g^{-1}_O$ and $g$. This optimization strategy allows to stabilize the attribute representation when minimizing \eqref{eq:Lrep}. The strategy was developed during the early experiments with a low-dimensional ($\X \subset \reals^2$) synthetic dataset. In the ablation study (Section \ref{sec:ablation} below), we show that a standard (non-alternating) optimization strategy achieves comparable results, \edit{but with somewhat higher bias toward seen accuracy}.

We used SGD with Nesterov Momentum to train with AO-CLEVr. Empirically, we found that SGD allowed finer control over $\L_{indep}$ than Adam \cite{adam}.

In practice, we weighed the loss of $||\hatphia-\ha||^2$ and $||\hatphio-\ho||^2$ according to the respective attribute and object frequencies in the training set. This detail has a relatively small effect on performance. Without weighing the loss, the Harmonic decreases by \edit{1.2\%} (from \edit{68.8\% to 67.6\%}) and Unseen accuracy decreases by \edit{1.3\%} \edit{(from 57.5\% to 56.2\%)}.

\textbf{Zappos}:
We optimized Zappos in a standard (non-alternating) fashion. We couldn’t use the \textit{alternating} optimization strategy, because in Zappos we also learn a mapping $f$ that projects pretrained image features to the feature space $\X$. Thus, updating the parameters of $f$ changes the mapping to $\X$ and we cannot keep $\ha$, $g^{-1}_A$ frozen once $\X$ changes.

As in \cite{ATTOP}, we used Adam \cite{adam} to train with Zappos.

\subsection{Early Stopping and Hyper-parameter selection}
We trained each model instantiation for a maximum of 1000 epochs and early stopped on the validation set.

We used two metrics for early-stopping and hyper-parameter selection on the validation set: (i) the Harmonic metric  for testing the unseen-accuracy, the seen-accuracy and the Harmonic; and (ii) accuracy of the Closed metric for testing the Closed accuracy. In Zappos, we followed \cite{TMN2019} and used the AUSUC for testing both the AUSUC metric and the Closed accuracy.

For our approach and all the compared methods, we tuned the hyper-params by first taking a coarse random search, and then further searching around the best performing values on the validation set. As a rule of thumb, we first stabilized the hyper-parameters of the learning-rate, weight-decay, and architecture. Then we searched in finer detail over the hyper-parameters of the loss functions. At the most fine-grained iteration of the random search, each combination of hyper-parameters was evaluated with 3 different random weight initializations, and metrics were averaged over 3 runs. We chose the set of hyper-parameters that maximized the average metric of interest.

Since AO-CLEVr has $6\cdot3 = 18$ different splits, we searched the hyper-parameters over a single split of each of the ratios \{2:8, 5:5, 6:4, 7:3\}. For \{3:7, 4:6\} ratios, we used the hyper-parameters chosen for the \{5:5\} split.

\subsection{Hyper-parameters for loss function}
\edit{
In practice, to weigh the terms of the loss function, we use an equivalent but different formulation. Specifically, we set $\lambda_{indep}\!\!=\!\!1$ and $\lambda_{invert}\!\!=\!\!1$, and use the following expressions for $\L_{indep}$ and $\L_{invert}$
\begin{eqnarray}
    \L_{indep} = \lambda_{oh} \L_{oh} + \lambda_{rep} \L_{rep} \\
    \L_{invert} = \lambda_{icore}\big(CE(a, f_a(\ha)) + CE(o, f_o(\ho))\big) + \\ \lambda_{ig}\big(CE(a,f_{ga}(g(\ha,\ho))) + CE(a,f_{go}(g(\ha,\ho)))\big).    \nonumber
\end{eqnarray}
where $\lambda_{oh}, \lambda_{rep}, \lambda_{icore}$ and $\lambda_{ig}$ weigh the respective loss terms.
}

\subsection{Grid-search ranges}
For \textit{Causal} with AO-CLEVr, we started the random grid-search over the following ranges: \textbf{Architecture:} (1) Number of layers for $\ha$ and $\ho$ $\in \{0,1,2\}$, (2) Number of layers for $g^{-1}_A$ and $g^{-1}_O$ $\in \{1,2,3\}$, (3) Number of layers for $g$ $\in \{2,4\}$, (4) Common size of hidden layers $d_h \in \{10,30,150,300, 1000\}$\footnote{We used 150 and not 100, in order to have total size of $2\cdot150=300$ for the concatenated representation of $[\hatphia, \hatphio]$. 300 is comparable to the default value for emb-dim in the LE baseline.}.
\textbf{Optimization:} (1) learning rate $\in \{1\text{e-}5,3\text{e-}5,1\text{e-}4,3\text{e-}4,1\text{e-}3\}$, when using alternate training, we used different learning rates for each alternation.  (2) weight-decay $\in \{1\text{e-}5,1\text{e-}4,1\text{e-}3,0.01,0.1,1\}$ (3) $\lambda_{rep} \in \{0, 0.001, 0.003, 0.01, 0.03, 0.1, 0.3, 1, 3, 10, 30, 100, \edit{300}\}$ (4) $\lambda_{oh} \in \{0, 0.001, 0.003, 0.01, 0.03, 0.1, 0.3, 1, 3, 10, 30, 100, \edit{300}\}$ (5)  $\lambda_{ao} \in \{0, 0.1, 0.3, 1, 3, 10\}$ (6) $\lambda_{icore} \in \{0, 0.01, 0.03, 0.1, 0.3, 1, 3, 10, 30, 100\}$ (7) $\lambda_{ig} \in \{0, 0.01, 0.03, 0.1, 0.3, 1, 3, 10, 30, 100\}$. We didn't tune batch-size, we set it to 2048. 

\edit{For \textit{Causal} with Zappos, we used some hyper-parameters found with Causal\&AO-CLEVr and some already known for LE with Zappos. Specifically, for the architecture, we set (1) Number of hidden layers for $\ha$ and $\ho$ = 0 (linear embedding), (2) Number of layers for $g^{-1}_A$ and $g^{-1}_O$ = 2, (3) Number of layers for $g$ = 2 (4) Common size of hidden layers $d_h = 300$. For the optimization, in order to find a solution around the solution used by LE, we selected $\lambda_{ao}=1000, \lambda_{ig}=0, \text{weight-decay} = 5\text{e-}5, \text{batch-size}=2048$; and $\lambda_{icore}=100$ as with Causal\&AO-CLEVr. We applied the random search protocol over $\lambda_{rep} \in 15\cdot\{0, 0.001, 0.003, 0.01, 0.03, 0.1, 0.3, 1, 3, 10, 30, \}$,  $\lambda_{oh} \in 15\cdot\{0, 0.001, 0.003, 0.01, 0.03, 0.1, 0.3, 1, 3, 10, 30, \}$, and learning rate $\in \{1\text{e-}4,3\text{e-}4\}$
}

For \textit{TMN}, we applied the random grid-search according to the ranges defined in the supplemental of \cite{TMN2019}. Specifically: $\text{lr} \in \{0.0001, 0.001, 0.01, 0.1\}$, $\text{lrg} \in \{0.0001, 0.001, 0.01, 0.1\}$, $\text{batch-size} \in \{64, 128, 256, 512\}$, $\text{concept-dropout} \in \{0, 0.05, 0.1, 0.2\}$, $\text{nmod} \in \{12, 18, 24, 30\}$, $\text{output-dimension} \in \{8, 16\}$, $\text{number-of-layers} \in \{1, 2, 3, 5\}$. Additionally, we trained TMN for a maximum of 30 epochs, which is $\times 6$ longer than the recommended length (4-5 epochs). As instructed, we chose the number of negatives to be ``all negatives''. With Zappos, our grid-search found a hyper-parameters' combination with better performance than the one reported by the authors.

For \textit{ATTOP} with AO-CLEVr, we used the following ranges: 
$\lambda_{aux} \in \{0, 0.001, 0.003, 0.01, 0.03, 0.1, 0.3, 1, 3, 10, 30, 100\}$, \\ $\lambda_{comm} \in \{0, 0.001, 0.003, 0.01, 0.03, 0.1, 0.3, 1, 3, 10, 30, 100\}$, \\ $\text{emb-dim} \in \{10, 30, 150, 300, 1000\}$, $\text{weight-decay} \in \{1\text{e-}5, 1\text{e-}4, 1\text{e-}3, 1\text{e-}2\}$. We used the default learning rate $1\text{e-}4$, and used a batch size of 2048. 

For \textit{ATTOP} with Zappos, we used the hyper-parameters combinations recommended by \cite{ATTOP, TMN2019}, and also searched for $\text{emb-dim} \in \{300, 1000\}$

For \textit{LE},  we used the following ranges: $\text{weight-decay} \in \{1\text{e-}4, 1\text{e-}3, 1\text{e-}2\}$, $\text{emb-dim} \in \{10, 30, 150, 300, 1000\}$. With Zappos we also followed the guideline of \cite{TMN2019} and searched $\text{lr} \in \{0.0001, 0.001\}$.

Note that for ATTOP and LE, when Glove embedding is enabled then $\text{emb-dim}$ is fixed to 300.

For \textit{VisProd} with AO-CLEVr, we used the following ranges: We used the same number of layers, weight-decay and learning rates as used to train ``Causal''. We also followed the alternate-training protocol. With Zappos, we searched for $\text{emb-dim} \in \{100, 300, 1000\}$, $\text{lr} \in \{1\text{e-}4, 1\text{e-}3\}$.

\subsection{AO-CLEVr dataset}
\label{supp:clevr}
\paragraph{Pretrained features:} Similar to Zappos, we extracted pretrained features for AO-CLEVr using a pretrained ResNet18 CNN.

\paragraph{Cross validation splits:}
For cross-validation, we used two types of splits. The first uses the same unseen pairs for validation and test. We call this split the ``overlapping'' split. The split allows us to quantify the potential generalization capability of each method. The second split, is harder, where unseen validation pairs are not overlapping with the unseen test pairs. We call this split the ``\textit{non}-overlapping'' split.

For the overlapping split, we varied the ratio of unseen:seen pairs on a range of (2:8, 3:7, \dots 7:3), and for each ratio we drew 3 random seen-unseen splits. For the non-overlapping split, we varied the ratio of unseen:seen pairs on a range of (2:6, 3:5, \dots 5:3), and for each ratio we drew 3 random seen-unseen splits. In addition, we always draw 20\% of the pairs for validation.

\section{AO-CLEVr with a non-overlapping validation set.}

Here we present results for AO-CLEVr with the ``\textit{non}-overlapping'' split. For this split, the unseen validation pairs are not overlapping with the unseen test pairs. It is harder than the ``overlapping'' split, which uses the same unseen pairs for validation and test.

The non-overlapping split is important because, in practice, we cannot rely on having labeled samples of the unseen pairs for validation purposes.

\begin{figure}[h]
  \centering   
  \hspace{-15pt}
  \includegraphics[width=0.11\textwidth, trim={15.cm 0cm 0.5cm 1.5cm},clip]{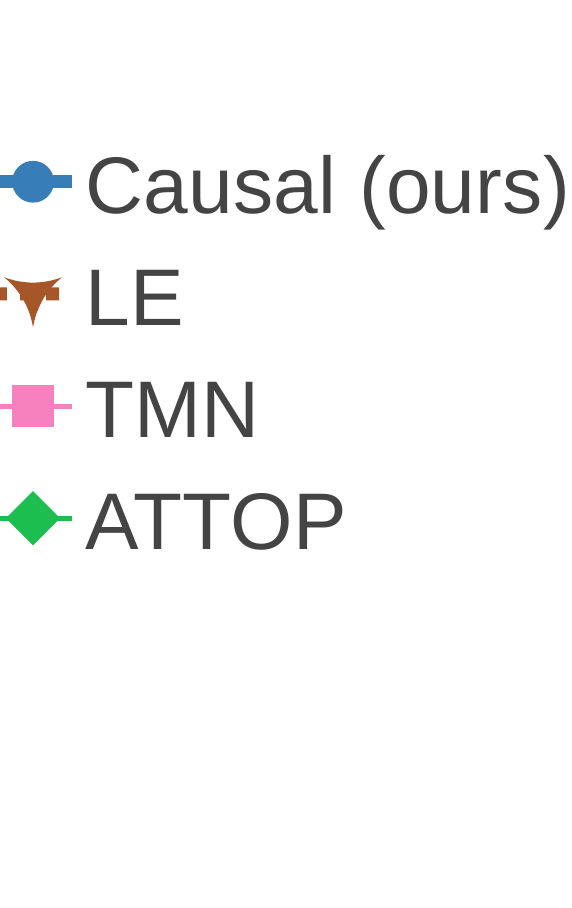}
  \includegraphics[width=\mywidth\textwidth, trim={0cm 0cm 2.5cm 1.3cm},clip ]{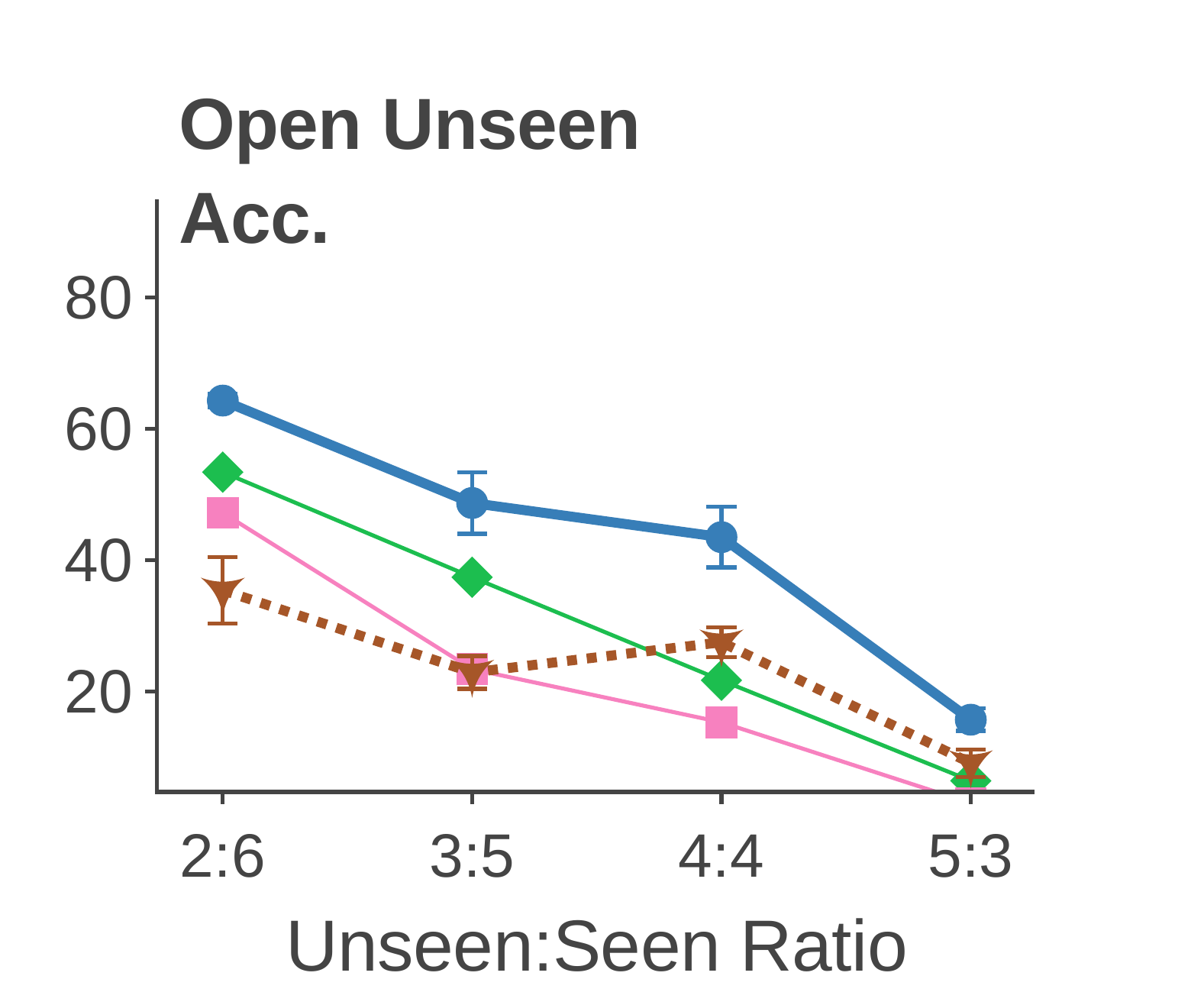}   
  \includegraphics[width=\mywidth\textwidth, trim={0cm 0cm 2.5cm 1.5cm},clip]{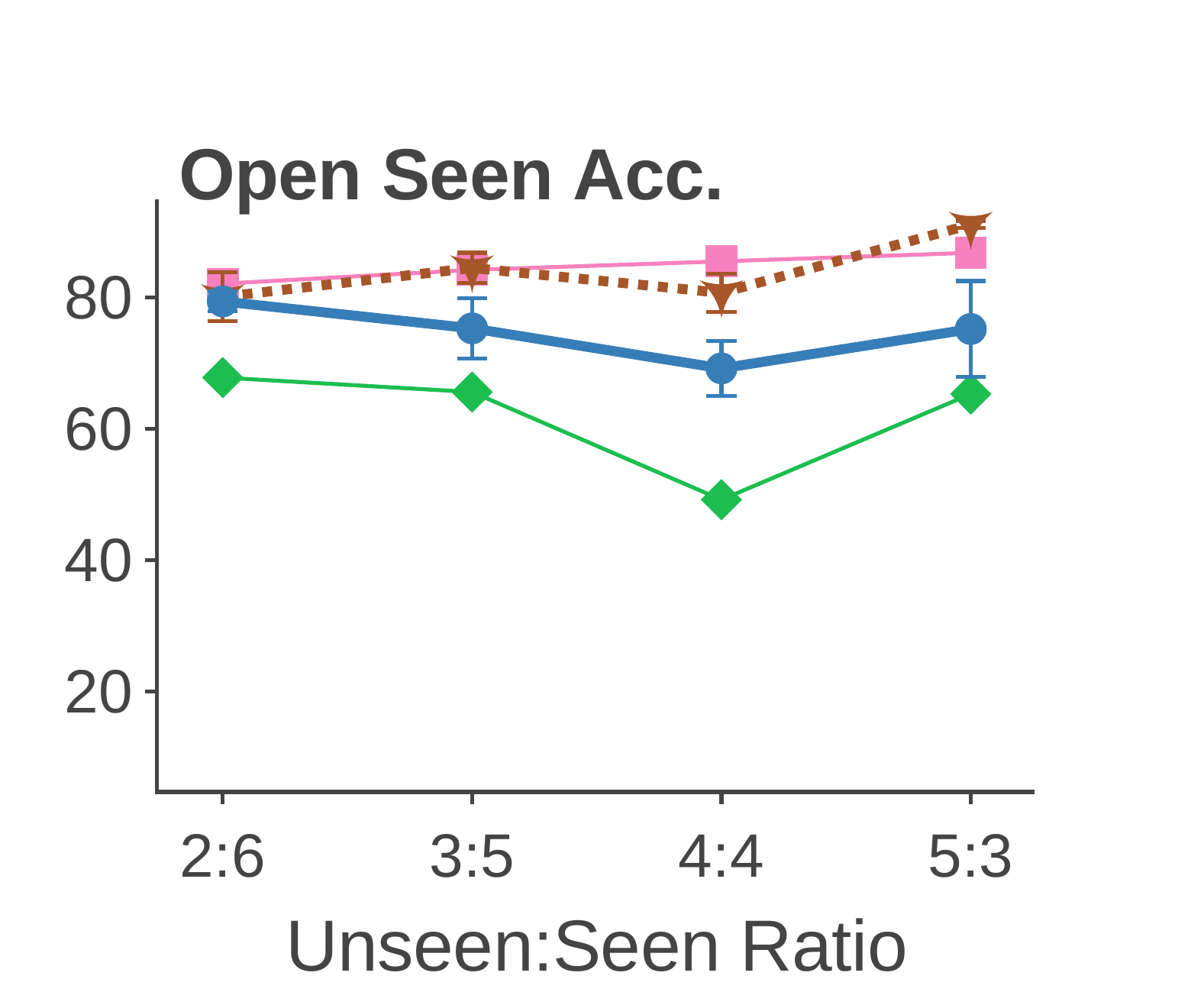} 
  \includegraphics[width=\mywidth\textwidth, trim={0cm 0cm 2.5cm 1.5cm},clip]{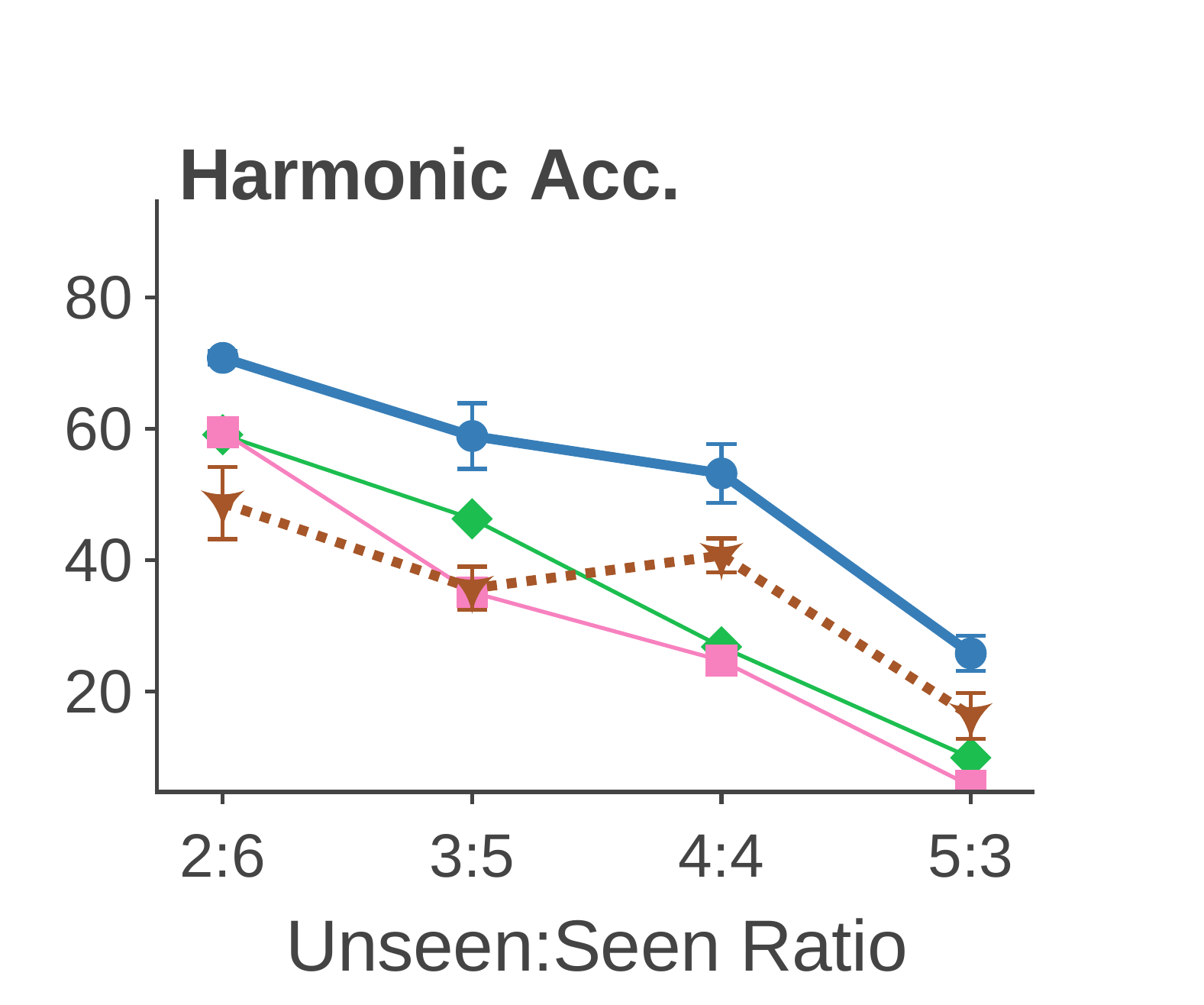} 
  \includegraphics[width=\mywidth\textwidth, trim={0cm 0cm 2.5cm 1.5cm},clip]{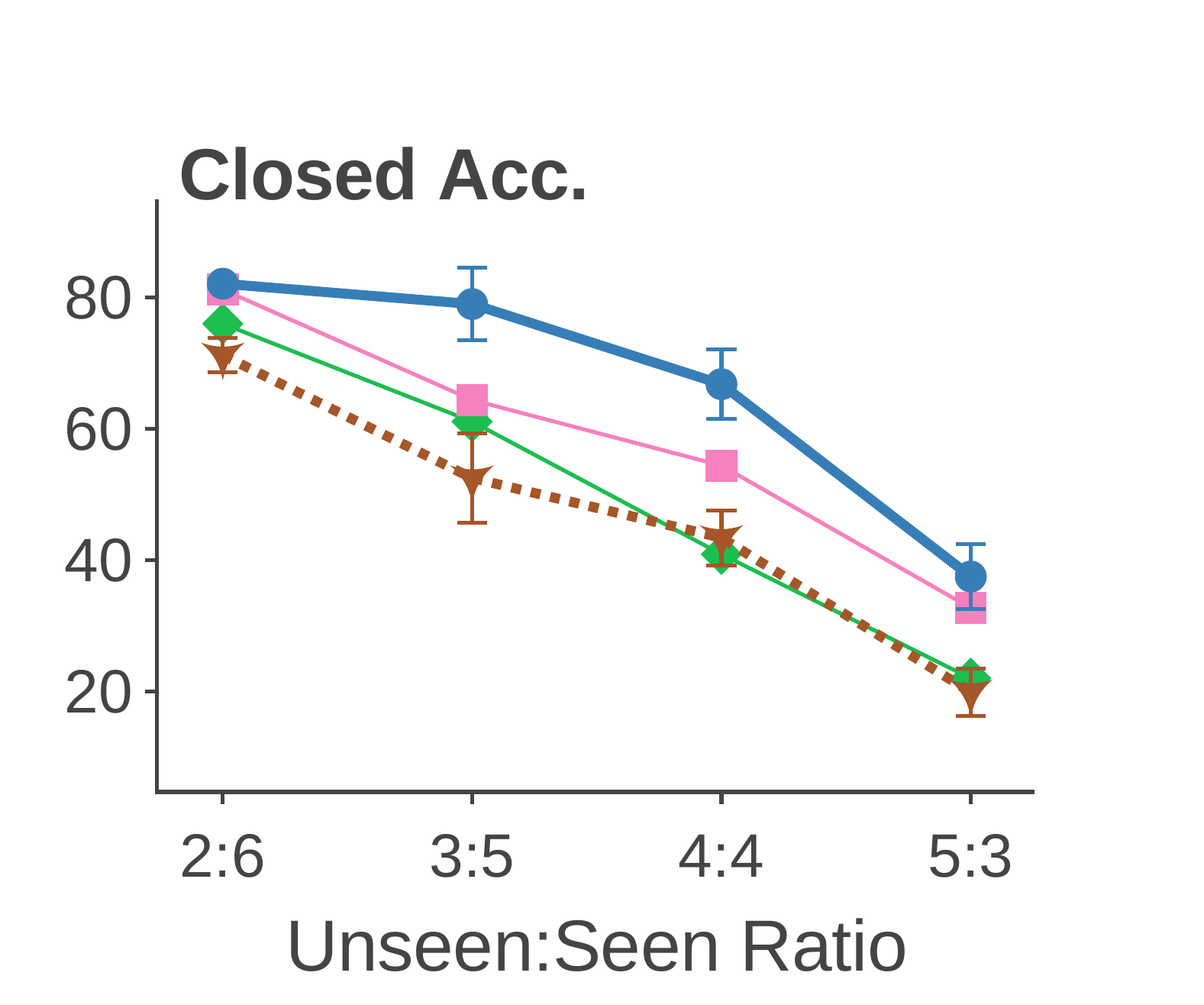} 

\caption{AO clevr with a \textit{non}-overlaping validation set.}
    \label{fig_results_clevr_UV}
\end{figure}

\figref{fig_results_clevr_UV} shows the measured metrics when comparing ``Causal'', LE, TMN, and ATTOP and varying the ratio of seen:unseen pairs between 2:6 to 5:3.

 For the main zero-shot metrics (Open-Unseen, Harmonic and Closed), our approach ``Causal'', performs substantially better than the compared methods.
ATTOP performs substantially worse on ``seen" pairs.

\section{Complete results for AO clevr with overlapping split}

\begin{figure}[h]
  \centering   
  \hspace{-15pt}
  \includegraphics[width=0.14\textwidth, trim={14.cm 2cm 0.5cm 1.5cm},clip]{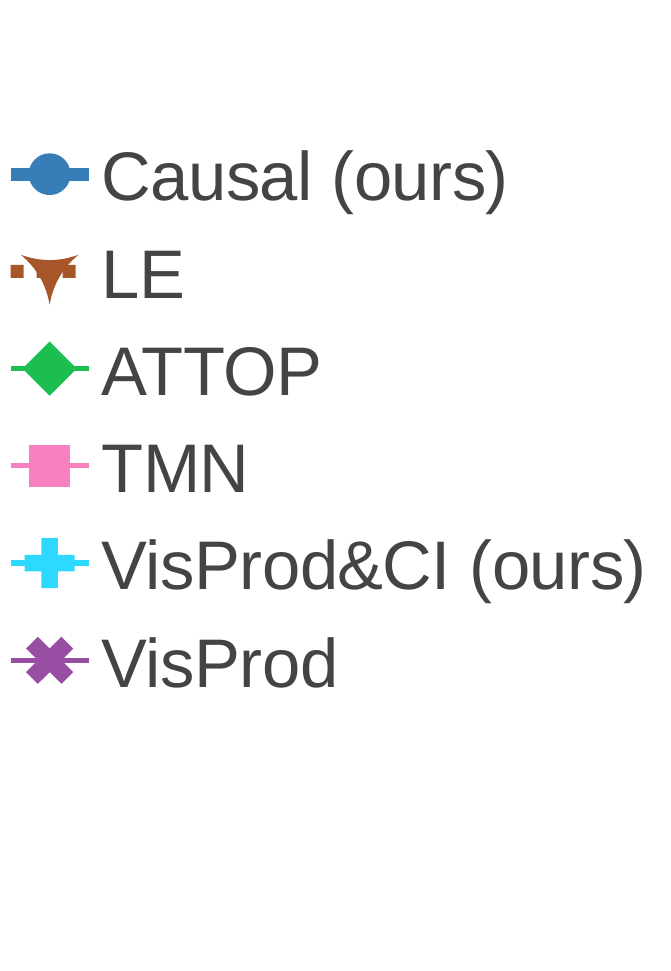}
  \includegraphics[width=\mywidth\textwidth, trim={0cm 0cm 2.5cm 1.3cm},clip ]{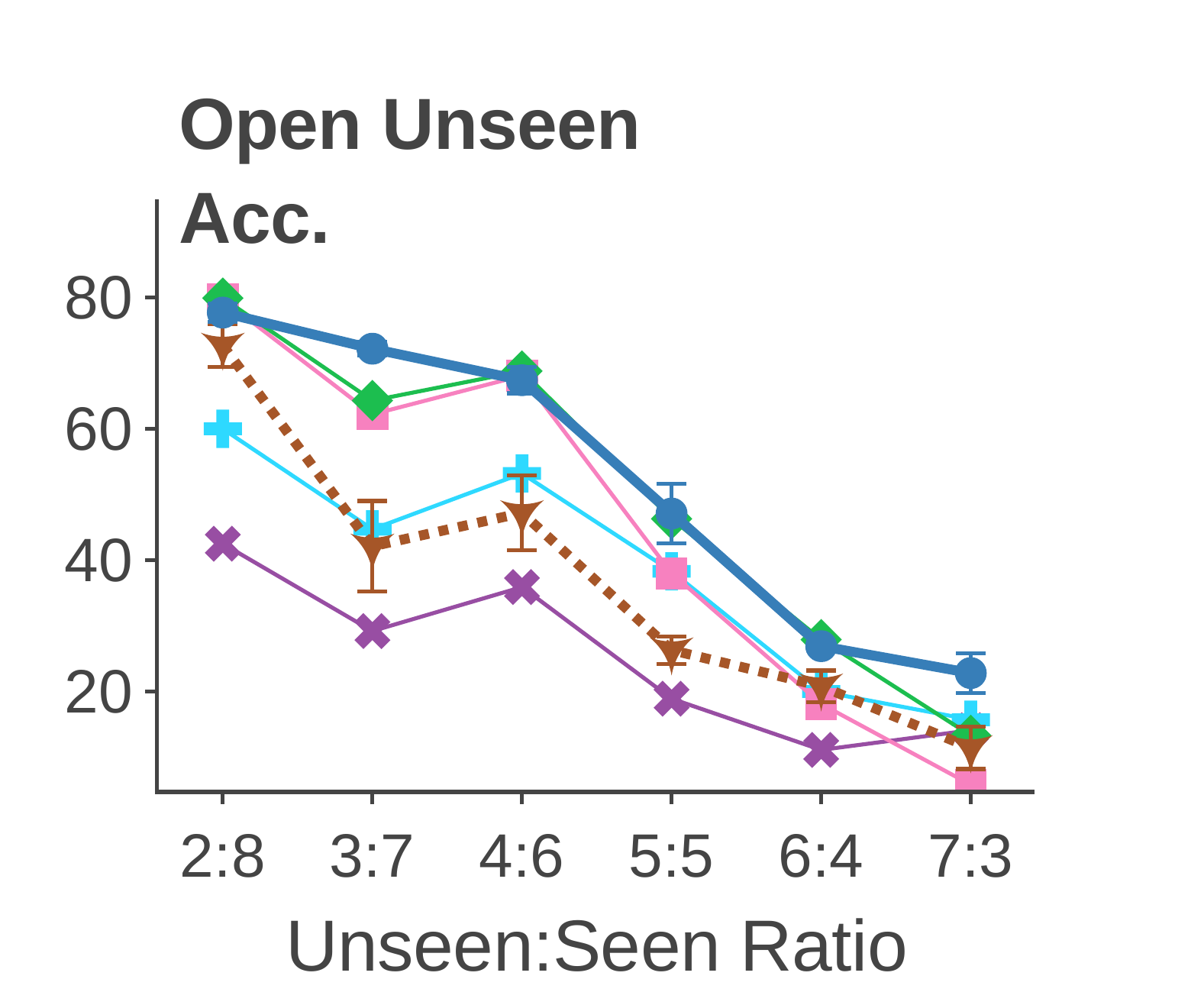}   
  \includegraphics[width=\mywidth\textwidth, trim={0cm 0cm 2.5cm 1.5cm},clip]{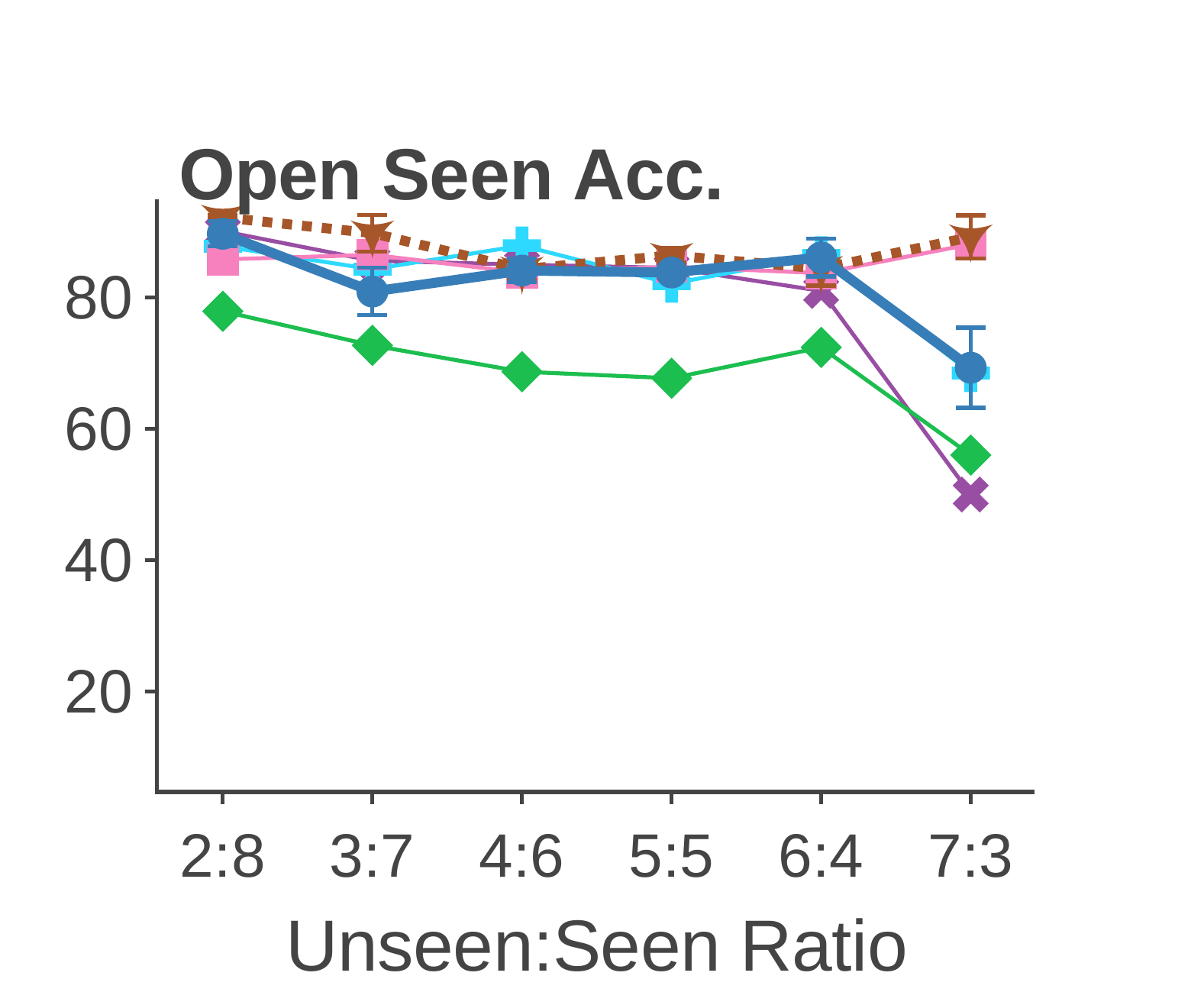} 
  \includegraphics[width=\mywidth\textwidth, trim={0cm 0cm 2.5cm 1.5cm},clip]{neurips2020/camera_ready/figures/H_acc.pdf} 
  \includegraphics[width=\mywidth\textwidth, trim={0cm 0cm 2.5cm 1.5cm},clip]{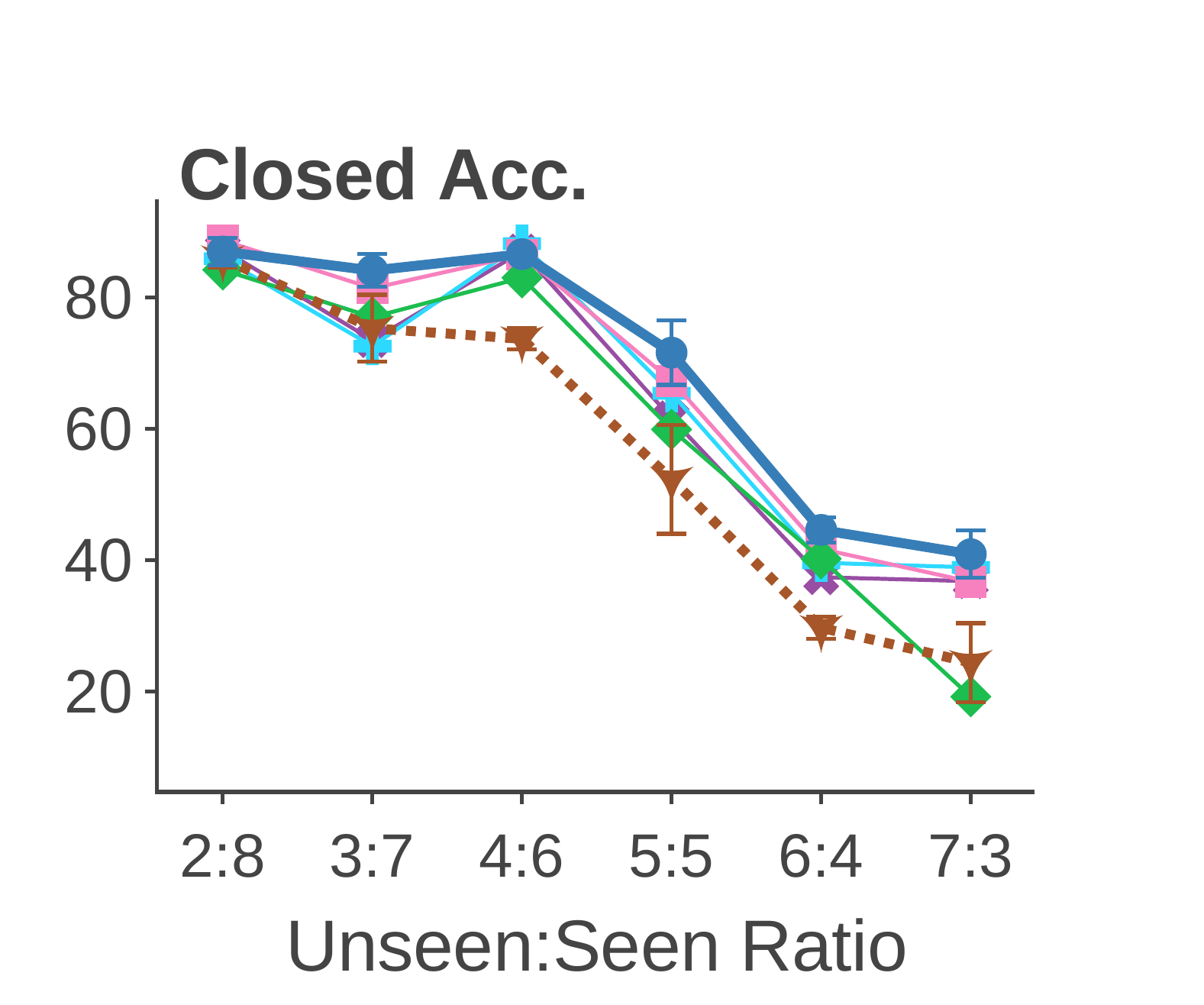} 
\\
  \hspace{-15pt}
  \hspace{0.13\textwidth}
  \includegraphics[width=\mywidth\textwidth, trim={0cm 0cm 2.5cm 1.5cm},clip]{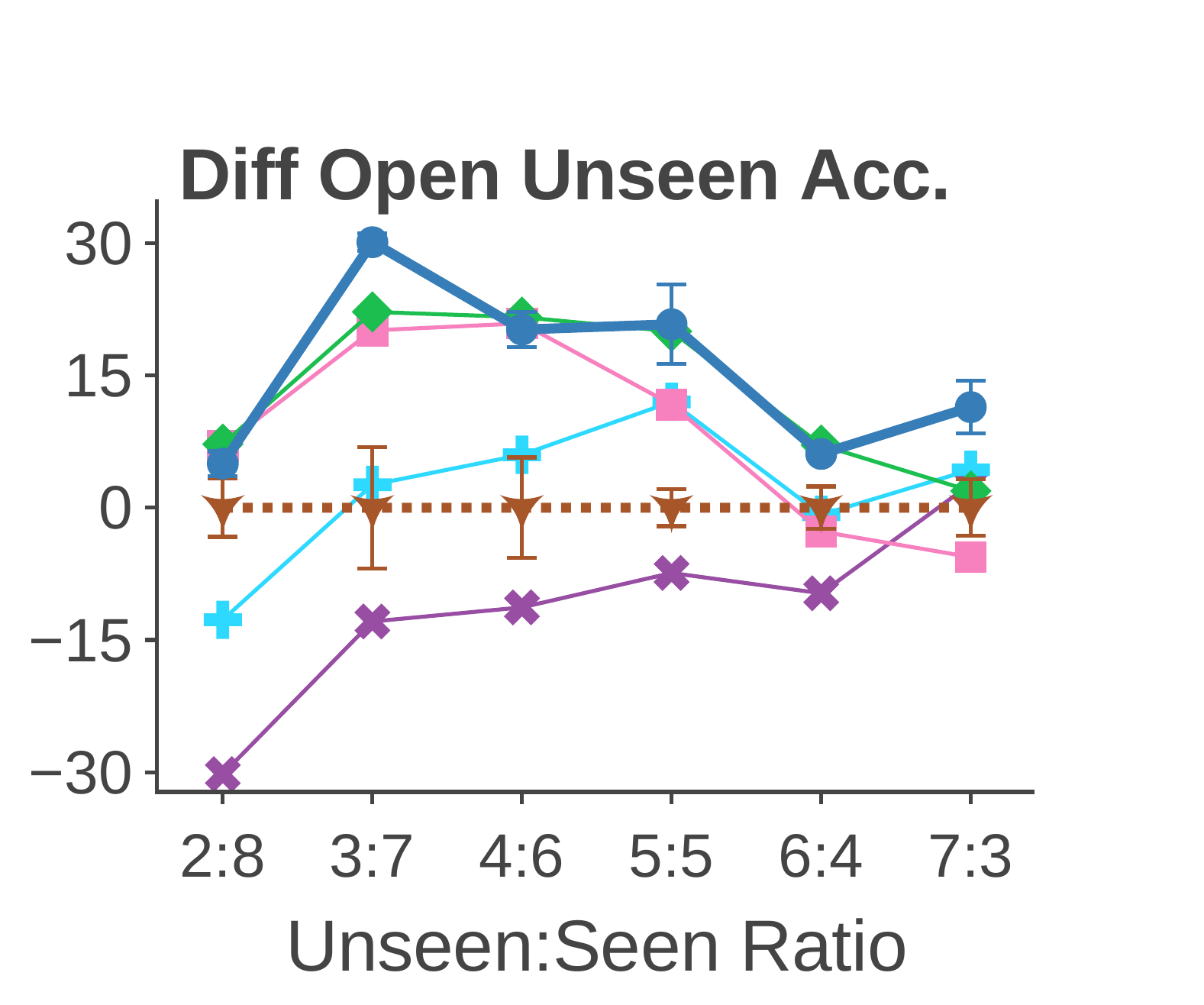} 
  \includegraphics[width=\mywidth\textwidth, trim={0cm 0cm 2.5cm 1.5cm},clip]{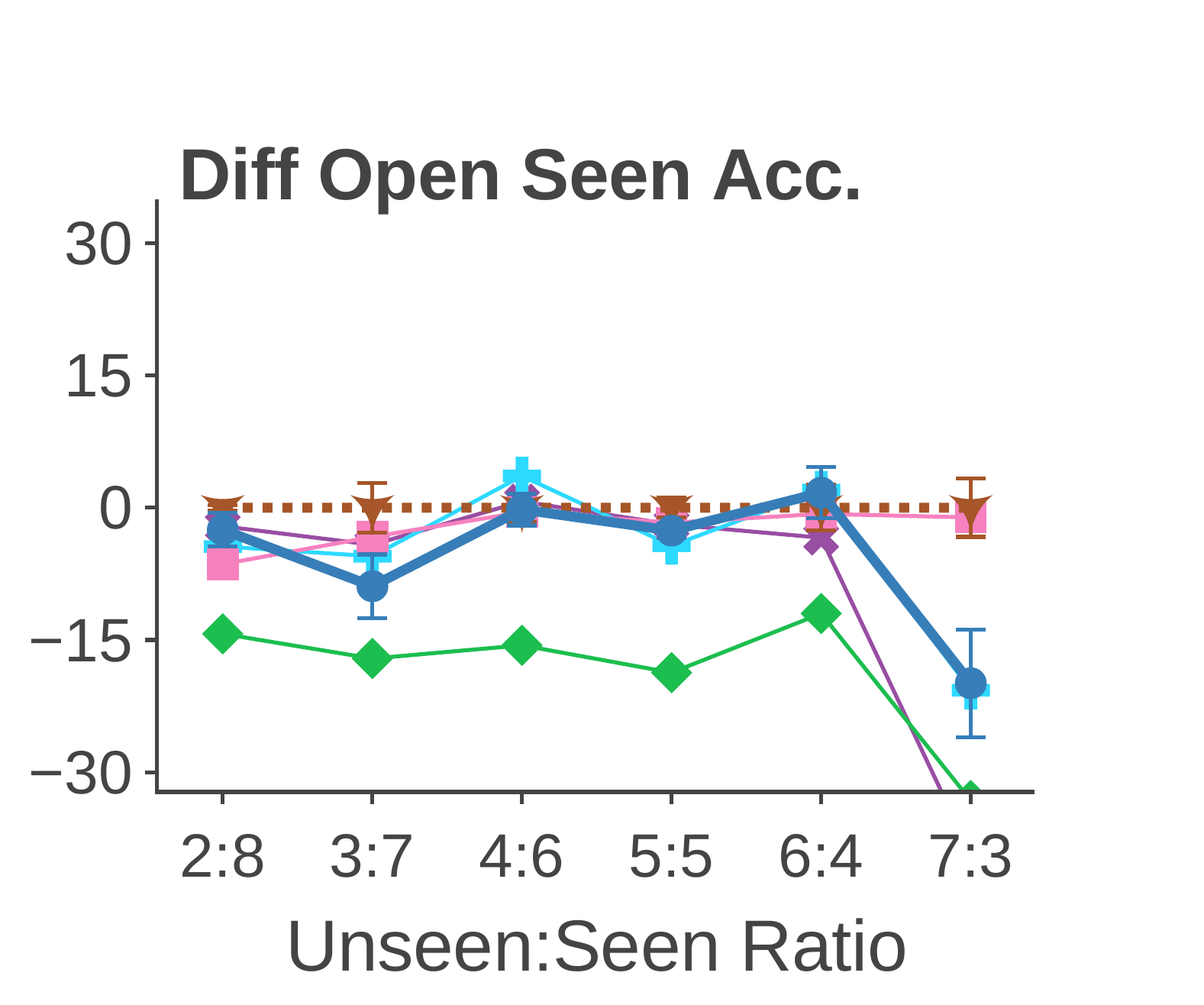} 
  \includegraphics[width=\mywidth\textwidth, trim={0cm 0cm 2.5cm 1.5cm},clip]{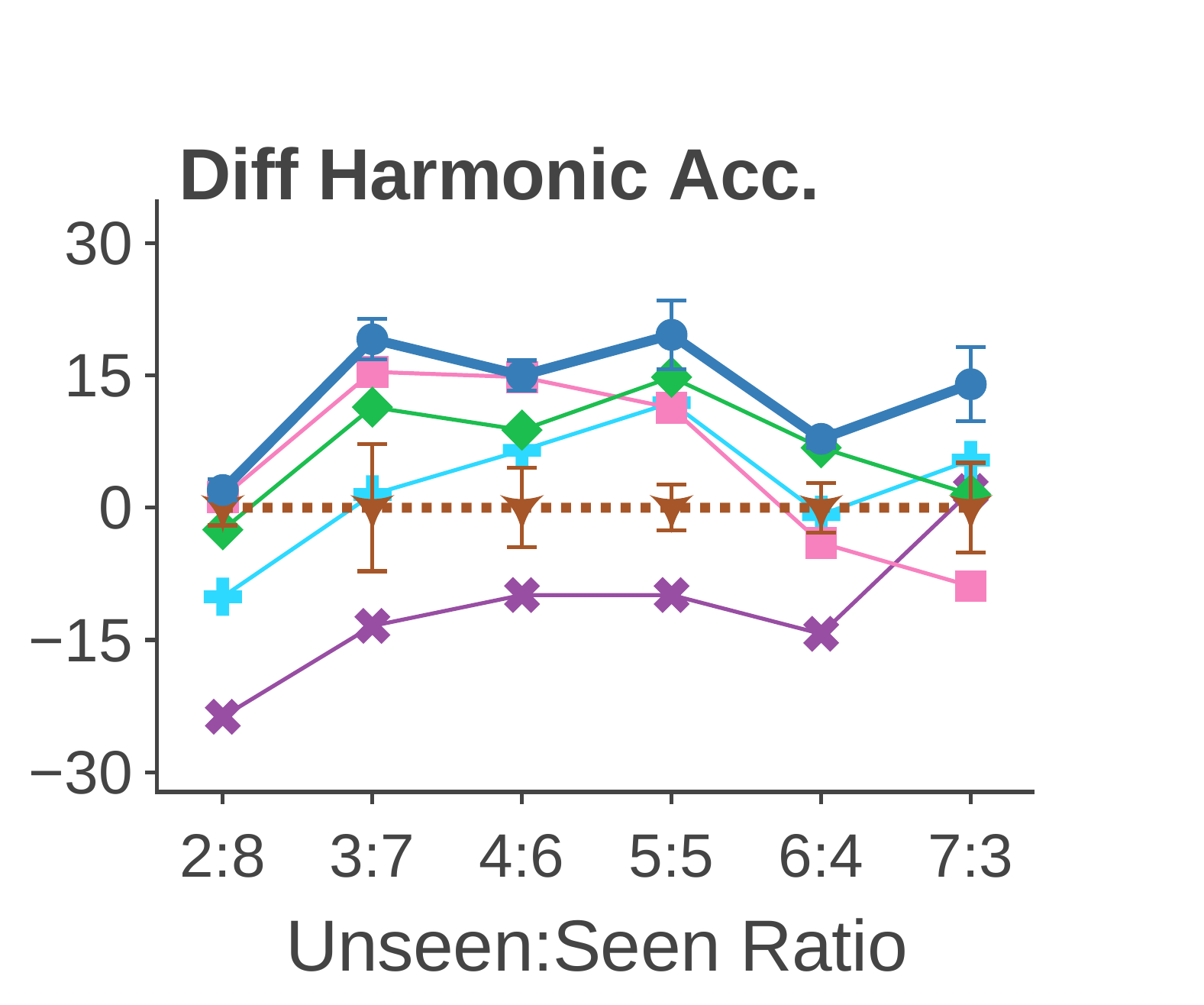}
  \includegraphics[width=\mywidth\textwidth, trim={0cm 0cm 2.5cm 1.5cm},clip]{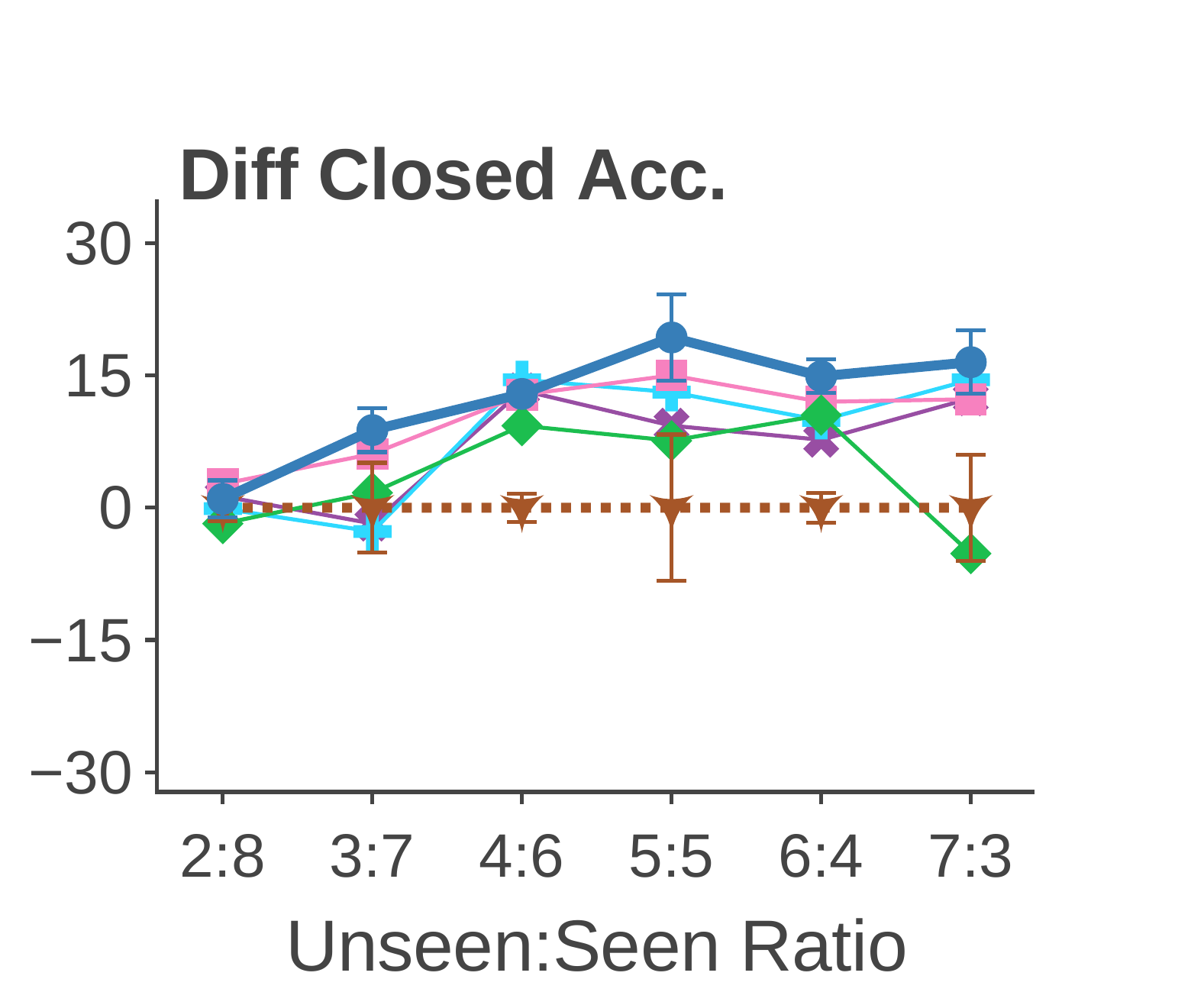} 
\caption{Accuracy metrics for AO-CLEVR on a sweep of 20\% unseen classes up to 70\% unseen classes. The top row show the measured metrics. The bottom row show the difference (\textit{subtraction}) of measured metrics from the LE baseline method. Error bars denote Standard Error of the Mean (S.E.M) over 3 random splits. To reduce visual clutter, error bars are shown only for our Causal method and for the reference baseline (LE). 
}
    \label{fig_full_results_clevr}
\end{figure}

\figref{fig_full_results_clevr} shows the accuracy metrics for compared approaches with AO-CLEVR when varying the fraction of seen:unseen classes (between 2:8 to 7:3). The top row in the figure shows the measured metrics. The bottom row shows the difference (subtraction) from LE.
We selected LE as the main reference baseline because its embedding loss approximately models $p(\x|a,o)$, but without modeling the core-features.

Across the full sweep of unseen:seen ratios, our approach ``\textit{Causal}'', performs better than or equivalent to all the compared methods for the main zero-shot metrics (Open-Unseen, Harmonic and Closed). VisProd, which approximates $p(a,o|\x)$, has a relatively low Unseen accuracy. \textit{VisProd\&CI}, the discriminatively-trained variant of our model, improves the Unseen performance by a large margin, while not hurting VisProd Seen accuracy. ATTOP is better than LE on open unseen pairs but performs substantially worse than all methods on the seen pairs. TMN performs equally well as our approach for splits with mostly seen pairs (unseen:seen @ 2:8, 3:7, 4:6), but degrades when the fraction of seen pairs reduces below 60\%.

\subsection{Baseline models without language embeddings}
\begin{figure}[h]
  \centering   
  \includegraphics[width=0.09\textwidth, trim={17.cm 2.5cm 0.5cm 1.5cm},clip]{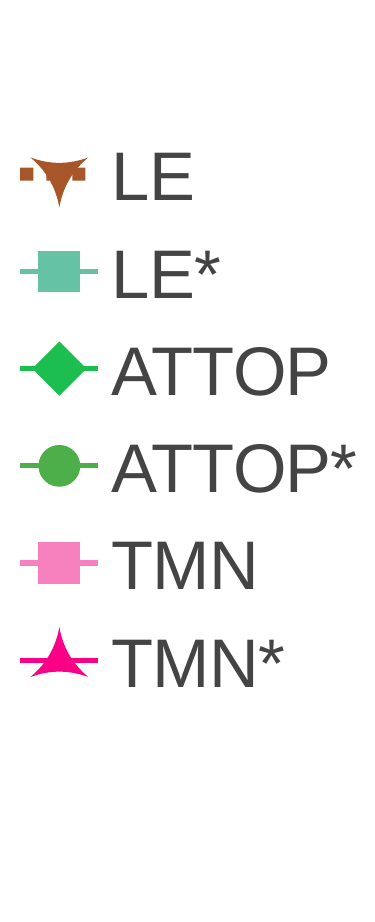}
  \includegraphics[width=\mywidth\textwidth, trim={0cm 0cm 2.5cm 1.3cm},clip ]{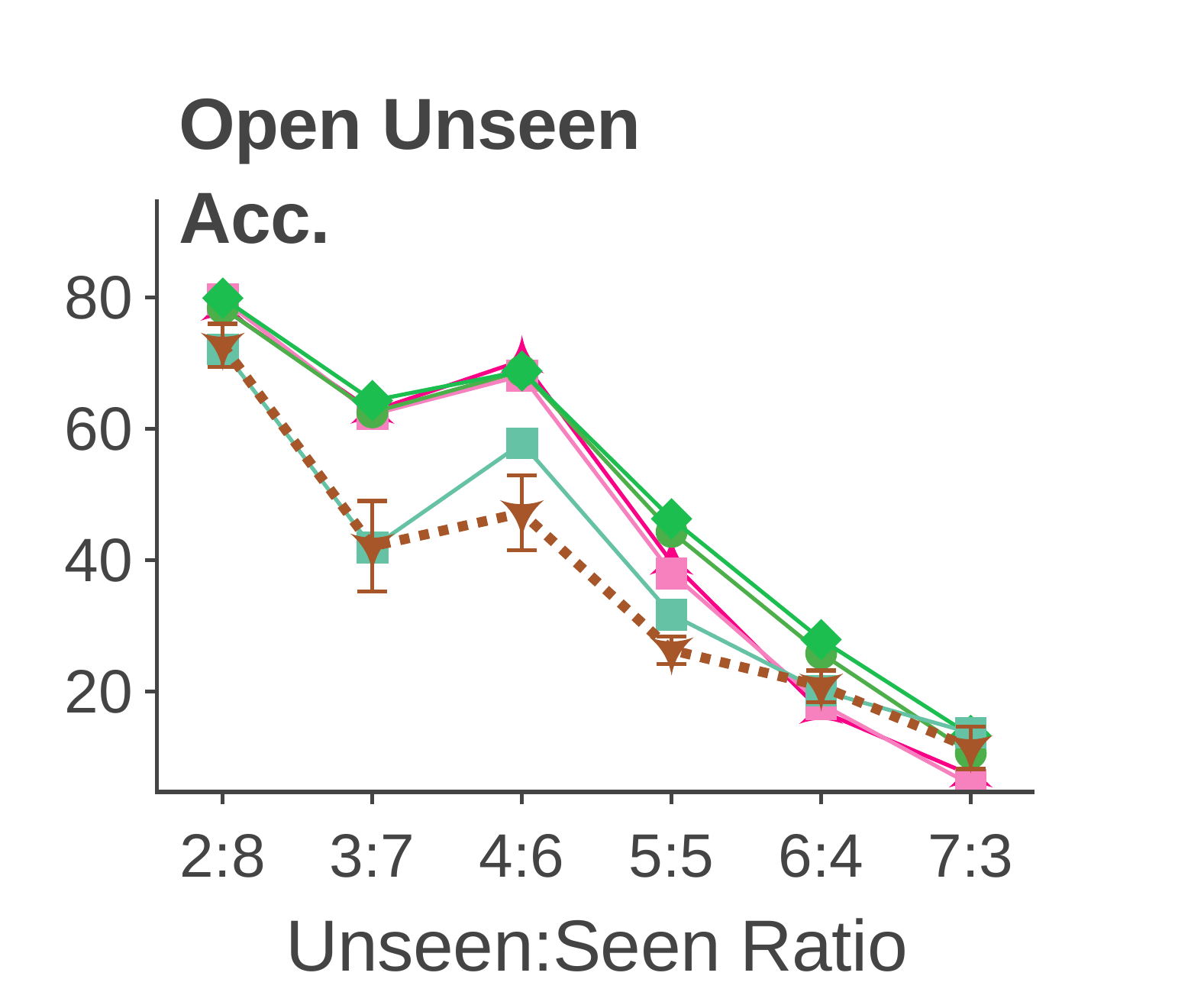}   
  \includegraphics[width=\mywidth\textwidth, trim={0cm 0cm 2.5cm 1.5cm},clip]{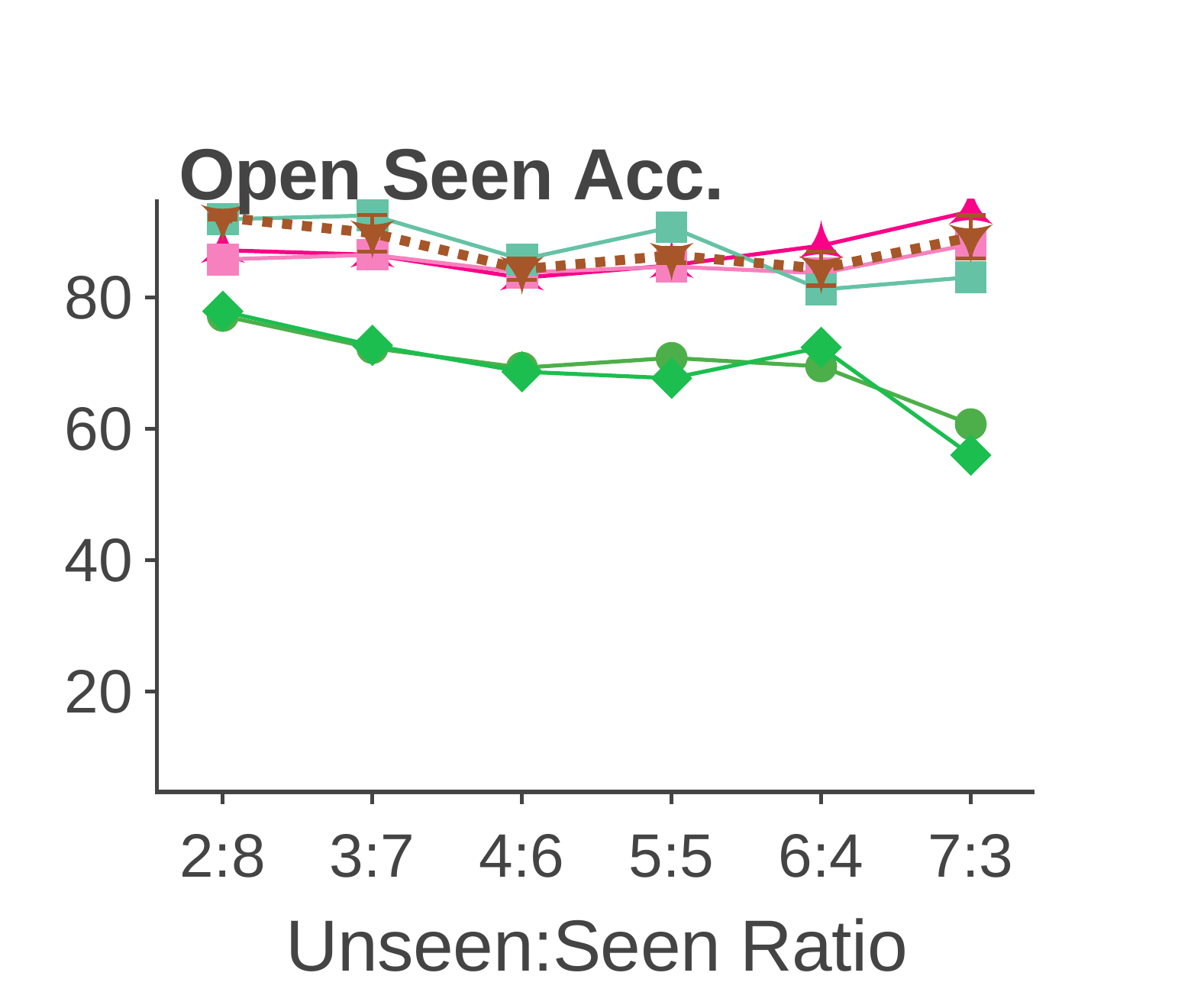} 
  \includegraphics[width=\mywidth\textwidth, trim={0cm 0cm 2.5cm 1.5cm},clip]{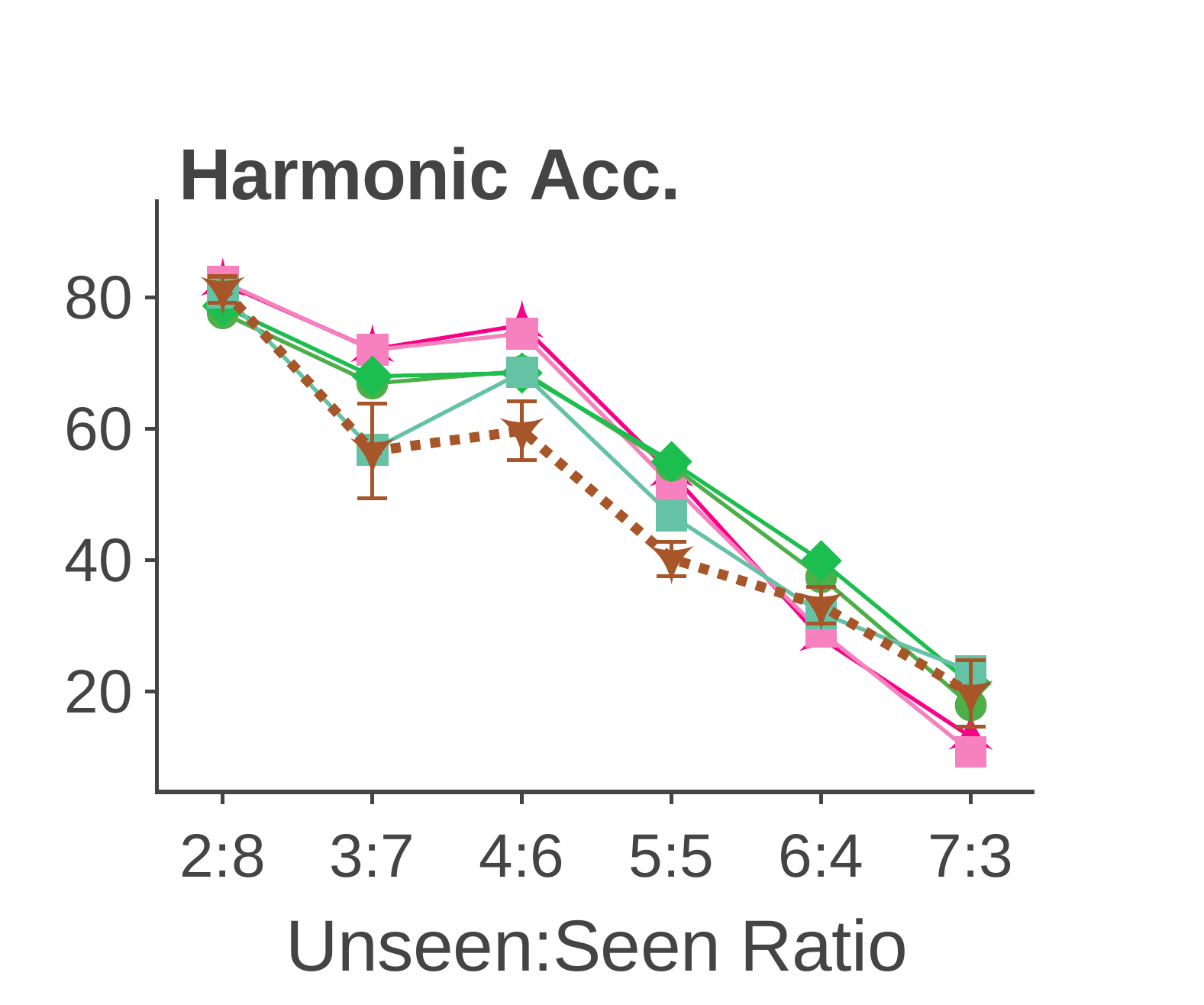} 
  \includegraphics[width=\mywidth\textwidth, trim={0cm 0cm 2.5cm 1.5cm},clip]{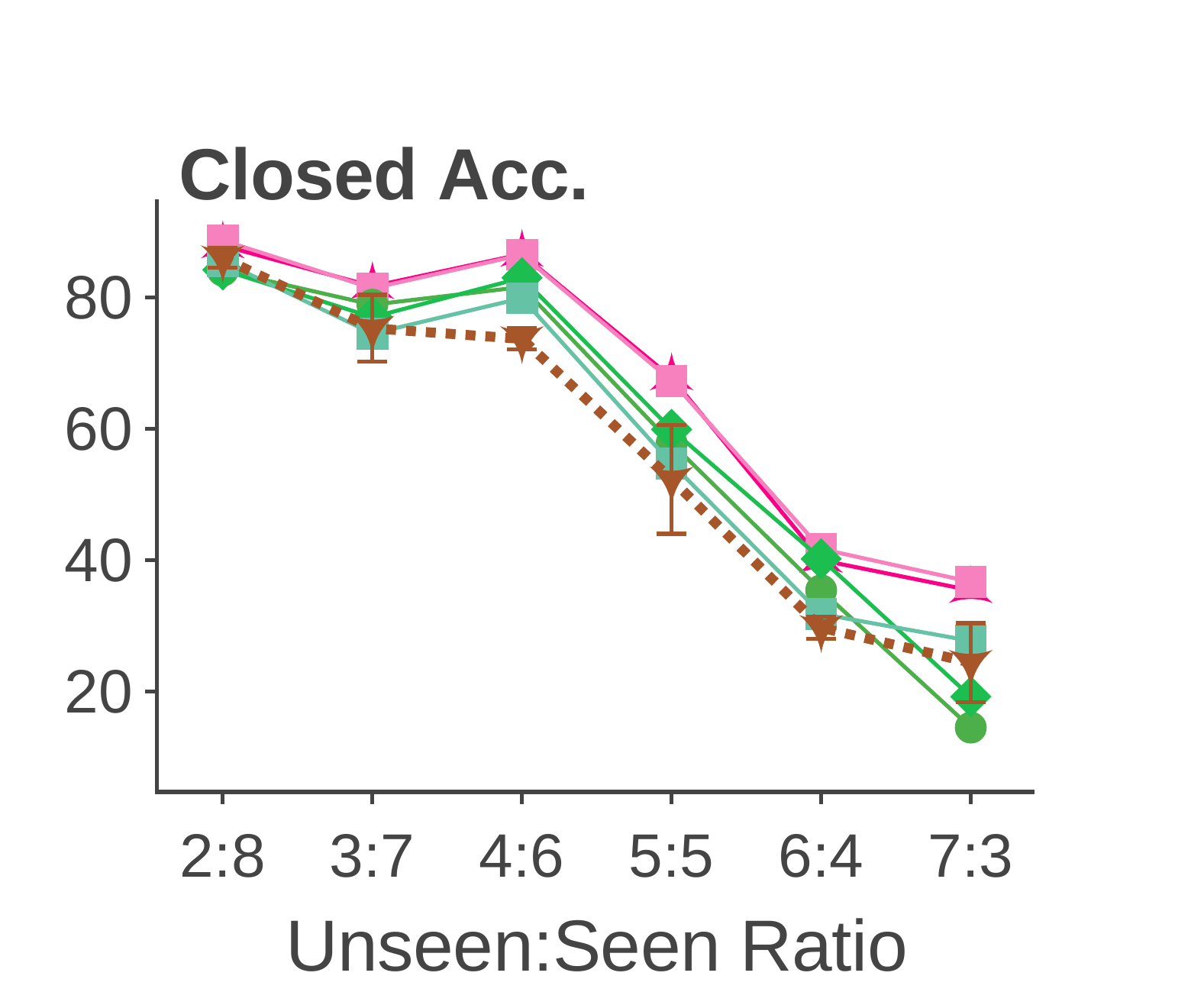} 

\caption{Comparing LE, ATTOP, and TMN with and without initialization by Glove word embedding. Glove initialization somewhat hurts LE, improves ATTOP, and is mostly equivalent for TMN.}
    \label{fig_compare_prior}
\end{figure}

\figref{fig_compare_prior} compares LE, ATTOP, and TMN with and without initialization by Glove word embedding. It demonstrates that for AO-CLEVr, Glove initialization somewhat hurts LE, improves ATTOP, and is mostly equivalent to TMN.

\section{Ablation study }
\label{sec:ablation}
To understand the contribution of the different components of our approach, we conducted an ablation study to quantify the effect of the components. We report test metrics for one of the 5:5 ``overlapping'' splits of AO-CLEVr. Specifically, the split used for hyper-parameter search.

Table \ref{tab:ablation_ao_clevr} reports the test metrics when ablating different components of our approach: We first compared the different components of the model while using alternate-training (see implementation details). Next, we compared the alternate-training strategy to standard (non-alternate) training.
Finally, we compared the different components of the conditional-independence loss.

We compared the following components: 
\begin{enumerate}
    \item ``Causal'' is our approach described in \secref{sec:approach}. We tested is with both alternate training, and standard (non-alternate) training. 
    \item $\lambda_{indep}=0$ indicates nullifying the loss terms that encourage the conditional independence relations.
    \item $\lambda_{ao}=0$ indicates nullifying the embedding to the image space $\X$.
    \item $\lambda_{invert}=0$ indicates nullifying the term that preserves information about source labels of the attribute and object embeddings.
    \item $\lambda_{rep}=0$ indicates nullifying $\L_{rep}$: The term that encourages invariance between $\hatphia$ to $\hatphio$.
    \item $\lambda_{oh}=0$ indicates nullifying $\L_{oh}$: The term that encourages invariance of $\hatphia$ to a categorical representation of an object (and similarly for $\hatphio$).
\end{enumerate}

First, we find that both alternate-training and standard non-alternate training result in a comparable Harmonic metric. However, alternate training has a %
\edit{better Unseen accuracy (57.5\% vs 53.3\%), but lower Seen accuracy (85.7\% vs 90.6\%)}

\begin{table}
{
    \begin{small}\begin{sc}
    \scalebox{0.82}{
    \setlength\tabcolsep{2pt} %
\begin{tabular}[t]{l|l|l|l|l}
\toprule
{} &        Unseen &         Seen &        Harmonic &        Closed \\
\midrule

\edit{Causal}                         &  \textbf{57.5 $\pm$ 2.3} &  85.7 $\pm$ 3.4 &  \textbf{68.8 $\pm$ 2.1} &  73.8 $\pm$ 1.0 \\
\edit{$\lambda_{indep}=0$}     &  30.0 $\pm$ 1.5 &  \textbf{97.2 $\pm$ 0.2} &  45.7 $\pm$ 1.8 &  68.4 $\pm$ 1.0 \\
\edit{$\lambda_{ao}=0$}        &  44.9 $\pm$ 1.9 &  83.5 $\pm$ 5.6 &  58.1 $\pm$ 0.3 &  72.9 $\pm$ 3.0 \\
\edit{$\lambda_{invert}=0$}   &  19.5 $\pm$ 2.2 &  46.5 $\pm$ 2.9 &  27.3 $\pm$ 2.3 &  28.5 $\pm$ 4.6 \\
\midrule
\multicolumn{5}{l}{Non-alternate training} \\

\edit{Causal}                     &  53.3 $\pm$ 2.0 &  90.6 $\pm$ 2.3 &  67.0 $\pm$ 1.5 &  \textbf{74.5 $\pm$ 1.2} \\
\edit{$\lambda_{rep}=0$}           &  53.0 $\pm$ 4.7 &  91.3 $\pm$ 1.4 &  66.7 $\pm$ 3.3 &  70.7 $\pm$ 1.3 \\
\edit{$\lambda_{oh}=0$}             &  52.8 $\pm$ 4.0 &  90.5 $\pm$ 0.6 &  66.5 $\pm$ 3.0 &  71.9 $\pm$ 2.0 \\
\edit{$\lambda_{indep}=0$} &  38.0 $\pm$ 2.3 &  94.6 $\pm$ 0.4 &  54.1 $\pm$ 2.4 &  68.9 $\pm$ 0.6 \\
\bottomrule
\end{tabular}

\hspace{10pt}
\begin{tabular}[t]{l|l|l|l|l}
    \toprule
    {} &        Unseen &         Seen &        Harmonic &        Closed \\
    \midrule
    \multicolumn{5}{l}{With prior embeddings} \\
    LE      &  21.4 $\pm$ 1.1 &  84.1 $\pm$ 1.8 &  34.0 $\pm$ 1.3 &  34.2 $\pm$ 2.4 \\
    ATTOP    &  48.7 $\pm$ 0.5 &  73.5 $\pm$ 0.8 &  58.5 $\pm$ 0.1 &  58.2 $\pm$ 0.5 \\
    TMN      &  32.3 $\pm$ 2.8 &  87.3 $\pm$ 4.1 &  47.0 $\pm$ 3.3 &  65.1 $\pm$ 3.0 \\
    \midrule
    \multicolumn{5}{l}{No prior embeddings} \\
    VisProd             &  19.1 $\pm$ 1.3 &  94.3 $\pm$ 1.1 &  31.7 $\pm$ 1.8 &  60.1 $\pm$ 0.2 \\
    LE*            &  28.2 $\pm$ 1.7 &  87.5 $\pm$ 0.5 &  42.5 $\pm$ 1.9 &  43.4 $\pm$ 2.7 \\
    ATTOP*         &  45.6 $\pm$ 0.5 &  76.3 $\pm$ 1.0 &  57.0 $\pm$ 0.1 &  54.6 $\pm$ 1.3 \\
    TMN*           &  36.6 $\pm$ 5.2 &  89.1 $\pm$ 3.5 &  51.6 $\pm$ 5.6 &  66.5 $\pm$ 4.3 \\
    \midrule
    \edit{VisP\&CI}    &  40.5 $\pm$ 2.7 &  84.7 $\pm$ 4.3 &  54.4 $\pm$ 1.8 &  59.9 $\pm$ 0.3 \\
    \bottomrule
\end{tabular}
}
\end{sc}\end{small}
}
\caption{\textbf{Left: }Ablation study on a 5:5 split of AO-CLEVr. We use the split used for hyper-param search. $\pm$ denote S.E.M on 3 random initializations. \textbf{Right: }Reference metrics for baselines.}\label{tab:ablation_ao_clevr}
\end{table}

Second, nullifying each of the major components of the loss has a substantial impact on the performance of the model. Specifically, (1) nullifying $\lambda_{indep}$ reduces the Harmonic from \edit{68.8\% to 45.7\%}, (2) nullifying $\lambda_{ao}$ reduces the Harmonic to \edit{58.1\%}, (3) nullifying $\lambda_{invert}$ reduces the Harmonic to \edit{27.3}\%.

Finally, $\L_{oh}$ and $\L_{rep}$ have a synergistic effect on the performance \edit{of AO-CLEVr}. Their individual performance metrics are comparable, but jointly they improve the %
Closed accuracy from \edit{$\tildeapprox$71.5\% to 74.5\%}.

\begin{figure}[h]
    \centering
     \includegraphics[width=0.50\linewidth]{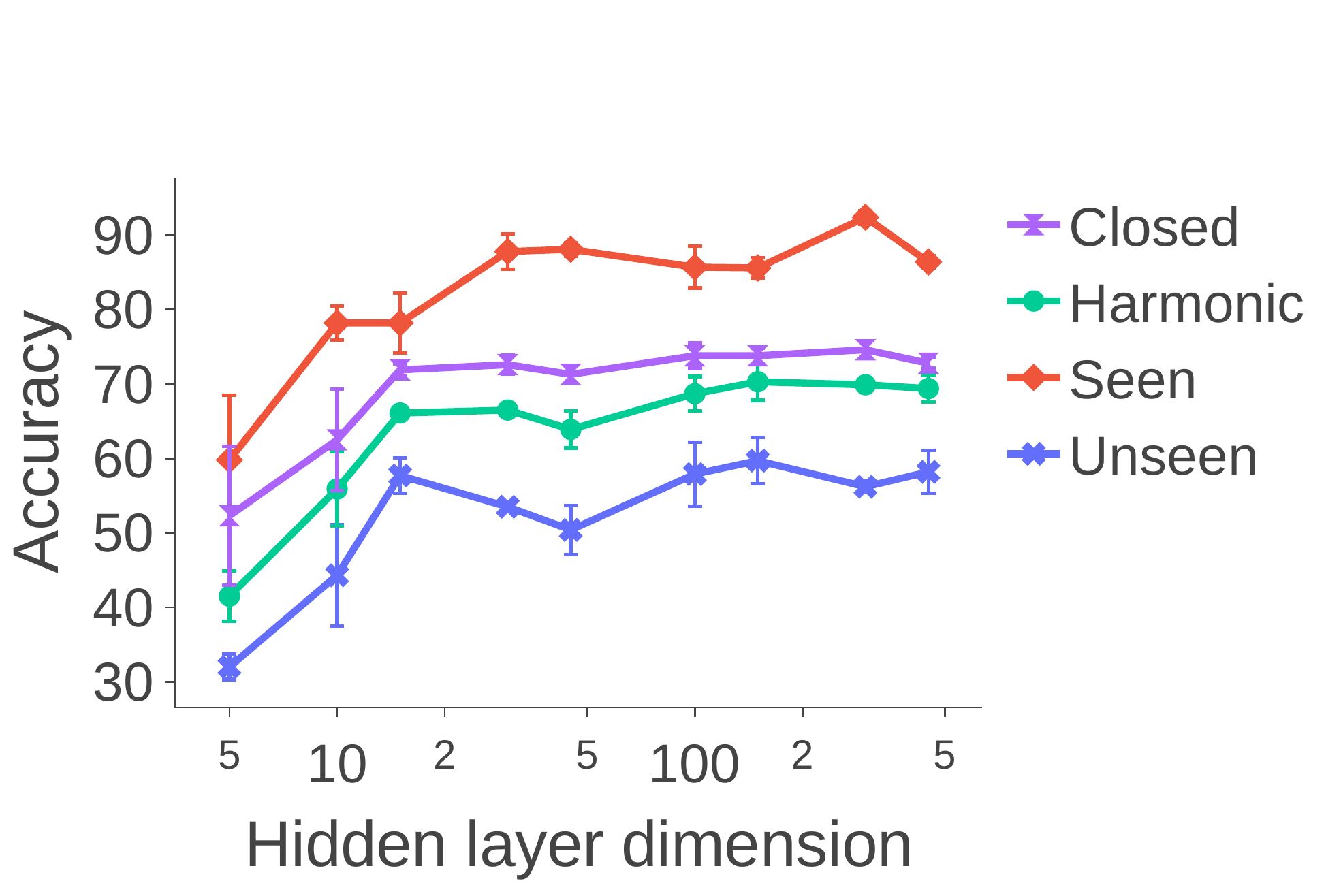}
    \caption{Ablation: Accuracy versus hidden layer size (in a logarithmic scale)} 
    \label{fig:AO_CLEVr_acc_vs_dim}
\end{figure}

\paragraph{Accuracy versus hidden layer size:}
\edit{
\figref{fig:AO_CLEVr_acc_vs_dim} shows the different accuracy metrics when changing the hidden layer size for the 5:5 split of AO-CLEVr. It shows that the seen accuracy increases with the layer size, presumably because it can capture more subtleties of the seen pairs. The unseen accuracy appears to be bi-modal, with peaks at 15-units and 150 units.  Indeed the cross-validation procedure selected a layer size of 150, because it maximized the harmonic-mean of the seen and unseen accuracy.}

\section{Error analysis}
\label{sec:error_analysis}
\subsection{Zappos}

\edit{We analyzed the \textit{errors} that \textit{Causal} makes when recognizing unseen pairs in Zappos (open). In $53\%$ of cases the object was predicted correctly, in $14\%$ the attribute was predicted correctly and in $33\%$ neither. It appears that in Zappos, recognizing the object transferred better to the unseen pairs, presumably because  recognizing the attributes is harder in this dataset. }

\edit{
To gain further intuition, we compared the errors that \textit{Causal} makes to those of \textit{LE*}, the strongest no-prior baseline. 
With \textit{Causal}, 39\% of unseen pairs (U) are confused for seen pairs (S), and 36\% of unseen pairs are confused for incorrect unseen-pairs.
This yields a balanced rate of $\frac{U\rightarrow S}{U\rightarrow U} = \frac{39\%}{36\%} \approx 1.1$.
For comparison \textit{LE*} errors are largely unbalanced: $\frac{U\rightarrow S}{U\rightarrow U} = \frac{67\%}{19\%} \approx 3.5$. 
}

\begin{figure}[h]
  \centering   
  (a) Ground-truth = \textit{(Suede, Slippers)}
  \includegraphics[width=0.87\textwidth, trim={0.cm 13.2cm 0.cm 7cm},clip]{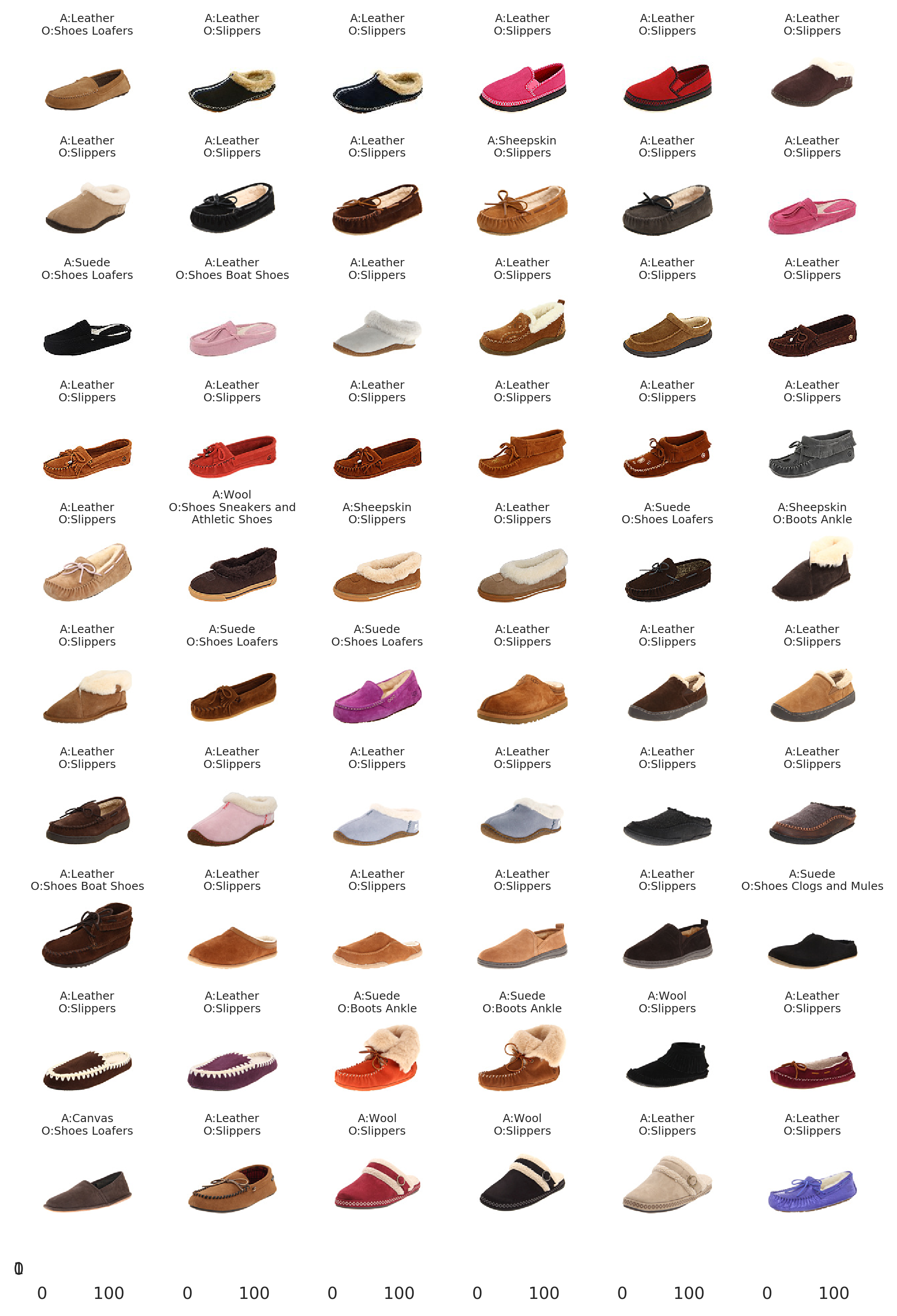}  \\ \vspace{2pt} %
  (b) Ground-truth = \textit{(Hair.Calf, Shoes.Heels)}
  \includegraphics[width=0.87\textwidth, trim={0.cm 4.1cm 0.cm 0cm},clip]{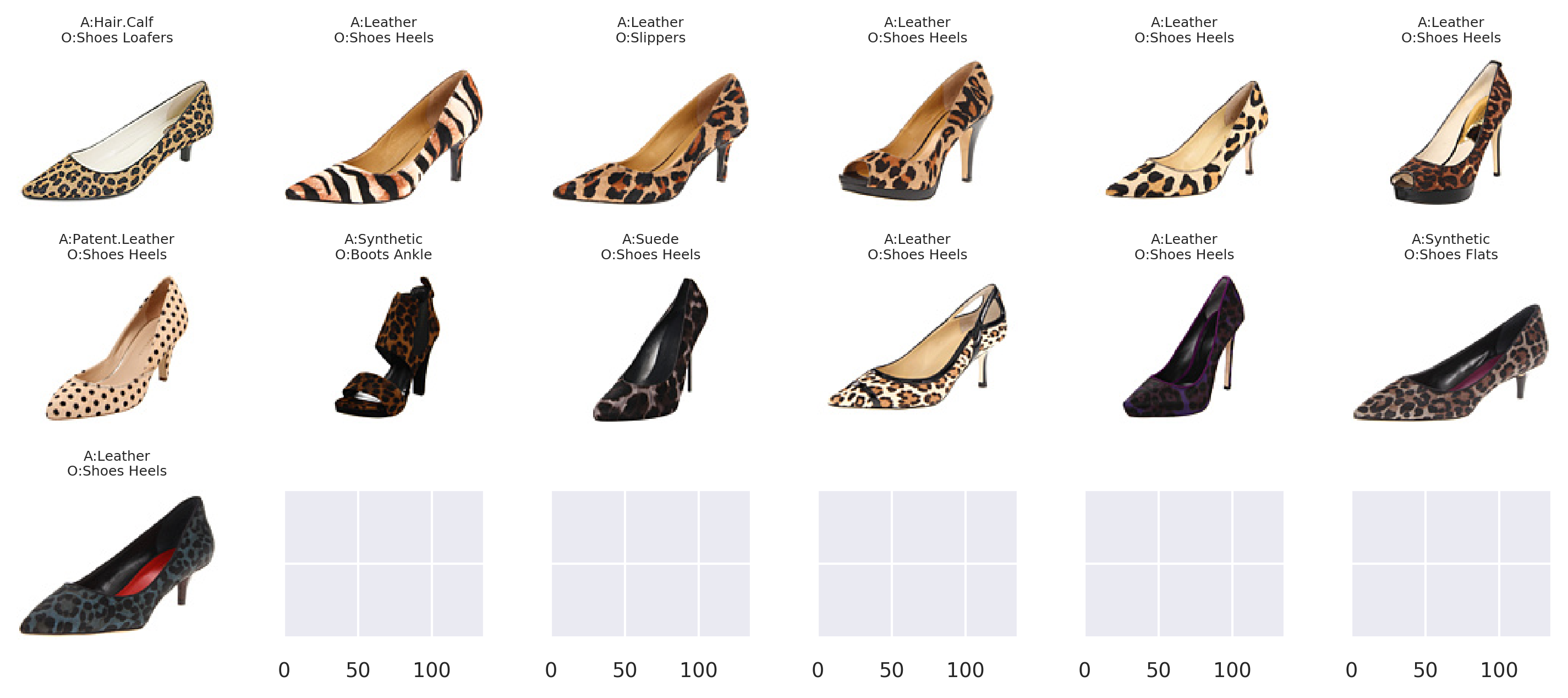} \\ \vspace{2pt} %
  (c) Ground-truth = \textit{(Patent.Leather , Shoes.Heels)}
  \includegraphics[width=0.87\textwidth, trim={0.cm 15.cm 0.cm 0cm},clip]{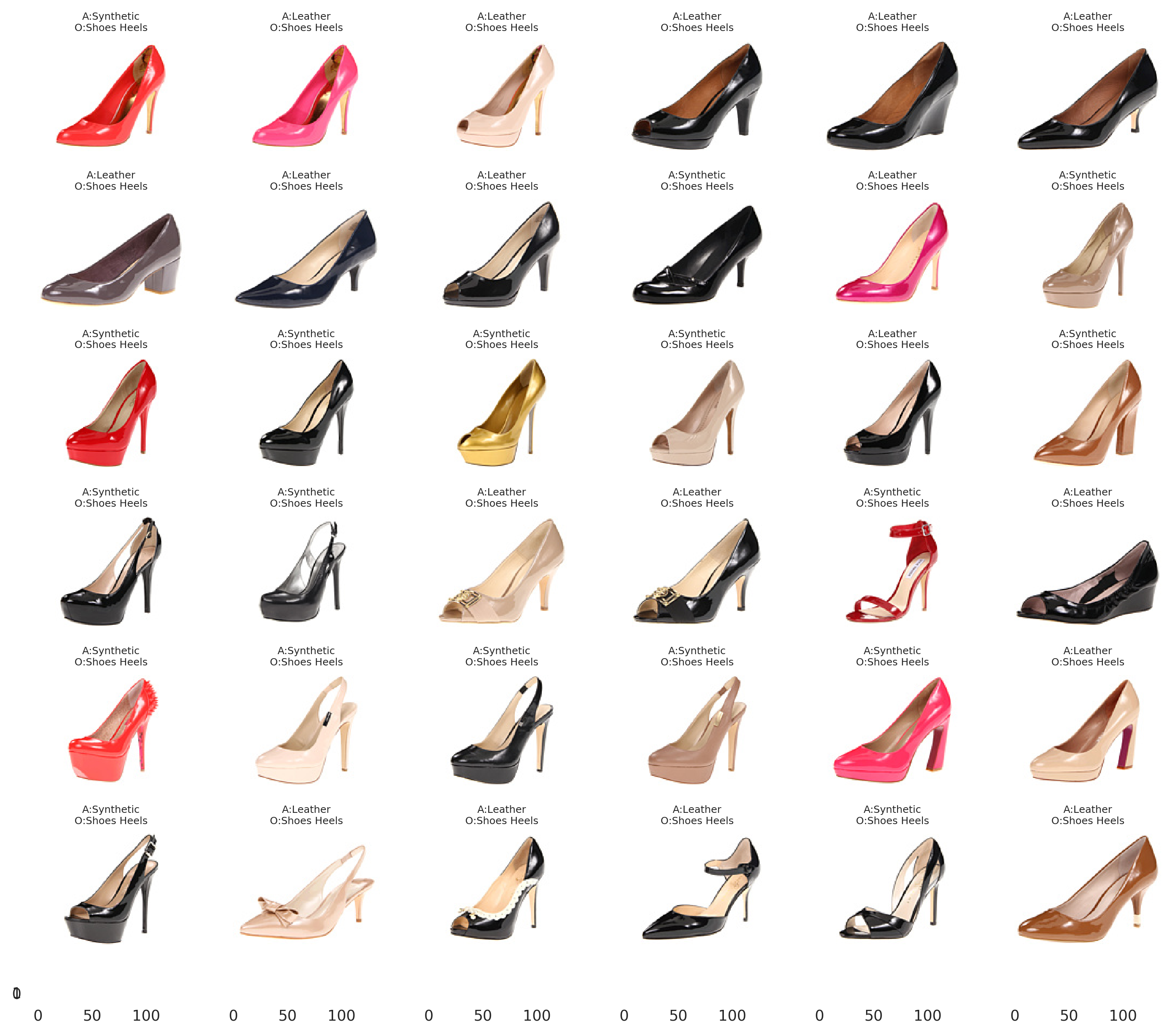} \\ %
    \caption{Qualitative examples of \textit{Causal} success cases for Zappos: Every image was predicted correctly by \textit{Causal}, but erroneously by \textit{LE*}. Above every image we denote the erroneous  prediction of \textit{LE*}.
    (a) %
    An unseen pair: the error rate was 40\% for \textit{Causal}, while it was 82\% for \textit{LE*}. (b) %
    An unseen pair: the error rate was 39\% for \textit{Causal}, while it was 58\% for \textit{LE*}. (c) %
    A \textbf{seen} pair: the error rate was 56\% for \textit{Causal}, while it was 83\% for \textit{LE*}.
    }
    \label{fig:zappos_success_cases}
\end{figure}

\edit{We further analyzed wins and losses of \textit{Causal} compared with \textit{LE*}, which are illustrated in \figref{fig:zappos_success_cases}. \textit{Causal} succeeded to overcome common failures of \textit{LE*}, sometimes overcoming domain-shifts that exist within the seen pairs. 
The main weakness of \textit{Causal} is that the error rate for seen pairs is higher compared to \textit{LE*}. This effect is strongest for (1) \big(A=\textit{Leather}, O=\textit{Boots.Ankle}\big), which was mostly confused for either A=\textit{Full.grain.leather} or A=\textit{Suede}, while the object-class was correctly classified. (2) \big(A=\textit{Leather}, O=\textit{Boots.Mid-Calf}\big), which was mostly confused for A=\textit{Faux.Leather} while the object-class was correctly classified. 
This result shows that \textit{Causal} is less biased toward predicting \textit{Leather}, which is the most common attribute in the training set.  }

\subsection{AO-CLEVr}
\edit{We analyzed the errors that Causal makes when recognizing unseen pairs at the $5\!\!:\!\!5$ split\footnote{The split used for ablation study} of AO-CLEVr (open). In $15\%$ of cases the object was predicted correctly, in $82\%$ the attribute was predicted correctly and in $3\%$ neither. It appears that in this case, recognizing the attribute transferred better to the unseen pairs, because in this dataset colors can be easily recognized. }

\edit{
Comparing the errors that Causal makes to those of \textit{TMN*}, the strongest no-prior baseline show that with \textit{Causal}, 19\% of unseen pairs (U) are confused for seen pairs (S), and 20\% of unseen-pairs are confused for incorrect unseen-pairs, a balanced rate of $\frac{U\rightarrow S}{U\rightarrow U} = \frac{19\%}{20\%} \approx 0.9$.
However, with \textit{TMN*}, confusion is largely unbalanced: $\frac{U\rightarrow S}{U\rightarrow U} = \frac{41\%}{13\%} \approx 3$. 
}

\begin{figure}[!h]
    \centering
     \includegraphics[width=0.80\linewidth]{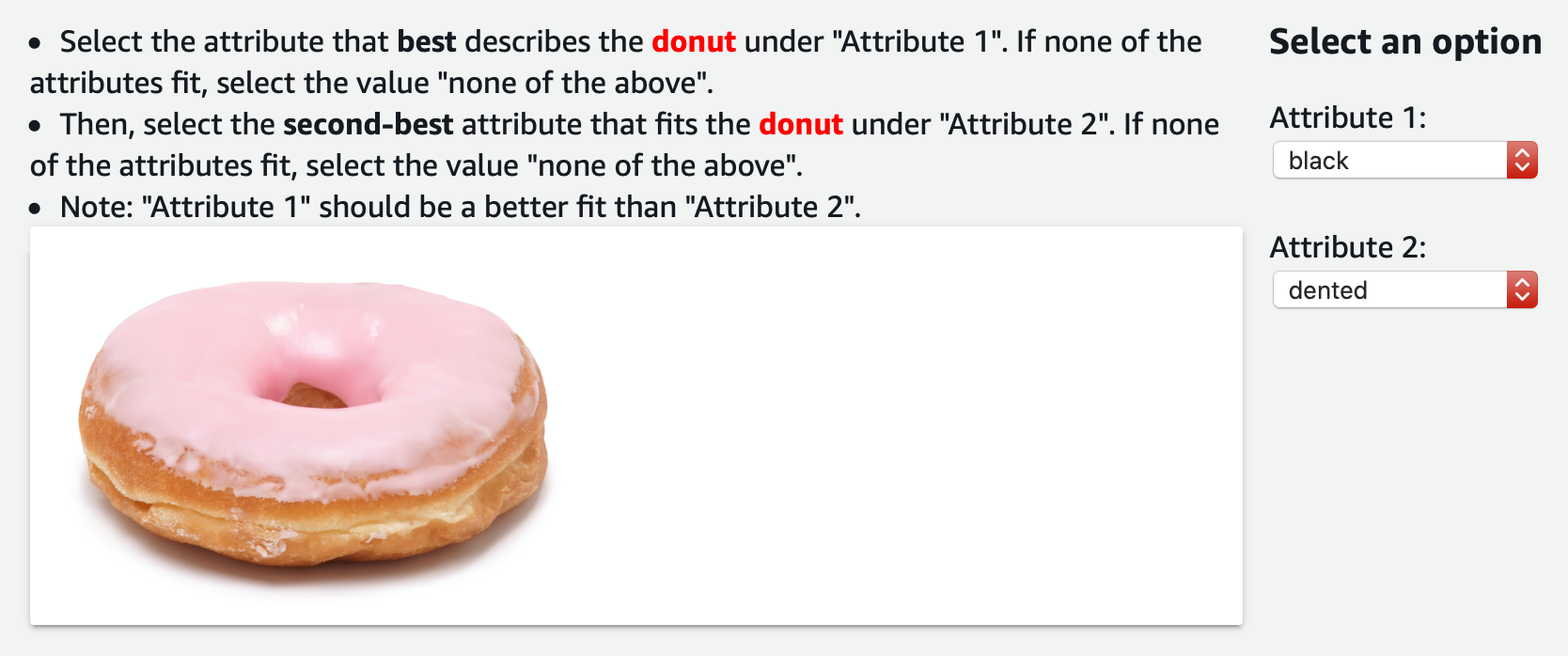}
    \caption{An example of a task on Amazon Mechanical Turk.} 
    \label{fig:amt_exmaple}
\end{figure}

\section{Label-quality evaluation: Human-Rater Experiments}
\label{sec:AMT}
To better understand the level of label noise in \textit{MIT-states} dataset we conducted an experiment with human raters. 

Each rater was presented with an image of an object and was asked to select \edit{the best and second-best} attributes that describe this object from a pre-defined list. The list was comprised of attributes that co-occur with the given object in the dataset. For example, the object ``apple'' had candidate attributes ``green'', ``yellow''  and ``red'', but not ``black''. Raters were also presented with an option ``none of the above'' and an option ``I don't know'', see Figure~\ref{fig:amt_exmaple}. 
We sampled 500 instances of objects and attributes, one from each attribute-object pair. As a label-quality score, we computed the balanced accuracy of rater responses compared with the label provided by the dataset across the 500 tasks. \edit{To verify that raters were attentive we also introduced a ``sanity'' set of 30 instances of objects for which there were two clear attributes as answers.} We also recorded the rate at which raters chose the ``none of the above'' and ``I don't know'' answers as a proxy for the difficulty of assigning labels to the dataset. Balanced accuracy was computed by averaging the accuracy per attribute. 

The average rater \edit{top-1 and top-2 accuracies were 31.79\% and 47\% respectively}, indicating a label noise level of $\tildeapprox$70\%. The fraction of tasks that raters selected ``none of the above'' or ``I don't know'' was 5\%, indicating that raters were confident in about $\tildeapprox$95\% of their rating. \edit{The top-1 accuracy on the ``sanity'' set was 88\% and the top-2 accuracy was 100\%, indicating that the raters were attentive and capable of solving the task at hand.}

\begin{figure}[h]
  \centering   
  \includegraphics[width=0.87\textwidth]{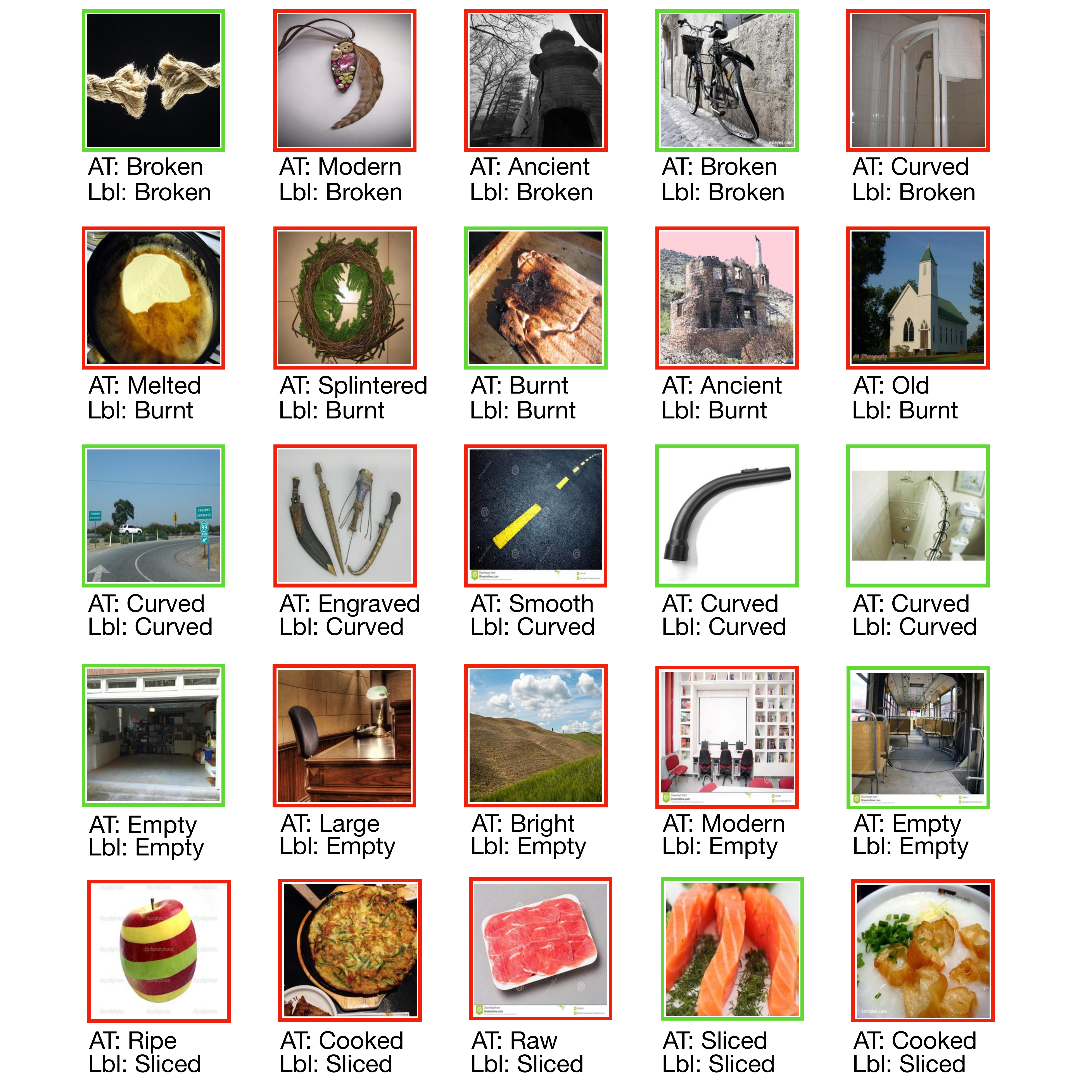}
    \caption{Label quality of MIT-States. Showing 5 attribute labels, selected by random. For each label, we show 5 images, selected by random. For each image, we show the choice of the amazon-turker (AT) and the provided attribute label (Lbl). Green image margins indicate that the turker choice agrees with the label. Red margins indicate that the turker choice disagrees with the label.}
    \label{fig_amt_MIT}
\end{figure}

Finally, Fig. \ref{fig_amt_MIT} shows qualitative examples for the label quality of MIT-States. For each label of 5 attribute labels, selected by random, we show 5 images, selected by random. Under each image, we show the choice of the amazon-turker in the label-quality experiment and the provided attribute label.

\newpage
\clearpage

\section{\edit{Numeric values for the metrics}}
\label{sec:numeric_all}

\subsection{Overlapping split}

\begin{table}[h!]
{
    \begin{small}\begin{sc}
    \scalebox{0.65}{
\setlength\tabcolsep{3.5pt} %
\begin{tabular}{l|llll|llll|llll}
\toprule
U:S & \multicolumn{4}{l}{Causal} & \multicolumn{4}{l}{VisProd\&CI} & \multicolumn{4}{l}{VisProd} \\
{} &        Unseen &        Seen &             Harmonic &        Closed &        Unseen &        Seen &             Harmonic &        Closed &        Unseen &        Seen &             Harmonic &        Closed \\
\midrule
2:8 &  77.7 $\pm$ 1.4 &  89.7 $\pm$ 1.9 &  83.2 $\pm$ 1.2 &  87.0 $\pm$ 2.1 &  60.0 $\pm$ 2.7 &  87.8 $\pm$ 2.3 &  71.1 $\pm$ 1.7 &  85.9 $\pm$ 1.0 &  42.5 $\pm$ 3.0 &  90.1 $\pm$ 1.4 &  57.5 $\pm$ 2.9 &  87.3 $\pm$ 1.3 \\
3:7 &  72.2 $\pm$ 1.0 &  80.9 $\pm$ 3.6 &  75.7 $\pm$ 2.3 &  84.1 $\pm$ 2.5 &  44.7 $\pm$ 5.0 &  84.3 $\pm$ 3.8 &  58.1 $\pm$ 4.9 &  72.6 $\pm$ 7.5 &  29.2 $\pm$ 3.7 &  85.6 $\pm$ 3.4 &  43.2 $\pm$ 4.4 &  73.5 $\pm$ 6.9 \\
4:6 &  67.4 $\pm$ 2.0 &  84.1 $\pm$ 1.8 &  74.7 $\pm$ 1.7 &  86.6 $\pm$ 0.7 &  53.2 $\pm$ 1.5 &  87.9 $\pm$ 1.1 &  66.2 $\pm$ 0.9 &  88.2 $\pm$ 1.0 &  35.9 $\pm$ 3.6 &  85.0 $\pm$ 2.2 &  49.8 $\pm$ 2.9 &  87.0 $\pm$ 1.6 \\
5:5 &  47.1 $\pm$ 4.5 &  83.8 $\pm$ 0.8 &  59.8 $\pm$ 3.9 &  71.6 $\pm$ 4.9 &  38.3 $\pm$ 1.1 &  82.1 $\pm$ 4.2 &  52.1 $\pm$ 1.8 &  65.4 $\pm$ 3.6 &  18.9 $\pm$ 0.2 &  84.5 $\pm$ 8.5 &  30.3 $\pm$ 0.8 &  61.6 $\pm$ 2.3 \\
6:4 &  26.9 $\pm$ 0.5 &  86.1 $\pm$ 2.9 &  40.9 $\pm$ 1.0 &  44.6 $\pm$ 1.9 &  20.0 $\pm$ 1.8 &  86.4 $\pm$ 1.0 &  32.3 $\pm$ 2.4 &  39.6 $\pm$ 2.6 &  11.1 $\pm$ 1.4 &  81.0 $\pm$ 9.5 &  18.8 $\pm$ 1.5 &  37.4 $\pm$ 3.5 \\
7:3 &  22.8 $\pm$ 3.0 &  69.3 $\pm$ 6.1 &  33.7 $\pm$ 4.2 &  40.9 $\pm$ 3.6 &  15.7 $\pm$ 2.5 &  68.5 $\pm$ 6.6 &  25.1 $\pm$ 3.4 &  38.9 $\pm$ 1.6 &  14.1 $\pm$ 1.5 &  50.0 $\pm$ 8.6 &  21.6 $\pm$ 2.4 &  36.8 $\pm$ 2.7 \\
\bottomrule
\end{tabular}

}
\end{sc}\end{small}
}
\end{table}

\begin{table}[h!]
{
    \begin{small}\begin{sc}
    \scalebox{0.65}{
\setlength\tabcolsep{3.5pt} %
\begin{tabular}{l|llll|llll|llll}
\toprule
U:S & \multicolumn{4}{l}{LE*} & \multicolumn{4}{l}{LE} & \multicolumn{4}{l}{ATTOP*} \\
{} &        Unseen &        Seen &             Harmonic &        Closed &        Unseen &        Seen &             Harmonic &        Closed &        Unseen &        Seen &             Harmonic &        Closed \\
\midrule
2:8 &  72.1 $\pm$ 2.8 &  91.9 $\pm$ 0.2 &  80.7 $\pm$ 1.7 &  85.5 $\pm$ 1.8 &  72.7 $\pm$ 3.3 &  92.2 $\pm$ 0.3 &  81.2 $\pm$ 2.0 &  86.0 $\pm$ 1.5 &  78.4 $\pm$ 3.3 &  77.2 $\pm$ 1.0 &  77.6 $\pm$ 1.4 &  84.1 $\pm$ 4.0 \\
3:7 &  41.9 $\pm$ 6.3 &  92.5 $\pm$ 0.3 &  56.8 $\pm$ 6.4 &  74.5 $\pm$ 4.5 &  42.1 $\pm$ 6.9 &  89.8 $\pm$ 2.8 &  56.6 $\pm$ 7.2 &  75.3 $\pm$ 5.1 &  62.5 $\pm$ 2.3 &  72.3 $\pm$ 1.4 &  66.9 $\pm$ 1.2 &  78.9 $\pm$ 4.1 \\
4:6 &  57.8 $\pm$ 4.5 &  85.8 $\pm$ 1.5 &  68.6 $\pm$ 2.7 &  79.9 $\pm$ 2.7 &  47.2 $\pm$ 5.7 &  84.3 $\pm$ 1.6 &  59.7 $\pm$ 4.5 &  73.7 $\pm$ 1.6 &  68.8 $\pm$ 1.2 &  69.3 $\pm$ 2.8 &  68.8 $\pm$ 0.9 &  81.6 $\pm$ 2.1 \\
5:5 &  31.7 $\pm$ 1.6 &  90.7 $\pm$ 1.4 &  46.8 $\pm$ 1.9 &  54.7 $\pm$ 5.6 &  26.3 $\pm$ 2.1 &  86.4 $\pm$ 1.1 &  40.2 $\pm$ 2.6 &  52.3 $\pm$ 8.3 &  44.3 $\pm$ 1.1 &  70.8 $\pm$ 2.3 &  54.4 $\pm$ 1.2 &  57.8 $\pm$ 2.5 \\
6:4 &  20.1 $\pm$ 1.0 &  81.2 $\pm$ 4.7 &  31.9 $\pm$ 0.9 &  31.8 $\pm$ 1.3 &  20.8 $\pm$ 2.4 &  84.4 $\pm$ 2.6 &  33.1 $\pm$ 2.8 &  29.7 $\pm$ 1.7 &  25.8 $\pm$ 1.9 &  69.5 $\pm$ 2.2 &  37.4 $\pm$ 1.7 &  35.4 $\pm$ 0.3 \\
7:3 &  13.7 $\pm$ 2.2 &  83.1 $\pm$ 3.8 &  23.1 $\pm$ 3.3 &  27.7 $\pm$ 3.4 &  11.4 $\pm$ 3.2 &  89.2 $\pm$ 3.3 &  19.7 $\pm$ 5.1 &  24.4 $\pm$ 6.0 &  10.6 $\pm$ 1.7 &  60.7 $\pm$ 7.7 &  17.9 $\pm$ 2.7 &  14.5 $\pm$ 1.6 \\
\bottomrule
\end{tabular}

}
\end{sc}\end{small}
}

\end{table}

\begin{table}[h!]
{
    \begin{small}\begin{sc}
    \scalebox{0.65}{
\setlength\tabcolsep{3.5pt} %
\begin{tabular}{l|llll|llll|llll}
\toprule
U:S & \multicolumn{4}{l}{ATTOP} & \multicolumn{4}{l}{TMN*} & \multicolumn{4}{l}{TMN} \\
{} &        Unseen &        Seen &             Harmonic &        Closed &        Unseen &        Seen &             Harmonic &        Closed &        Unseen &        Seen &             Harmonic &        Closed \\
\midrule
2:8 &  79.9 $\pm$ 3.5 &  77.9 $\pm$ 1.1 &  78.7 $\pm$ 1.3 &  84.2 $\pm$ 3.9 &  78.4 $\pm$ 5.2 &  87.2 $\pm$ 0.8 &  82.2 $\pm$ 2.7 &  87.9 $\pm$ 1.5 &  79.7 $\pm$ 4.4 &  85.8 $\pm$ 0.9 &  82.4 $\pm$ 2.1 &  88.7 $\pm$ 1.6 \\
3:7 &  64.3 $\pm$ 3.3 &  72.7 $\pm$ 1.8 &  68.0 $\pm$ 1.4 &  77.0 $\pm$ 3.9 &  62.7 $\pm$ 6.1 &  86.5 $\pm$ 0.4 &  72.1 $\pm$ 4.2 &  81.7 $\pm$ 5.0 &  62.2 $\pm$ 4.9 &  86.5 $\pm$ 0.3 &  72.0 $\pm$ 3.4 &  81.4 $\pm$ 5.0 \\
4:6 &  68.8 $\pm$ 1.2 &  68.7 $\pm$ 3.1 &  68.5 $\pm$ 1.4 &  83.0 $\pm$ 2.0 &  70.4 $\pm$ 3.0 &  83.0 $\pm$ 3.2 &  75.8 $\pm$ 0.5 &  86.6 $\pm$ 1.3 &  68.1 $\pm$ 4.1 &  83.8 $\pm$ 2.6 &  74.5 $\pm$ 1.6 &  86.5 $\pm$ 2.5 \\
5:5 &  46.3 $\pm$ 1.2 &  67.7 $\pm$ 2.5 &  55.0 $\pm$ 1.5 &  59.9 $\pm$ 2.8 &  39.7 $\pm$ 3.7 &  84.9 $\pm$ 3.7 &  53.2 $\pm$ 2.4 &  67.8 $\pm$ 5.1 &  38.0 $\pm$ 3.0 &  84.7 $\pm$ 2.2 &  51.5 $\pm$ 2.2 &  67.3 $\pm$ 4.7 \\
6:4 &  27.9 $\pm$ 2.3 &  72.4 $\pm$ 2.7 &  39.9 $\pm$ 2.0 &  40.2 $\pm$ 0.3 &  17.0 $\pm$ 2.6 &  87.9 $\pm$ 1.2 &  28.1 $\pm$ 3.6 &  40.0 $\pm$ 0.9 &  18.1 $\pm$ 2.9 &  83.7 $\pm$ 0.4 &  29.1 $\pm$ 3.7 &  41.7 $\pm$ 0.7 \\
7:3 &  13.3 $\pm$ 1.5 &  56.0 $\pm$ 5.2 &  21.2 $\pm$ 2.2 &  19.2 $\pm$ 2.1 &   7.3 $\pm$ 1.6 &  93.1 $\pm$ 3.3 &  13.1 $\pm$ 2.7 &  35.4 $\pm$ 2.2 &   5.8 $\pm$ 0.9 &  88.1 $\pm$ 2.5 &  10.8 $\pm$ 1.6 &  36.7 $\pm$ 1.0 \\
\bottomrule
\end{tabular}
}
\end{sc}\end{small}
}
\caption{Numeic values for results of \figref{fig_full_results_clevr} (top row)}
\end{table}

\subsection{Non-overlapping split}

\begin{table}[h!]
{
    \begin{small}\begin{sc}
    \scalebox{0.65}{
\setlength\tabcolsep{3.5pt} %
\begin{tabular}{l|llll|llll|llll}
\toprule
U:S & \multicolumn{4}{l}{Causal} & \multicolumn{4}{l}{LE} & \multicolumn{4}{l}{ATTOP} \\
{} &        Unseen &        Seen &             Harmonic &        Closed &        Unseen &        Seen &             Harmonic &        Closed &        Unseen &        Seen &             Harmonic &        Closed \\
\midrule
2:8 &  64.3 $\pm$ 1.0 &  79.4 $\pm$ 1.5 &  70.8 $\pm$ 1.0 &  82.1 $\pm$ 0.7 &  35.4 $\pm$ 5.1 &  80.1 $\pm$ 3.7 &  48.7 $\pm$ 5.5 &  71.2 $\pm$ 2.6 &  53.4 $\pm$ 3.7 &  67.8 $\pm$ 3.2 &   59.1 $\pm$ 1.1 &   76.0 $\pm$ 5.4 \\
3:7 &  48.7 $\pm$ 4.7 &  75.3 $\pm$ 4.6 &  58.9 $\pm$ 5.0 &  79.0 $\pm$ 5.5 &  22.9 $\pm$ 2.5 &  84.5 $\pm$ 2.3 &  35.7 $\pm$ 3.3 &  52.5 $\pm$ 6.8 &  37.4 $\pm$ 5.5 &  65.6 $\pm$ 2.4 &   46.3 $\pm$ 4.2 &  61.1 $\pm$ 10.9 \\
4:6 &  43.5 $\pm$ 4.6 &  69.2 $\pm$ 4.2 &  53.2 $\pm$ 4.5 &  66.8 $\pm$ 5.3 &  27.5 $\pm$ 2.3 &  80.7 $\pm$ 2.9 &  40.7 $\pm$ 2.6 &  43.4 $\pm$ 4.2 &  21.7 $\pm$ 9.5 &  49.2 $\pm$ 5.3 &  26.8 $\pm$ 10.6 &  40.9 $\pm$ 10.1 \\
5:5 &  15.7 $\pm$ 1.7 &  75.2 $\pm$ 7.3 &  25.8 $\pm$ 2.7 &  37.5 $\pm$ 4.9 &   9.1 $\pm$ 2.1 &  91.1 $\pm$ 0.5 &  16.3 $\pm$ 3.5 &  19.9 $\pm$ 3.6 &   6.4 $\pm$ 1.9 &  65.3 $\pm$ 7.5 &    9.9 $\pm$ 2.0 &   22.0 $\pm$ 4.2 \\
\bottomrule
\end{tabular}
}
\end{sc}\end{small}
}
\end{table}

\begin{table}[h!]
{
    \begin{small}\begin{sc}
    \scalebox{0.65}{
\setlength\tabcolsep{3.5pt} %
\begin{tabular}{l|llll}
\toprule
U:S & \multicolumn{4}{l}{TMN} \\
{} &        Unseen &        Seen &             Harmonic &        Closed \\
\midrule
2:8 &  47.2 $\pm$ 2.7 &  82.1 $\pm$ 2.9 &  59.5 $\pm$ 1.6 &  81.2 $\pm$ 2.2 \\
3:7 &  23.4 $\pm$ 4.2 &  84.2 $\pm$ 0.2 &  35.1 $\pm$ 5.4 &  64.4 $\pm$ 5.6 \\
4:6 &  15.3 $\pm$ 4.4 &  85.5 $\pm$ 3.1 &  24.7 $\pm$ 6.3 &  54.4 $\pm$ 4.7 \\
5:5 &   3.0 $\pm$ 1.1 &  86.8 $\pm$ 3.4 &   5.6 $\pm$ 2.0 &  32.7 $\pm$ 1.6 \\
\bottomrule
\end{tabular}
}
\end{sc}\end{small}
}
\caption{Numeric values for results of \figref{fig_results_clevr_UV}}
\end{table}

\end{document}